\def\paperID{4154}
\def\confName{CVPR}
\def\confYear{2025}

\def\paperTitle{FrugalNeRF: Fast Convergence for Extreme Few-shot Novel View Synthesis without Learned Priors}

\makeatletter
\newcommand{\printfnsymbol}[1]{%
        \textsuperscript{\@fnsymbol{#1}}%
}
\makeatother

\def\authorBlock{
Chin-Yang Lin$^{1}$\footnote{Authors contributed equally to the paper.}
\quad
Chung-Ho Wu$^{1}$\printfnsymbol{1}
\quad
Chang-Han Yeh$^1$
\\
Shih-Han Yen$^1$
\quad
Cheng Sun$^2$
\quad
Yu-Lun Liu$^1$\vspace{0.5em}
\\
\centerline{$^1$National Yang Ming Chiao Tung University \quad $^2$NVIDIA Research}\vspace{0.5em}
\\
{\url{https://linjohnss.github.io/frugalnerf/}}
}

\newif\ifreview 
\newif\ifarxiv \newcommand{\arxiv}{\arxivtrue}
\newif\ifcamera 
\newif\ifrebuttal 

\arxiv 

\pdfoutput=1
\documentclass[10pt,twocolumn,letterpaper]{article}
\ifreview \usepackage[review]{cvpr} \fi
\ifarxiv \usepackage[pagenumbers]{cvpr} \fi
\ifrebuttal \usepackage[rebuttal]{cvpr} \fi
\ifcamera \usepackage{cvpr} \fi

\usepackage{graphicx}	
\usepackage{amsmath}	
\usepackage{amssymb}	
\usepackage{booktabs}
\usepackage{times}
\usepackage{microtype}
\usepackage{epsfig}
\usepackage{caption}
\usepackage{float}
\usepackage{placeins}
\usepackage{color, colortbl}
\usepackage{stfloats}
\usepackage{enumitem}
\usepackage{tabularx}
\usepackage{xstring}
\usepackage{multirow}
\usepackage{xspace}
\usepackage{url}
\usepackage{subcaption}
\usepackage{xcolor}
\usepackage[hang,flushmargin]{footmisc}
\usepackage{wrapfig}

\ifcamera \usepackage[accsupp]{axessibility} \fi

% ...

\ifarxiv  \fi

\newcommand{\R}[1]{{%
    \textbf{%
        \ifstrequal{#1}{1}{\textcolor{red}{R#1}}{%
        \ifstrequal{#1}{2}{\textcolor{blue}{R#1}}{%
        \ifstrequal{#1}{3}{\textcolor{magenta}{R#1}}{%
        \ifstrequal{#1}{4}{\textcolor{teal}{R#1}}{%
                           \textcolor{cyan}{R#1}%
        }}}}%
    }%
}}

\usepackage{xr-hyper}

\makeatletter
\newcommand*{\addFileDependency}[1]{
  \typeout{(#1)}
  \@addtofilelist{#1}
  \IfFileExists{#1}{}{\typeout{No file #1.}}
}

\makeatother
\newcommand*{\myexternaldocument}[1]{
    \externaldocument{#1}
    \addFileDependency{#1.tex}
    \addFileDependency{#1.aux}
}

\definecolor{cvprblue}{rgb}{0.21,0.49,0.74}
\usepackage[pagebackref,breaklinks,colorlinks,allcolors=cvprblue]{hyperref}
\usepackage[capitalize]{cleveref}
\crefname{section}{Sec.}{Secs.}
\crefname{table}{Table}{Tables}
\crefname{figure}{Fig.}{Figs.}

\ifarxiv \crefname{appendix}{App.}{Apps.}
\else \crefname{appendix}{Suppl.}{Suppls.} \fi

\unless\ifarxiv \myexternaldocument{_supplementary} \fi

\begin{document}
\title{\paperTitle}
\author{\authorBlock}

\definecolor{teaserorange}{RGB}{242,169,59}
\definecolor{teasergreen}{RGB}{55,126,34}

\twocolumn[{%
\renewcommand\twocolumn[1][]{#1}%
\maketitle
\begin{center}
\centering
\captionsetup{type=figure}
\vspace{-4mm}
\resizebox{1.0\textwidth}{!} 
{
\includegraphics[width=\textwidth]{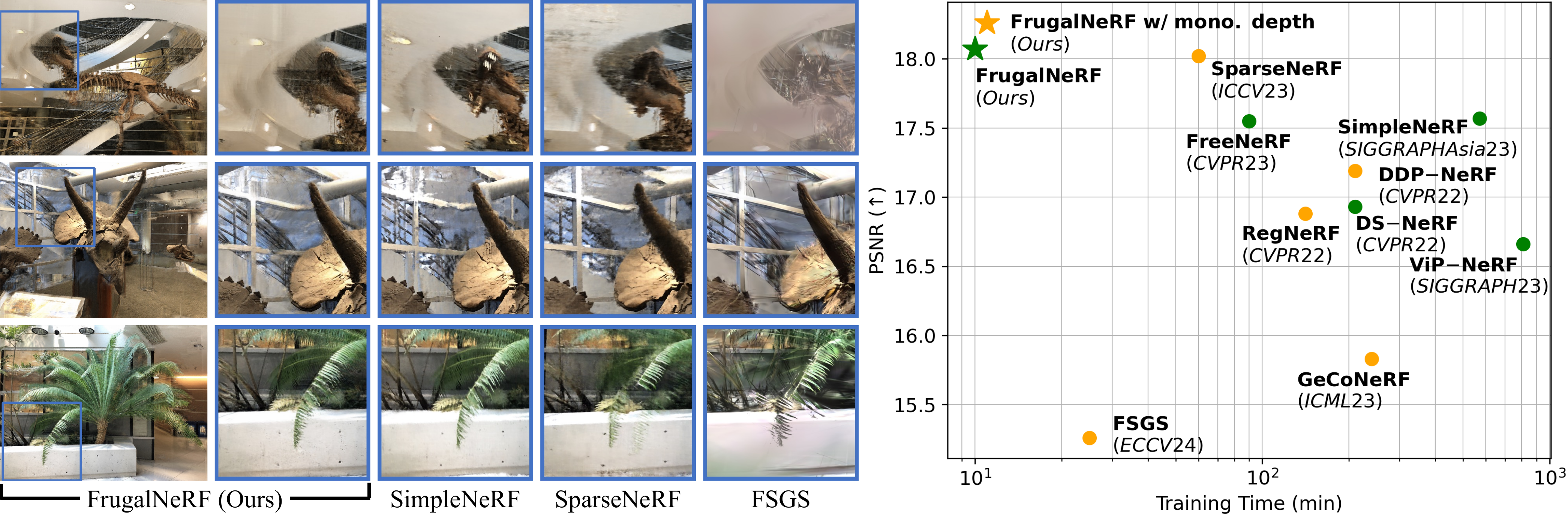}
}
\vspace{-4mm}
\caption{\textbf{Comparisons between FrugalNeRF and state-of-the-art methods with only two views for training.}
SimpleNeRF~\citep{somraj2023simplenerf} suffers from long training times, SparseNeRF~\citep{wang2023sparsenerf} produces blurry results, and FSGS~\citep{zhu2023FSGS} quality drops with few input views. Our FrugalNeRF achieves rapid, robust voxel training without learned priors, demonstrating superior efficiency and realistic synthesis. It can also integrate pre-trained priors for enhanced quality. \textbf{\textcolor{teasergreen}{Green}}: methods \emph{without} learned priors. \textbf{\textcolor{teaserorange}{Orange}}: \emph{with} learned priors.
}
\label{teaser}
\end{center}
}]


\maketitle

%



\begin{abstract}

\vspace{-3mm}
Neural Radiance Fields (NeRF) face significant challenges in extreme few-shot scenarios, primarily due to overfitting and long training times. Existing methods, such as FreeNeRF and SparseNeRF, use frequency regularization or pre-trained priors but struggle with complex scheduling and bias. We introduce FrugalNeRF, a novel few-shot NeRF framework that leverages weight-sharing voxels across multiple scales to efficiently represent scene details. Our key contribution is a cross-scale geometric adaptation scheme that selects pseudo ground truth depth based on reprojection errors across scales. This guides training without relying on externally learned priors, enabling full utilization of the training data. It can also integrate pre-trained priors, enhancing quality without slowing convergence.  Experiments on LLFF, DTU, and RealEstate-10K show that FrugalNeRF outperforms other few-shot NeRF methods while significantly reducing training time, making it a practical solution for efficient and accurate 3D scene reconstruction. 
\vspace{-3mm}
\end{abstract}

\section{Introduction}
\label{sec:intro}
Few-shot novel view synthesis, generating new views from limited imagery, is a substantial challenge in computer vision. While Neural Radiance Fields (NeRF) \citep{mildenhall2020nerf} have revolutionized high-fidelity 3D scene recreation, they demand considerable computational resources and time, often relying on external priors. This paper introduces \emph{FrugalNeRF}, a novel approach to accelerate NeRF training in extreme few-shot scenarios. It leverages the training data without relying on external priors and markedly reduces training overhead.

\begin{figure}[t]
\centering
\resizebox{1.0\columnwidth}{!} 
{
\begin{tabular}{cc}
\includegraphics[width=0.5\columnwidth]{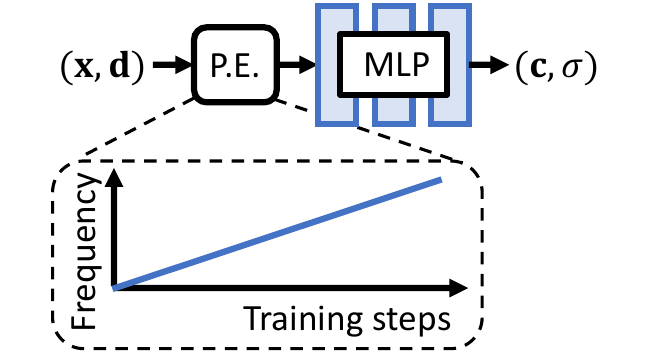} & 
\includegraphics[width=0.5\columnwidth]{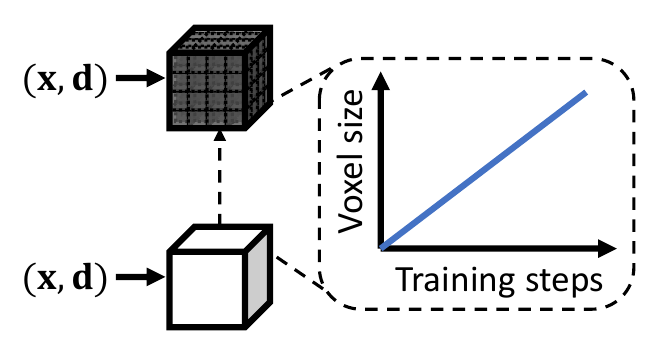}  \\
(a) Frequency regularization & (b) Voxel upsampling \\
  \\
\includegraphics[width=0.5\columnwidth]{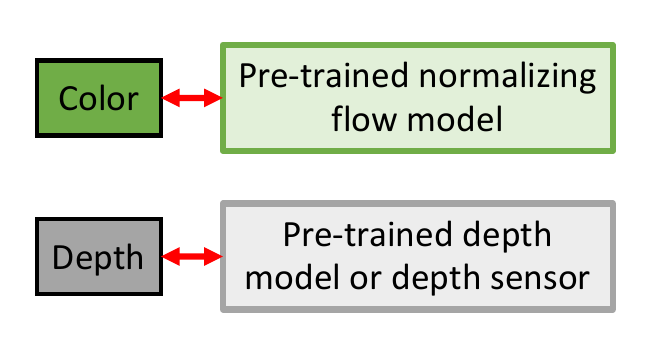} & 
\includegraphics[width=0.5\columnwidth]{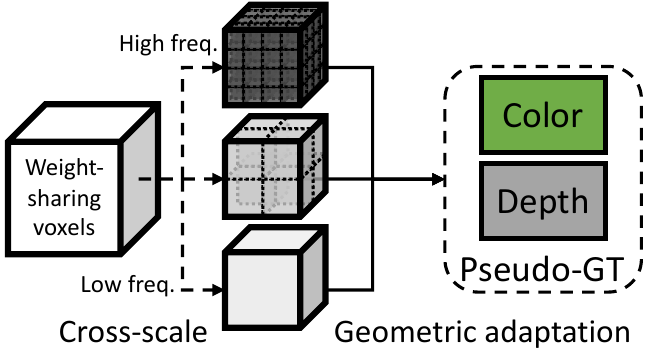}  \\
(c) Pre-trained models & (d) FrugalNeRF (Ours) \\
\end{tabular}%
}
\caption{\textbf{Comparisons between few-shot NeRF approaches.} 
(a) Frequency regularization gradually increases the visibility of high-frequency signals in positional encoding but slows training.
(b) Replacing MLPs with voxels and using gradual voxel upsampling offers similar regularization but lacks generalization.
(c) Other approaches use pre-trained models to supervise color or depth.
(d) Our FrugalNeRF uses weight-sharing voxels across scales for frequency representation, with cross-scale geometric adaptation for efficient supervision.
}
\label{fig:motivation}
\end{figure}

Traditional NeRF methods, despite producing high-quality outputs, suffer from long training time and rely on frequency regularization~\cite{yang2023freenerf} via multi-layer perceptrons (MLPs) and positional encoding, slowing convergence (\cref{fig:motivation}(a)). Alternatives like voxel upsampling (\cref{fig:motivation}(b)) attempt to overcome these challenges but struggle with generalizing to varied scenes~\cite{chen2022tensorf,sun2022direct,sun2023vgos}. Furthermore, using pre-trained models (\cref{fig:motivation}(c)) creates dependencies on external priors, which might not be readily available or could introduce biases from their training datasets~\cite{niemeyer2022regnerf,roessle2022dense,wang2023sparsenerf}.

FrugalNeRF differs from these approaches by incorporating a cross-scale, geometric adaptation mechanism, facilitating rapid training while preserving high-quality view synthesis (\cref{fig:motivation} (d)). Our method efficiently utilizes weight-sharing voxels across various scales to encapsulate the scene's frequency components. Our proposed adaptation scheme projects rendered depths and colors from different voxel scales onto the closest training view to compute reprojection errors. The most accurate scale becomes the pseudo-ground truth and guides the training across scales, thus eliminating the need for complex voxel upsampling schedules and enhancing generalizability across diverse scenes.

FrugalNeRF significantly reduces computational demands and accelerates training through self-adaptive mechanisms that exploit the multi-scale voxel structure, ensuring quick convergence without compromising the synthesis quality. By fully leveraging the training data and eliminating reliance on externally learned priors and their inherent limitations, FrugalNeRF provides a pathway toward more scalable and efficient few-shot novel view synthesis.
In conclusion, FrugalNeRF efficiently bypasses the need for external pre-trained prior and complex scheduling for voxel.

We evaluate the FrugalNeRF's effectiveness on three prominent datasets: LLFF~\citep{mildenhall2019local}, DTU~\citep{jensen2014large}, and RealEstate-10K~\cite{zhou2018stereo} dataset to assess both the rendering quality and convergence speed. Our results show that FrugalNeRF is not only faster but also achieves superior quality in comparison to existing methods (\cref{teaser}), showcasing FrugalNeRF's proficiency in generating perceptually high-quality images. The main contributions of our work are:
\begin{itemize}
\item We introduce a novel weight-sharing voxel representation that encodes multiple frequency components of the scene, significantly enhancing the efficiency and quality of few-shot novel view synthesis.
\item Our geometric adaptation selects accurate rendered depth across different scales by reprojection errors to create pseudo geometric ground truth that guides the training process, enabling a robust learning mechanism that is less reliant on complex scheduling and more adaptable to various scenes.
\item Our training scheme relies only on available data, eliminating the need for external priors or pre-trained models, and ensures fast convergence without losing quality. It also remains flexible, allowing learned priors to be added for better quality without slowing down training.
\end{itemize}

\section{Related Work}
\label{sec:related}
Neural Radiance Fields (NeRF)~\cite{mildenhall2020nerf} excels in synthesizing novel views of complex scenes~\cite{chen2022hallucinated,martin2021nerf,yuan2022nerf,xu2023neurallift,tao2023lidar,chen2023cunerf,peng2021pi,xu2022sinnerf,fridovich2022plenoxels,zhang2020nerf++,wang2021nerf,ye2022deformable,zheng2023editablenerf,bian2023nope}. In computer vision and 3D scene representation~\cite{hu2021worldsheet,sitzmann2019scene,dai2017scannet}, numerous research works focus on multi-view 3D view synthesis~\cite{oechsle2021unisurf,chen2021mvsnerf,jensen2014large,yariv2020multiview,wang2021ibrnet,su2024boostmvsnerfs,meuleman2023progressively}, single view synthesis~\cite{gao2020portrait,tucker2020single,han2022single,wiles2020synsin,wimbauer2023behind}, 3D image generation~\cite{chan2021pi,wang2022traditional,chan2022efficient,hong2022eva3d,li2021enforcing}, and dynamic 3D scene synthesis~\cite{pumarola2021d,mildenhall2022nerf,liu2023robust}. Few-shot Neural Radiance Fields (Few-shot NeRF)~\cite{chibane2021stereo,hu2023sherf,hu2023consistentnerf,chen2022geoaug} have gained interest in recent years, aiming to reconstruct 3D scenes from sparse input~\cite{zhang2021ners,jain2022zero,zhou2023sparsefusion,kim2022infonerf,bortolon2022vm,lee2023extremenerf,seo2023mixnerf,kwak2023geconerf}. However, they often face challenges such as overfitting to limited training images or poor generalization to novel viewpoints. To mitigate these issues, some approaches~\cite{yu2021pixelnerf,jain2021putting,wang2023sparsenerf,niemeyer2022regnerf} use pre-trained models, leveraging prior~\cite{johari2022geonerf,deng2023nerdi} knowledge to improve NeRF's ability in synthesizing unseen points or modeling a better geometry~\cite{chen2016single,uy2023scade} while others introduce additional regularization to improve performance ~\cite{yang2023freenerf,niemeyer2022regnerf,somraj2023vip,deng2022depth}. 


\vspace{3pt}  \noindent {\bf Depth regularizations.}
Recent works emphasize depth constraints during training. DS-NeRF~\citep{deng2022depth} uses sparse depth from an SfM model, focusing on sparse point regularization, while DDP-NeRF~\citep{roessle2022dense} extends DS-NeRF by completing depth priors from sparse points. SparseNeRF~\citep{wang2023sparsenerf} distills spatial continuity and depth ranking using a monocular depth estimator (MDE)\citep{Ranftl2021,Ranftl2020}. D{\"a}RF~\citep{song2023d} jointly optimizes NeRF and MDE to reduce MDE dataset bias. FSGS~\citep{zhu2023FSGS} employs MDE regularization from both seen and unseen views for improved adaptive density control. These methods rely on pre-trained MDE priors, which may contain errors from data bias and limited geometric detail. ViP-NeRF~\citep{somraj2023vip} uses visibility maps from plane sweep volume for regularization, but computing these priors is costly and lacks generalizability. In contrast, our FrugalNeRF regularizes geometry through geometrically adapted pseudo-GT depth, avoiding pre-trained models and extensive computation.

\vspace{3pt}  \noindent {\bf Novel pose regularization.}
Limited overlapping in sparse inputs often causes floaters in synthesized novel views. RegNeRF~\citep{niemeyer2022regnerf} uses view sampling via a normalizing flow model to regulate color rendering in unobserved viewpoints. PixelNeRF~\citep{yu2021pixelnerf} employs CNNs~\citep{NIPS2012_c399862d} to extract scene priors from input features, guiding unseen view rendering. DietNeRF~\citep{jain2021putting} uses a CLIP-based Vision Transformer~\citep{radford2021learning,caron2021emerging,li2022automated,lin2023vision} for color consistency constraints. FlipNeRF~\citep{seo2023flipnerf} samples reflection rays from estimated surface normals but relies on predefined reflection masks. Without ground truth for novel views, these constraints often depend on pre-trained models, adding inference time and potential bias. In contrast, our approach applies geometric adaptation to novel pose rendering, avoiding using pre-trained models while suppressing novel view floaters.

\vspace{3pt}  \noindent {\bf Frequency regularization.}
Positional encoding~\citep{sitzmann2020implicit,tancik2020fourier,wang2022hf} enables MLP-based NeRF to capture high-frequency details but risks overfitting in few-shot settings. FreeNeRF~\citep{yang2023freenerf} addresses this by gradually increasing input frequency. For voxel-based methods, gradually upsampling voxels aids radiance fields in avoiding overfitting, while VGOS~\citep{sun2023vgos} suppresses peripheral voxel optimization early to avoid overfitting, but both methods require complex scheduling and lack generalization. SimpleNeRF~\citep{somraj2023simplenerf} uses separate models for low- and high-frequency details, increasing resource and optimization costs. Our work leverages weight-sharing voxels across scales for various frequency representations, avoiding the need for complex scheduling.

\vspace{3pt}  \noindent {\bf Fast convergence.}
A key challenge in NeRF is slow training due to MLP queries. To address this, several methods~\citep{sun2023vgos,sun2022direct,chen2022tensorf,sitzmann2019deepvoxels,chen2024improving} replace most MLPs with faster-converging representations. Instant-NGP~\citep{muller2022instant} uses multiresolution hash encoding with density bitfields, while DVGO~\citep{sun2022direct} uses dense voxel grids with shallow MLPs. TensoRF~\citep{chen2022tensorf} improves voxel efficiency by decomposing radiance fields into low-rank tensors. ZeroRF~\citep{shi2023zerorf} adapts TensoRF for few-shot scenarios but focuses on object-level scenes. SparseCraft~\cite{younes2024sparsecraft} uses multiresolution hash encoding for 3D reconstruction but requires dense point clouds. DNGaussian~\citep{li2024dngaussian} employs compact Gaussian primitives but can create holes in novel view. Our FrugalNeRF builds on TensoRF for fast training and introduces a cross-scale, weight-sharing voxel framework for geometric adaptation.

\vspace{3pt}  \noindent {\bf Self-supervised consistency.}
Consistency modeling between sparse images and their warped counterparts is essential for Few-shot NeRFs. Traditional methods~\citep{darmon2022improving, chen2023structnerf, fu2022geo} warp images to minimize reprojection errors but struggle with limited data. SinNeRF~\citep{xu2022sinnerf} and PANeRF~\citep{ahn2022panerf} use warped views as pseudo labels to enforce geometric consistency but require RGB-D inputs. SE-NeRF~\citep{jung2023self} and Self-NeRF~\citep{bai2023self} leverage a teacher NeRF's outputs as labels. GeCoNeRF~\citep{kwak2023geconerf} uses rendered depth for warping but depends on a pre-trained feature extractor, slowing training. ReVoRF~\citep{xu2024learning} enhances geometric accuracy via bilateral consistency but smooths details. Our FrugalNeRF combines frequency regularization with cross-scale geometric adaptation, using the best render depth at different scales as a pseudo label to ensure geometric consistency without relying on learned priors.

\section{Method}
\label{sec:method}


\begin{figure*}[t]
\centering
\includegraphics[width=\textwidth]{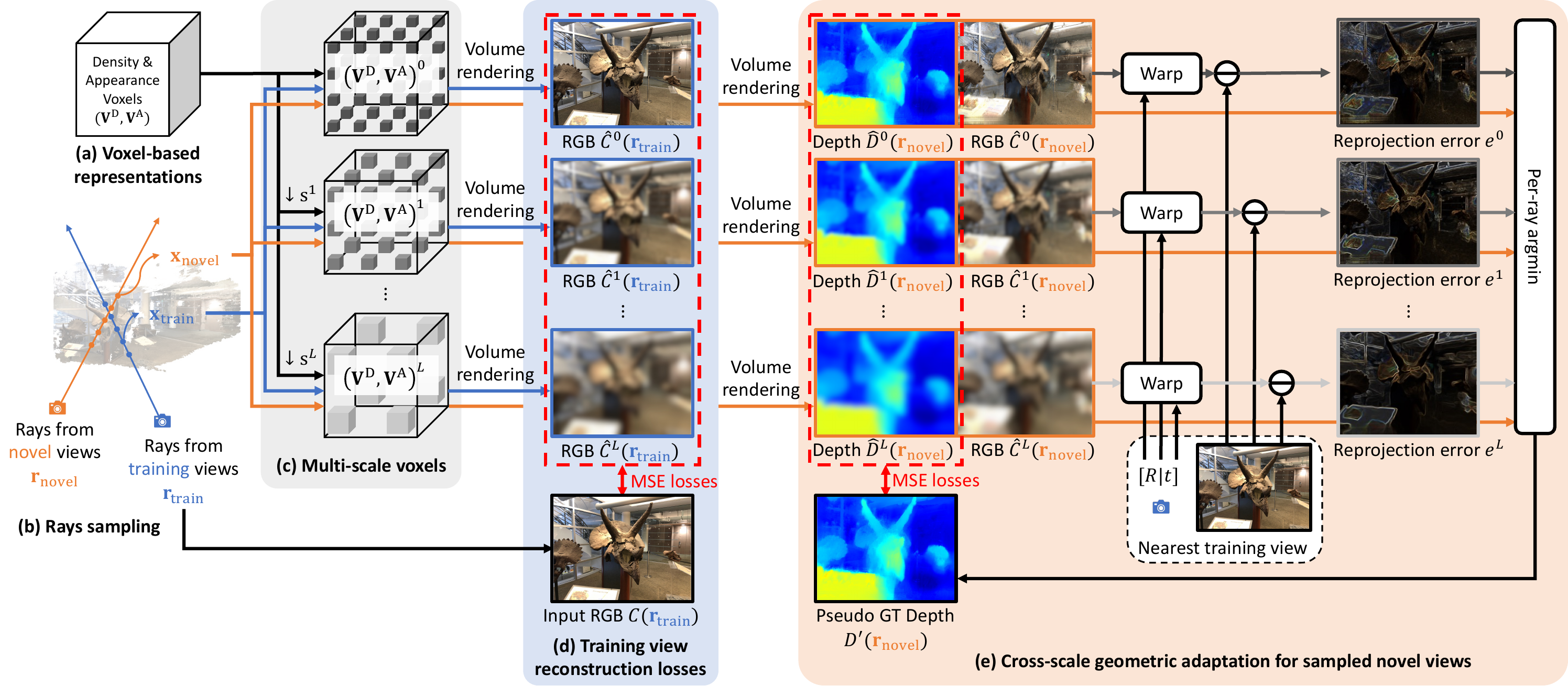}
\caption{\textbf{Overview of FrugalNeRF architecture.}
(a) Our FrugalNeRF represents a scene with a pair of density and appearance voxels $(\bf{V}^\text{D}, \bf{V}^\text{A})$. For a better graphical illustration, we show only one voxel in the figure.
(b) We sample rays from not only training input views $\bf{r}_{\text{train}}$ but also randomly sampled novel views $\bf{r}_{\text{novel}}$.
(c) We then create $L+1$ multi-scale voxels by hierarchical subsampling, where lower-resolution voxels ensure global geometry consistency and reduce overfitting but suffer from representing detailed structures, while higher-resolution voxels capture fine details but may get stuck in the local minimum or generate floaters.
(d) For the rays from training views $\bf{r}_\text{train}$, we enforce an MSE reconstruction loss between the volume rendered RGB color $\hat{C}^l$ and input RGB $C$ at each scale.
(e) We introduce a cross-scale geometric adaptation loss for novel view rays $\bf{r}_{\text{novel}}$, warping volume-rendered RGB to the nearest training view using predicted depth, calculating projection errors $e^l$ at each scale, and using the depth with the minimum reprojection error as pseudo-GT for depth supervision. This adaptation involves rays from both training and novel views, though the figure only depicts novel view rays for clarity.
}
\label{fig:framework}
\end{figure*}

\subsection{Preliminaries}

\noindent {\bf Neural radiance fields.}
NeRF~\cite{mildenhall2020nerf} uses a neural network $f$ to map 3D location $\mathbf{x}$ and viewing direction\ $\mathbf{d}$ to density $\sigma$ and color $\mathbf{c}$ for image rendering: $f: (\mathbf{x}, \mathbf{d}) \rightarrow (\sigma, \mathbf{c})$.
Then we use the densities and colors to render a pixel color $\hat{C}(\mathbf{r})$ by integrating the contributions along a ray $\mathbf{r}$ cast through the scene: $\hat{C}(\mathbf{r}) = \sum^{N}_{i=1} T_i (1-\text{exp}(-\sigma_i \delta_i)) \mathbf{c}_i$, where $T(t)=\text{exp}(-\sum_{j=i}^{i-1} \sigma_j \delta_j)$ is the transmittance along the ray, and $N$ is the number of points along the ray.
NeRF seeks to minimize the MSE between the rendered image and the actual image: $\mathcal{L} = \sum_{\bf{r} \in \mathcal{R}} \left \| \hat{C}(\mathbf{r}) - C(\mathbf{r}) \right \|^2$, where $\mathcal{R}$ denotes a set of rays.

\vspace{3pt}  \noindent {\bf Voxel-based NeRFs.}
Voxel-based NeRFs~\cite{sun2022direct,chen2022tensorf,muller2022instant} accelerate color and density querying by using voxel grids with trilinear interpolation for efficient data retrieval. They commonly use a coarse-to-fine strategy and refine view-dependent effects with a shallow MLP.

\vspace{3pt}  \noindent {\bf Few-shot NeRFs.}
Recent methods address under-constrained optimization with limited images by regularizing visible frequencies in positional encoding~\cite{yang2023freenerf} (\cref{fig:motivation}(a)), incrementally expanding voxel ranges~\cite{sun2023vgos} (\cref{fig:motivation}(b)), and using external priors from pre-trained models~\cite{wang2023sparsenerf} (\cref{fig:motivation}(c)). Our FrugalNeRF leverages a weight-sharing voxel across scales to capture a range of frequency, self-adapting to learn optimal geometric frequencies for faster training without pre-trained models (\cref{fig:motivation}(d)).


\subsection{Overview of FrugalNeRF}

FrugalNeRF introduces an efficient architecture for novel view synthesis from sparse inputs, eliminating the need for external priors. This novel approach leverages voxel-based NeRFs~\cite{chen2022tensorf,muller2022instant,sun2022direct} to estimate 3D geometry and shorten training time with limited 2D images. The key feature is hierarchical subsampling with weight-sharing multi-scale voxels, capturing diverse geometric details (\cref{sec:voxel}). To prevent overfitting in extreme few-shot scenarios, we apply geometric adaptation for regularization (\cref{sec:self}), along with novel view sampling and additional regularization losses to reduce artifacts (\cref{sec:novel}).  FrugalNeRF integrates data from both training and sampled views for robust and accurate scene representation (\cref{sec:loss}).

%
\subsection{Weight-Sharing Multi-Scale Voxels}
\label{sec:voxel}

To tackle data sparsity in few-shot scenarios, FrugalNeRF uses weight-sharing multi-scale voxels to balance frequency characteristics. Inspired by FreeNeRF~\cite{yang2023freenerf}, which addresses overfitting from high-frequency inputs, we use a voxel-based representation to manage frequency. Lower-resolution voxels capture broad scene outlines, while higher resolutions model finer details, similar to NeRF's positional encoding~\cite{mildenhall2020nerf}.

Unlike methods such as VGOS~\cite{sun2023vgos}, which starts with a coarse geometry and progressively refines details, our approach maintains generalization without intricate tuning. 
We construct multi-scale voxels by downsampling from a single density and appearance voxel, ensuring consistent scene representation(\cref{fig:framework} (c)).
This technique effectively balances different frequency bands in the training pipeline without increasing model size or memory demands.

With multi-scale voxels, we can further utilize \emph{multi-scale voxel color loss} to guide the training (\cref{fig:framework}(d)), which is crucial for few-shot scenarios in ensuring a balanced representation of geometry and detail.  The multi-scale voxel color loss is defined as:
\begin{equation}
    \mathcal{L}_\text{ms-color} = \sum_{l=0}^{L} \sum_{\bf{r}_\text{train} \in \mathcal{R}_\text{train}} {\left \| \hat{C}^l(\mathbf{r}_\text{train}) - C(\mathbf{r}_\text{train}) \right \|^2},
\end{equation}
where $\hat{C}^l$ is the rendered color from the voxel at scale $l$, $C$ is the ground truth color, $L$ is the number of scales, $\mathcal{R}_\text{train}$ is a set of rays from training views, and $\bf{r}_\text{train}$ is a ray sampled from  $\mathcal{R}_\text{train}$.
We compute a weighted average MSE loss across scales to ensure color rendering accuracy at each scale, enhancing overall robustness and fidelity.

\subsection{Cross-scale geometric adaptation}
\label{sec:self}

Our \emph{cross-scale geometric adaptation} approach effectively addresses the challenges of extreme few-shot scenarios by supervising geometry without ground truth depth data. Recognizing the diverse frequency representation by different voxel scales in a scene, it is essential to identify the optimal frequency band for each region of the scene.

For each ray from a training view $i$, we compute depth values at multiple scales through volume rendering and then warp~\cite{luo2020consistent,kopf2021robust,li2021enforcing} view $i$'s input RGB to the nearest training view $j$ using these depths. The reprojection error with view $j$'s input RGB determines the most suitable scale for each scene area. The depth of this scale serves as a pseudo-ground truth, guiding the model in maintaining geometric accuracy across frequencies (\cref{fig:framework}(e)).



Mathematically, for a pixel ${\bf{p}}_{i}$ in a training frame $i$, with its depth $D_{i}^{l}({\bf{p}}_{i})$ at scale $l$ and camera intrinsic $K_{i}$, we can lift ${\bf{p}}_{i}$ to a 3D point ${\bf{x}}_{i}^{l}$, then transform it to world coordinate ${\bf{x}}^{l}$, and subsequently transform to frame $j$'s camera coordinate ${\bf{x}}_{i \rightarrow j}^{l}$. This 3D point is then projected back to 2D in frame $j$, obtaining the pixel coordinate ${\bf{p}}_{i \rightarrow j}^{l}$. Due to the space limit, we provide the details for reprojection calculation in the supplementary. We calculate the reproject error $e^l({\bf{p}}_{i})$ using the RGB values of frame $i$ and $j$ for each scale $l$. 
\begin{equation}
\label{eq:reprojection_error_train}
    e^l({\bf{p}}_{i}) = \left \| C_{i}({\bf{p}}_{i}) - C_{j}({{\bf{p}}}_{i\rightarrow j}^{l})\right \|^2,
\end{equation}
where $C_{i}$ and $C_{j}$ are the input RGB images from view $i$ and $j$, respectively. For a pixel location $\bf{p}$ from which the training view ray $\mathbf{r}_\text{train}$ originates, we denote it simply as $\mathbf{r}_\text{train}$. The pseudo-ground truth depth for this pixel is the depth at the scale with the minimum reprojection error:
\begin{equation}
\label{eq:level_train}
    l'(\mathbf{r}_\text{train})=\arg \min _{l}(e^{l}(\mathbf{r}_\text{train})).
\end{equation}
\begin{equation}
\label{eq:depth_train}
    D'(\mathbf{r}_\text{train}) = \hat{D}^{l'(\mathbf{r}_\text{train})} (\mathbf{r}_\text{train}),
\end{equation}
where $\hat{D}^l$ is the rendered depth from the voxel at scale $l$, and $l'$ denotes the scale with minimum reprojection error:
This pseudo-ground truth depth $D'$ is used to compute a geometric adaptation loss, $\mathcal{L}_\text{geo}(\mathbf{r}_\text{train})$, an MSE loss that ensures the model maintains scene geometry effectively, even without explicit depth ground truth:
\begin{equation}
\label{eq:loss_train}
    \mathcal{L}_\text{geo}(\mathbf{r}_\text{train}) = \sum_{l=0}^{L} \sum_{\bf{r}_\text{train} \in \mathcal{R}_\text{train}} {\left \| \hat{D}^l(\mathbf{r}_\text{train}) - D'(\mathbf{r}_\text{train}) \right \|^2}.
\end{equation}

We further define a threshold for reprojection error to determine the reliability of depth estimation. Specifically, we do not compute the loss of those pixels in which the projection error exceeds this pre-defined threshold.
Geometric adaptation is critical by allowing the model to refine its understanding of the scene's geometry in a self-adaptive manner.

\subsection{Novel View Regularizations}
\label{sec:novel}
In few-shot scenarios, we extend geometric adaptation to \emph{novel views} to address the limitations in areas with less overlap among training views (\cref{fig:framework}(e)). Our novel view sampling strategy involves a spiral trajectory around training views, promoting comprehensive coverage and model robustness. In the absence of ground truth RGB for novel views, we rely on rendered color $\hat{C}$ for reprojection error calculation, similar to~\cref{eq:reprojection_error_train} in~\cref{sec:self}, but focusing on rays from novel views $\textbf{r}_\text{novel}$:
\begin{equation}
    e^l({\bf{p}}_{n}) = \left \| \hat{C}_{n}({\bf{p}}_{n}) - C_{j}({{\bf{p}}}_{n\rightarrow j}^{l})\right \|^2.
\end{equation}
In this context, ${\bf{p}}_{n}$ denotes a pixel coordinate in the sampled novel frame $n$, and ${{\bf{p}}}_{n\rightarrow j}^{l}$ represents the coordinates on its nearest training pose $j$ after warping ${{\bf{p}}}_{n}$ at scale $l$. This reprojection error helps refine the model's rendering for novel views. For each ray from a novel view, similar to~\cref{eq:depth_train,eq:level_train,eq:loss_train}, we first determine the scale with the minimum reprojection error, then determine its pseudo-ground truth depth and calculate geometric adaptation loss:
\begin{equation}
    l'(\mathbf{r}_\text{novel}) = \arg \min _{l}(e^{l}(\mathbf{r}_\text{novel})),
\end{equation}
\begin{equation}
    D'(\mathbf{r}_\text{novel}) = \hat{D}^{l'(\mathbf{r}_\text{novel})} (\mathbf{r}_\text{novel}),
\end{equation}
\begin{equation}
    \mathcal{L}_\text{geo}(\mathbf{r}_\text{novel}) = \sum_{l=0}^{L} \sum_{\bf{r}_\text{novel} \in \mathcal{R}_\text{novel}} {\left \| \hat{D}^l(\mathbf{r}_\text{novel}) - D'(\mathbf{r}_\text{novel}) \right \|^2},
\end{equation}
where $\mathcal{R}_\text{novel}$ is the set of rays from sampled novel views, and $\bf{r}_\text{novel}$ is a sampled ray from the set $\mathcal{R}_\text{novel}$.

We combine this loss with the geometric adaptation loss from training views to enhance the overall training process:
\begin{equation}
    \mathcal{L}_\text{geo} = \mathcal{L}_\text{geo}(\mathbf{r}_\text{train}) + \mathcal{L}_\text{geo}(\mathbf{r}_\text{novel}).
\end{equation}
This approach of novel view sampling and applying regularization through reprojection error computation is critical in training our model. It ensures that the model not only learns from the limited training views but also adapts to and accurately renders novel perspectives, thereby enhancing the overall performance and reliability of FrugalNeRF.

\vspace{3pt}  \noindent {\bf Additional global regularization losses.}
To further improve the geometry and reduce artifacts, we introduce an additional global regularization loss $\mathcal{L}_\text{reg}$, including total variation loss~\citep{chen2022tensorf, sun2023vgos}, patch-wise depth smoothness loss~\citep{niemeyer2022regnerf}, L1 sparsity loss~\citep{chen2022tensorf}, and distortion loss~\citep{sun2022direct, barron2022mip}. These losses help smooth the scene globally and suppress artifacts like floaters and background collapse.

\subsection{Total Loss}
\label{sec:loss}
The total loss for FrugalNeRF, essential for accurate scene rendering from sparse views, combines various components: color fidelity, geometric adaptation, global regularization, and sparse depth constraints. It is formulated as:
\begin{equation}
    \mathcal{L} = \mathcal{L}_\text{ms-color} + \lambda _\text{geo}\mathcal{L}_\text{geo} + \lambda _\text{reg}\mathcal{L}_\text{reg} + \lambda _\text{sd}\mathcal{L}_\text{sd} + \lambda _\text{d}\mathcal{L}_\text{d}.
\end{equation}
$\mathcal{L}_\text{ms-color}$ is the multi-scale voxel color loss, crucial for maintaining color accuracy across different scales.
$\mathcal{L}_\text{geo}$ is the geometric adaptation loss, providing geometric guidance in the absence of explicit depth information.
$\mathcal{L}_\text{reg}$ is the global regularization loss, addressing artifacts and inconsistencies in unseen areas.
And $\mathcal{L}_\text{sd}$ is the sparse depth loss~\cite{deng2022depth}, utilizing sparse depth data for absolute scale constraints derived from COLMAP~\cite{schoenberger2016sfm,schoenberger2016mvs}. We \emph{optionally} incorporate the depth loss, $\mathcal{L}_\text{d}$, using the Dense Prediction Transformer (DPT)~\citep{ranftl2021vision} to generate depth maps from the training views.

\section{Experiments}
\label{sec:experiments}

\begin{table*}[t]
\centering
\caption{\textbf{Quantitative results on the LLFF~\cite{mildenhall2019local} dataset.} 
FrugalNeRF performs competitively with baseline methods in extreme few-shot settings, offering shorter training time without relying on externally learned priors. Integrating monocular depth regularization further improves quality while maintaining fast convergence. 
}
\label{tab:quantitative_llff}
\resizebox{\textwidth}{!}{%
\begin{tabular}{l|c|c|ccc|ccc|ccc|c}
\toprule
 & & Learned & \multicolumn{3}{c|}{2-view} & \multicolumn{3}{c|}{3-view} & \multicolumn{3}{c|}{4-view} & Training \\
Method & Venue & priors & PSNR $\uparrow$ & SSIM $\uparrow$ & LPIPS $\downarrow$ & PSNR $\uparrow$ & SSIM $\uparrow$ & LPIPS $\downarrow$ & PSNR $\uparrow$ & SSIM $\uparrow$ & LPIPS $\downarrow$ & time $\downarrow$\\
\midrule
TensoRF~\citep{chen2022tensorf} & ECCV22 & - & 11.97 & 0.32 & 0.64 & 12.63 & 0.32 & 0.63 & 13.32 & 0.35 & 0.60 & 6 mins \\
\midrule
DS-NeRF~\citep{deng2022depth} & CVPR22 & - & 16.93 & 0.51 & 0.42 & 18.97 & 0.58 & 0.36 & 20.07 & 0.61 & 0.34 & 3.5 hrs \\
FreeNeRF~\citep{yang2023freenerf} & CVPR23 & - & 17.55 & \cellcolor{orange!25}0.54 & 0.38 & 19.30 & 0.60 & 0.34 & \cellcolor{orange!25}20.45 & 0.63 & 0.33 & \cellcolor{orange!25}1.5 hrs\\
ViP-NeRF~\citep{somraj2023vip} & SIGGRAPH23 & - & 16.66 & 0.52 & \cellcolor{orange!25}0.37 & 18.89 & 0.59 & 0.34 & 19.34 & 0.62 & 0.32 & 13.5 hrs \\
SimpleNeRF~\citep{somraj2023simplenerf} & SIGGRAPH Asia23 & - & \cellcolor{orange!25}17.57 & \cellcolor{red!25}0.55 & 0.39 & \cellcolor{orange!25}19.47 & \cellcolor{red!25}0.62 & \cellcolor{orange!25}0.33 & 20.44 & \cellcolor{red!25}0.65 & \cellcolor{orange!25}0.31&9.5 hrs \\
FrugalNeRF (Ours) & - & - & \cellcolor{red!25}18.07 & \cellcolor{orange!25}0.54 & \cellcolor{red!25}0.35 & \cellcolor{red!25}19.66 & \cellcolor{orange!25}0.61 & \cellcolor{red!25}0.30 & \cellcolor{red!25}20.70 & \cellcolor{red!25}0.65 & \cellcolor{red!25}0.28 & \cellcolor{red!25}10 mins\\
\midrule
RegNeRF~\citep{niemeyer2022regnerf} & CVPR22 & normalizing flow & 16.88 & 0.49 & 0.43 & 18.65 & 0.57 & 0.36 & 19.89 & 0.62 & 0.32 & 2.35 hrs\\
DDP-NeRF~\citep{roessle2022dense} & CVPR22 & depth completion & 17.19 & \cellcolor{orange!25}0.54 & \cellcolor{orange!25}0.39 & 17.71 & 0.56 & 0.39 & 19.19 & 0.61 & 0.35 & 3.5 hrs \\
GeCoNeRF~\citep{kwak2023geconerf} & ICML23 & VGG19 feature &  15.83 & 0.45 & 0.52 & 17.44 & 0.50 & 0.47 &  19.14 &  0.56 & 0.42 & 4 hrs \\
SparseNeRF~\citep{wang2023sparsenerf} & ICCV23 & monocular depth  &  \cellcolor{orange!25}18.02 & 0.52 & 0.45 & \cellcolor{orange!25}19.52 & \cellcolor{orange!25}0.59 & \cellcolor{orange!25}0.37 & \cellcolor{red!25}20.89 & \cellcolor{orange!25}0.65 & 0.34 & 1 hrs \\
FSGS~\citep{zhu2023FSGS} & ECCV24 & monocular depth  &  15.26 & 0.45 & 0.41 & 19.21 & \cellcolor{red!25}0.61 & \cellcolor{red!25}0.30 & \cellcolor{orange!25}20.07 & \cellcolor{red!25}0.66 & \cellcolor{red!25}0.22 & \cellcolor{orange!25}25 mins \\
FrugalNeRF (Ours) & - & monocular depth & \cellcolor{red!25}18.26 & \cellcolor{red!25}0.55 & \cellcolor{red!25}0.35 & \cellcolor{red!25}19.87 & \cellcolor{red!25}0.61 & \cellcolor{red!25}0.30 & \cellcolor{red!25}20.89 & \cellcolor{red!25}0.66 & \cellcolor{orange!25}0.26 & \cellcolor{red!25}11 mins\\
 \bottomrule
\end{tabular}%
}
\end{table*}


\begin{figure*}[t]
\centering
\resizebox{1.0\textwidth}{!} 
{
\includegraphics[width=\textwidth]{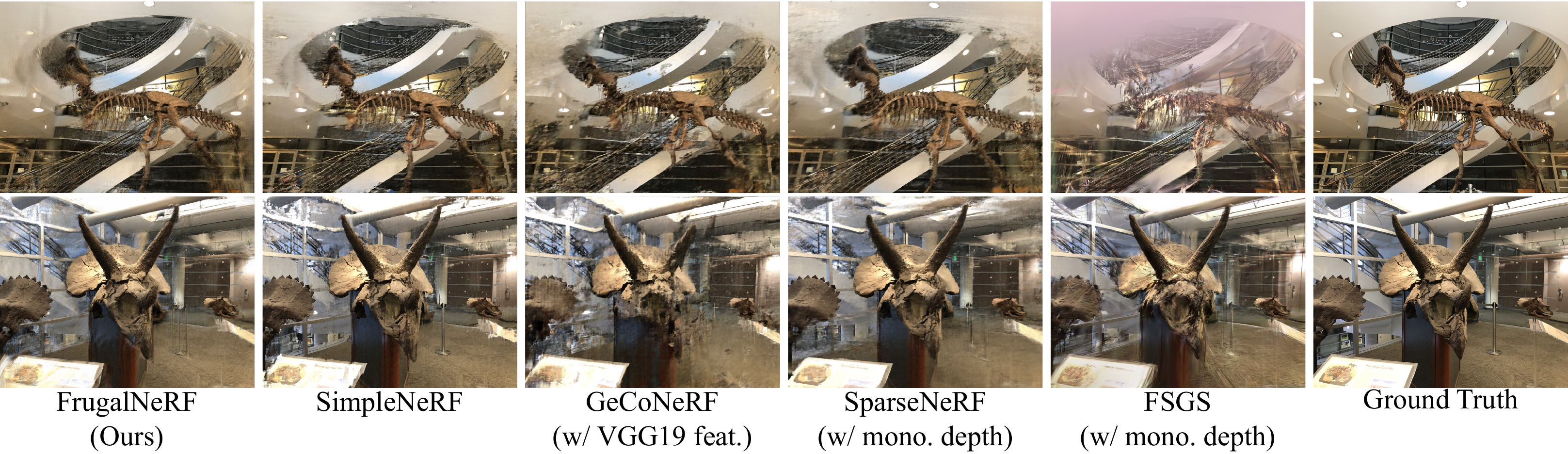}
}
\caption{\textbf{Qualitative comparisons on the LLFF~\cite{mildenhall2019local} dataset with two input views.} FrugalNeRF achieves better synthesis quality.}
\label{fig:llff_visual}
\end{figure*}

\noindent {\bf Datasets \& evaluation metrics.}
We conduct experiments on three datasets: LLFF~\citep{mildenhall2019local}, DTU~\citep{jensen2014large}, and RealEstate-10K~\cite{zhou2018stereo}. For those datasets, we use the test sets defined by pixelNeRF~\citep{yu2021pixelnerf} and ViP-NeRF~\citep{somraj2023vip}. We follow the same evaluation protocol as ViP-NeRF evaluate in the more challenging setup of 2, 3, or 4 input views, unlike prior work which uses 9–18 input views~\cite{niemeyer2022regnerf, yang2023freenerf}.
Specifically, there are 12 scenes\footnote{There are 15 scenes in total in ViP-NeRF's DTU test sets. 
However, COLMAP can only run successfully on 12 scenes.} in the test sets of the DTU dataset.
We assume that camera parameters are known, which is relevant for applications with available calibrated cameras. We provide further details in the supplementary materials.


We follow standard evaluation protocols, using PSNR, SSIM~\cite{wang2004image}, and LPIPS~\cite{zhang2018perceptual}. For DTU, we follow SparseNeRF~\cite{yang2023freenerf} in removing background to mitigate bias as noted by RegNeRF~\cite{niemeyer2022regnerf} and pixelNeRF~\cite{yu2021pixelnerf}. We also report training time on a single NVIDIA RTX 4090 GPU to fairly evaluate the efficiency of the methods.

\vspace{3pt}  \noindent {\bf Implementation details.}
We implement FrugalNeRF based on TensoRF~\citep{chen2022tensorf} using the official PyTorch framework. Training is optimized with the Adam optimizer~\citep{kingma2014adam} at an initial learning rate of 0.08, which decays to 0.002 over the course of training. To compute the $\mathcal{L}_\text{sd}$ loss, we utilize COLMAP to generate a sparse point cloud from the few-shot training views. For reprojection error, we use a $5 \times 5$ pixel patch with a 0.5 threshold to reliably identify robust pseudo-GT depth values. The batch size is set to 4,096 for both training and novel view rays, with 120 novel poses sampled along a spiral trajectory around the training view. Each scene is trained for 5,000 iterations. Specific voxel resolutions are applied to different datasets: $640^3$ for LLFF and RealEstate-10K, and $300^3$ for DTU, with a voxel downsample ratio of $s=4$ and $L=2$ (three total scale levels) to capture varying scene details. We set the loss weights as follows: $\lambda _\text{geo} = 0.01$, $\lambda _\text{sd} = 0.5$, and $\lambda _\text{d} = 0.01$, while $\lambda _\text{reg}$ combines multiple losses to enhance training stability and performance. Further details are provided in the supplementary materials.
\begin{figure*}[t]
\centering
\resizebox{1.0\textwidth}{!} 
{
\includegraphics[width=\textwidth]{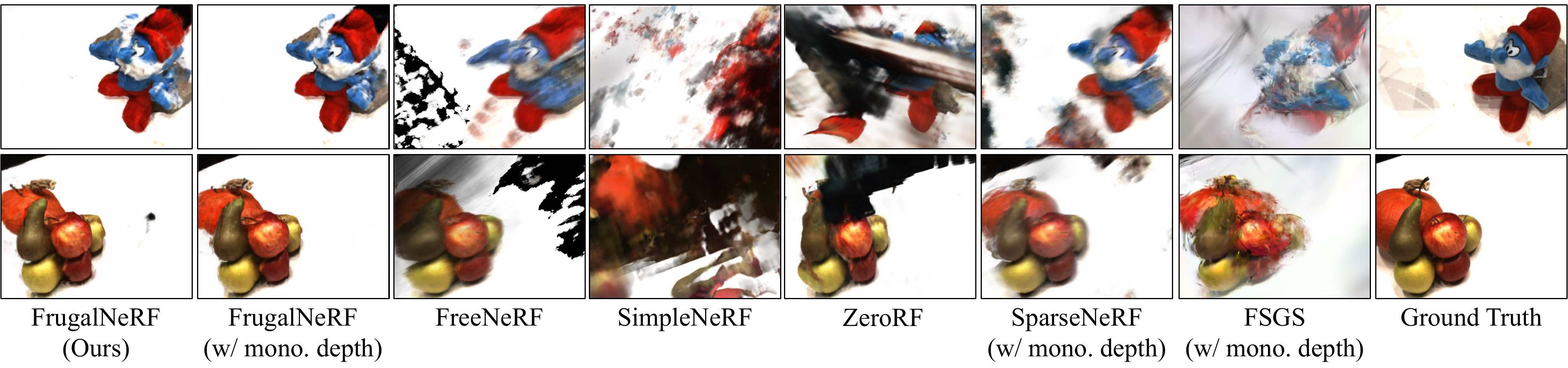}
}
\vspace{-6mm}
\caption{\textbf{Qualitative comparisons on the DTU~\cite{jensen2014large} dataset with two input views.} 
FrugalNeRF achieves better synthesis quality.
}
\label{fig:dtu_visual}
\end{figure*}

\begin{table}[t]
\centering
\caption{\textbf{Quantitative results on the DTU ~\cite{jensen2014large} dataset.} 
FrugalNeRF outperforms most baselines in extreme few-shot settings with shorter training time and no reliance on external priors. Integrating monocular depth regularization further enhances quality while preserving fast convergence. 
}
\label{tab:quantitative_dtu_2v}
\vspace{-3mm}
\resizebox{\columnwidth}{!}{%
\begin{tabular}{l|c|ccc|c}
\toprule
 & Learned & \multicolumn{3}{c|}{2-view} & Training \\
Method & priors & PSNR $\uparrow$ & SSIM $\uparrow$ & LPIPS $\downarrow$ & time $\downarrow$\\
\midrule
TensoRF~\cite{chen2022tensorf} & - & 8.81 & 0.34 & 0.71 & 5 mins \\
\midrule
FreeNeRF~\cite{yang2023freenerf} & - & \cellcolor{orange!25}18.05 & \cellcolor{orange!25}0.73 & \cellcolor{orange!25}0.22 & 1 hrs \\
ViP-NeRF~\cite{somraj2023vip} & - & 14.91 & 0.49 & 0.24 & 2.2 hrs \\
SimpleNeRF~\cite{somraj2023simplenerf} & - & 14.41 & 0.79 & 0.25 & 1.38 hrs \\
ZeroRF~\cite{shi2023zerorf} & - & 
14.84 & 0.60 & 0.30 & 25 mins
\\ 
FrugalNeRF (Ours) & - & \cellcolor{red!25}19.72 & \cellcolor{red!25}0.78 & \cellcolor{red!25}0.16 & \cellcolor{red!25}6 mins \\
\midrule
SparseNeRF~\cite{wang2023sparsenerf} & monocular depth & \cellcolor{orange!25}19.83 & \cellcolor{orange!25}0.75 & \cellcolor{orange!25}0.20 & 30 mins \\
FSGS~\cite{zhu2023FSGS} & monocular depth & 16.82 & 0.64 & 0.27 & 20 mins \\
FrugalNeRF (Ours) & monocular depth & \cellcolor{red!25}20.77 & \cellcolor{red!25}0.79 & \cellcolor{red!25}0.15  & \cellcolor{red!25}7 mins\\
\bottomrule
\end{tabular}%
}
\end{table}

\subsection{Comparisons} 
\noindent {\bf LLFF dataset.}
We compare FrugalNeRF to RegNeRF~\citep{niemeyer2022regnerf}, DS-NeRF~\citep{deng2022depth}, DDP-NeRF~\citep{roessle2022dense}, FreeNeRF~\citep{yang2023freenerf}, ViP-NeRF~\citep{somraj2023vip}, SimpleNeRF~\citep{somraj2023simplenerf}, GeCoNeRF~\citep{kwak2023geconerf}, SparseNeRF~\citep{wang2023sparsenerf}, and FSGS~\citep{zhu2023FSGS}. As shown in \cref{tab:quantitative_llff}, whether learned priors are used, FrugalNeRF outperforms other methods in PSNR and LPIPS, with comparable SSIM. FrugalNeRF achieves an ideal balance of quality and efficiency, completing training in just 10 minutes. Our cross-scale geometric adaptation generalizes better than frequency regularization methods like FreeNeRF, and adding monocular depth regularization further enhances quality without slowing convergence. Qualitative comparisons (\cref{fig:llff_visual}) show FrugalNeRF renders scenes with richer details and sharper edges compared to SparseNeRF’s blurry results. Unlike SimpleNeRF and FSGS, which show floaters and holes, FrugalNeRF models geometry smoothly and consistently, demonstrating its capability for high-fidelity scene modeling.

\vspace{3pt}  \noindent {\bf DTU dataset.}
We compare FrugalNeRF with  FreeNeRF~\citep{yang2023freenerf}, ViP-NeRF~\citep{somraj2023vip}, SimpleNeRF~\citep{somraj2023simplenerf}, SparseNeRF~\citep{wang2023sparsenerf}, ZeroRF~\citep{shi2023zerorf}, and FSGS~\citep{zhu2023FSGS} on the dataset preprocessed by pixelNeRF~\citep{yu2021pixelnerf}. \cref{tab:quantitative_dtu_2v} shows FrugalNeRF achieves state-of-the-art performance in most cases, with the shortest training time.
Qualitative comparisons (\cref{fig:dtu_visual}) demonstrate FrugalNeRF's superior visual results, consistently rendering fine details (e.g., the blue elf's eyes) without noticeable artifacts, unlike other methods. 
\subsection{Ablation Studies}


\noindent {\bf Number of scales.}
We examine the effect of scale numbers in \cref{tab:ablation_voxel_amount}. Results show that increasing scales improves rendering quality, as multiple voxel resolutions enable FrugalNeRF to represent scene details more effectively through geometric adaptation. We use $L=2$ (three total scales) in our experiments to balance quality and training time.

\vspace{3pt}  \noindent {\bf Weight-sharing voxels.}
We compared the performance and memory usage of weight-sharing voxels against three independent voxels.
\cref{tab:abalation_with_losses} indicates that weight-sharing not only enhances performance but also reduces the model size. 

\begin{table}[t]
    \centering
    \small
    \setlength{\tabcolsep}{8pt}
    \renewcommand{\arraystretch}{0.8}
    \caption{\textbf{Comparison of different number of scales on LLFF.}}
    \label{tab:ablation_voxel_amount}
\vspace{-3mm}
    \begin{tabular}{l|cccc}
    \toprule
    \# of scales & PSNR $\uparrow$ & SSIM $\uparrow$ & LPIPS $\downarrow$ &  Time $\downarrow$\\
    \midrule
    1 ($L=0$) & 15.22 & 0.46 & 0.43 & \bf{6 mins} \\
    2 ($L=1$) & 16.58 & 0.53 & 0.37 & 7 mins \\
    3 ($L=2$) & 18.07 & \bf{0.54} & \bf{0.35} & 10 mins \\
    4 ($L=3$) & \bf{18.08} & \bf{0.54} & 0.36 & 15 mins \\
    \bottomrule
    \end{tabular}%
\end{table}

\begin{table}[t]
    \centering
    \small
    \setlength{\tabcolsep}{2pt}
    \renewcommand{\arraystretch}{0.8}
    \caption{\textbf{Ablation of different components on the LLFF dataset with two input views.}}
    \label{tab:abalation_with_losses}
    \resizebox{1.0\columnwidth}{!}{%
    \begin{tabular}{cccc|ccccc}
    \toprule
    Weight-sharing & $\mathcal{L}_\text{ms-color}$ & $\mathcal{L}_\text{geo}$ & $\bf{r}_\text{novel}$ & PSNR $\uparrow$ & SSIM $\uparrow$ & LPIPS $\downarrow$ & Model size $\downarrow$ \\
    \midrule
    - & \checkmark   &   \checkmark   &    \checkmark & 17.54 & 0.52 & 0.37 & 221.14 MB \\
    \checkmark & -   &   \checkmark  &    \checkmark   &   16.89  &   0.44  &  0.46 & \bf{183.04 MB} \\
    \checkmark & \checkmark   &   -   &    \checkmark   &   15.97  &   0.49  &  0.41 & \bf{183.04 MB}  \\
    \checkmark & \checkmark   &   \checkmark   &  -  & 17.84   &   0.52   &   0.36 & \bf{183.04 MB}  \\
    \checkmark & \checkmark   &   \checkmark   &    \checkmark   &   \bf{18.07}   &   \bf{0.54}  &  \bf{0.35} & \bf{183.04 MB}\\
    \bottomrule
    \end{tabular}%
    }
\end{table}


\begin{table}[t]
    \centering
    \small
    \setlength{\tabcolsep}{12pt}
    \renewcommand{\arraystretch}{0.8}
    \caption{\textbf{Comparison of the efficiency between $\mathcal{L}_\text{geo}$ and $\mathcal{L}_\text{d}$}.}
    \label{tab:abalation_with_depth}
    \vspace{-3mm}
    \begin{tabular}{cc|cccc}
    \toprule
    $\mathcal{L}_\text{geo}$ & $\mathcal{L}_\text{d}$ & PSNR $\uparrow$ & SSIM $\uparrow$ & LPIPS $\downarrow$ \\
    \midrule
    - & \checkmark  & 17.51 & 0.53 & 0.37 \\
    \checkmark & -   &   18.07  &   0.54  & \textbf{0.35} \\
    \checkmark & \checkmark   &   \textbf{18.26}  &   \textbf{0.55}  & \textbf{0.35} \\
    \bottomrule
    \end{tabular}%
\end{table}

\begin{figure}[t]
\centering
\resizebox{1.0\columnwidth}{!} 
{
\includegraphics[width=\textwidth]{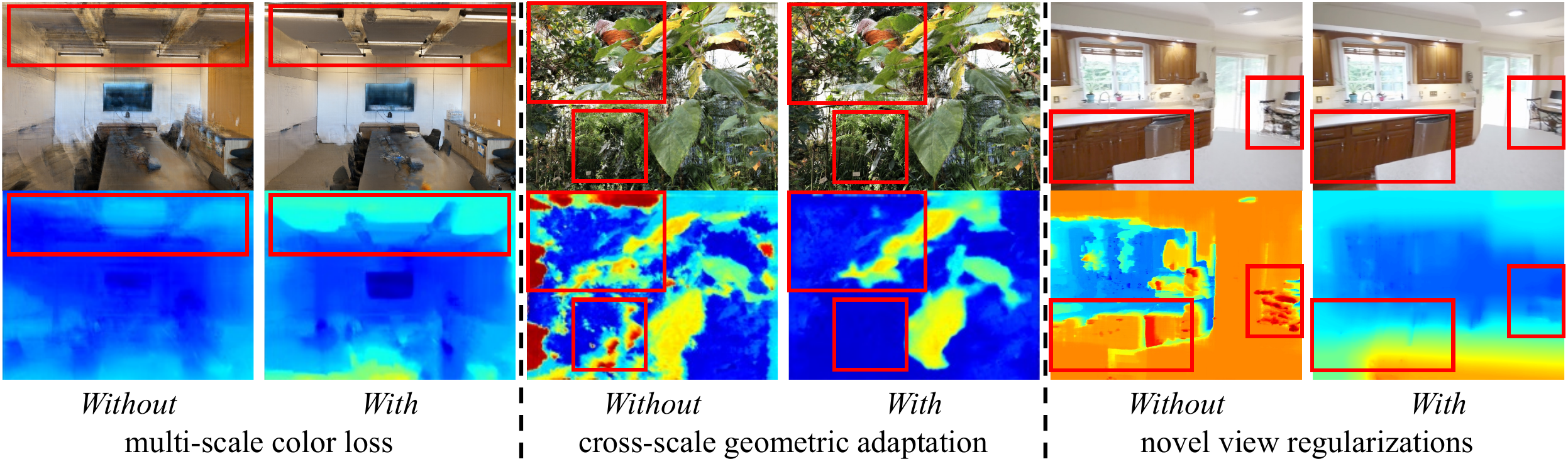}
}
\vspace{-6mm}
\caption{\textbf{Visual comparisons on ablation studies.} 
(\emph{Left}) Multi-scale color loss enhances varied scene details.
(\emph{Mid}) Geometric adaptation greatly suppresses floaters.
(\emph{Right}) Novel view regularizations add supervision for high-fidelity geometry.
}
\label{fig:ablation_vis}
\end{figure}

\vspace{3pt}  \noindent {\bf Multi-scale voxel color loss.}
We demonstrate the effectiveness of multi-scale voxel color loss $\mathcal{L}_\text{ms-color}$ over single-scale color loss, as shown in \cref{tab:abalation_with_losses} and \cref{fig:ablation_vis}(\emph{Left}). Multi-scale loss captures varied scene details, enhancing rendering and geometry. Without geometric adaptation, FrugalNeRF falls short of FreeNeRF. While our voxel grid trains faster than MLPs, its discrete structure limits initial continuity, which geometric adaptation greatly improves across scales.

\begin{figure}[t]
\resizebox{\columnwidth}{!}{%
\begin{tabular}{cc}
\includegraphics[width=\textwidth]{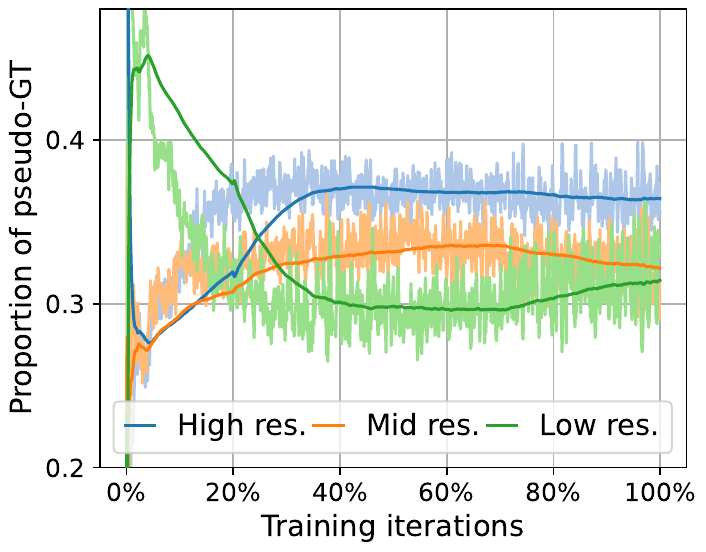} & \includegraphics[width=\textwidth]{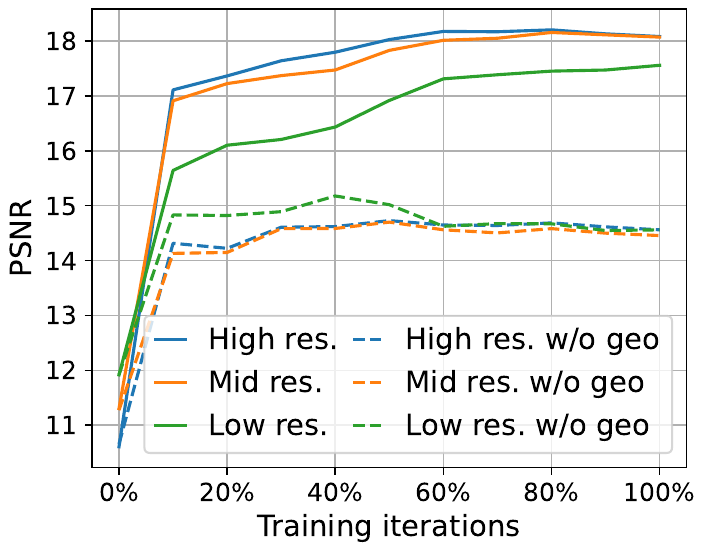} \\
\end{tabular}%
}
\vspace{-3mm}
\caption{\textbf{Cross-scale geometric adaptation in training.}  
(\emph{Left}) Early training uses low-resolution voxels as pseudo-ground truth to guide initial geometry learning, with medium- and high-resolution voxels refining details as training advances. (\emph{Right}) Without geometric adaptation, all scales perform sub-optimally, whereas adaptation drives convergence to higher quality across scales.
}
\label{fig:self_depth_ratio}
\end{figure}

\vspace{3pt}  \noindent {\bf Cross-scale geometric adaptation.}
\cref{tab:abalation_with_losses} shows performance declines on all metrics without the geometric adaptation loss $\mathcal{L}_\text{geo}$. \cref{fig:ablation_vis}(\emph{Mid}) illustrates that geometric adaptation significantly reduces floaters. In \cref{fig:self_depth_ratio}(\emph{Left}), low-frequency components from low-resolution voxels guide coarse geometry early in training, with mid- and high-frequency components gradually increasing to refine details. This self-adaptive process resembles MLP-based frequency regularization but requires no complex scheduling, allowing FrugalNeRF to generalize well to diverse scenes. \cref{fig:self_depth_ratio}(\emph{Right}) confirms that geometric adaptation improves convergence quality across scales.

\begin{figure}[h]
\centering
\resizebox{0.95\columnwidth}{!} 
{
\includegraphics[width=\textwidth]{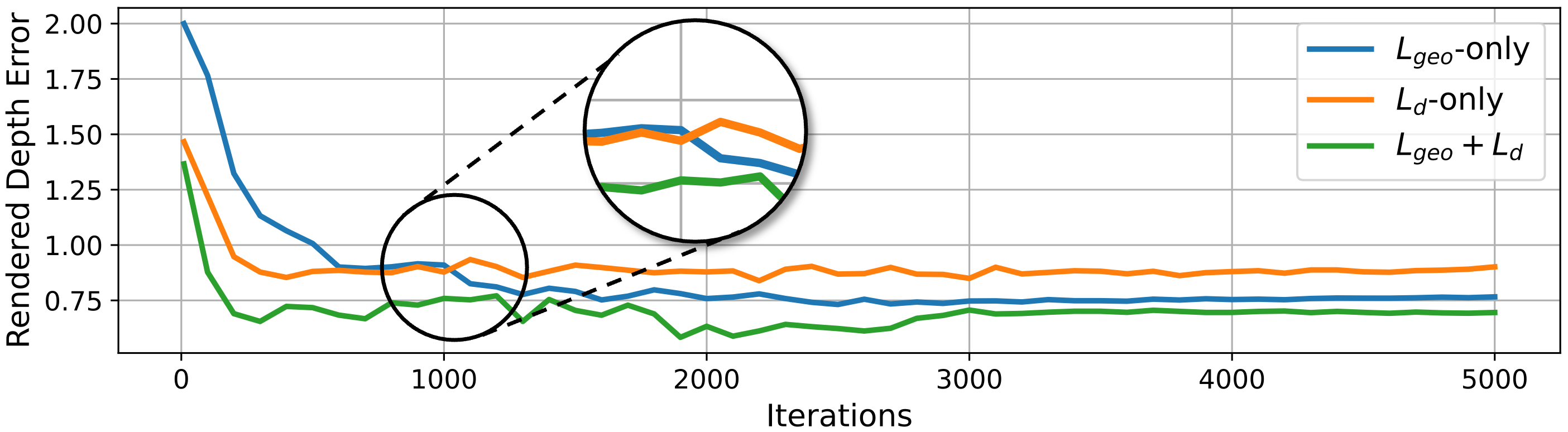}}
\vspace{-3mm}
\caption{\textbf{Rendered depth error between $\mathcal{L}_\text{geo}$ and $\mathcal{L}_\text{d}$.} \(L_{\text{geo}}\)-only achieves lower final error. Combining both yields the best results.}
\label{fig:render_depth_error}
\end{figure}

\vspace{3pt}  \noindent {\bf Efficiency between $\mathcal{L}_\text{geo}$ and $\mathcal{L}_\text{d}$.}
\cref{fig:render_depth_error} shows that $\mathcal{L}_\text{d}$ alone initially offers good performance but later suffers from higher error due to scale mismatches, whereas $\mathcal{L}_\text{geo}$ alone refines depth progressively over time, achieving lower final error. This indicates that cross-scale geometric adaptation provides a stronger and more robust constraint than the depth prior. Combining both $\mathcal{L}_\text{geo}$ and $\mathcal{L}_\text{geo}$ yields even better results, demonstrating their complementarity and highlighting FrugalNeRF's flexibility, as further validated by \cref{tab:abalation_with_depth}.


\begin{figure}[t]
\includegraphics[width=\columnwidth]{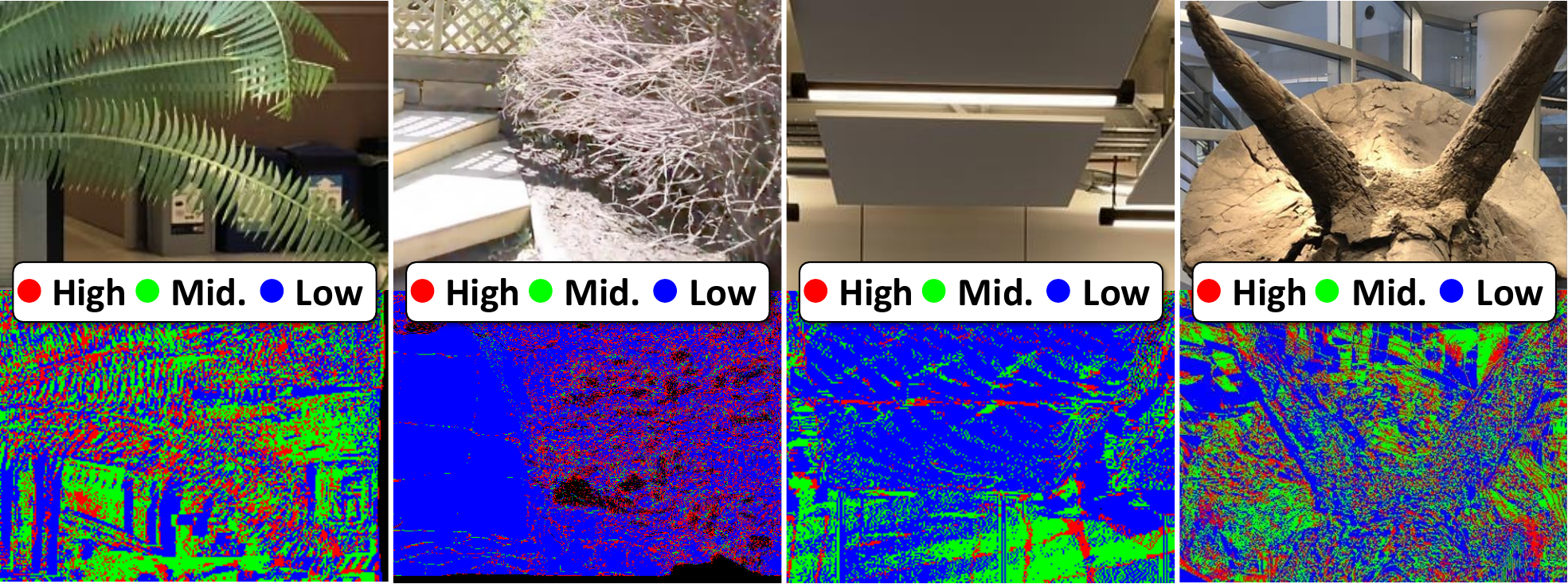}
\vspace{-6mm}
\caption{\textbf{Scene dependency analysis of the multi-scale voxels.} Cross-scale geometric adaptation can adapt to diverse scenes.
}
\label{fig:scene_dependency_analysis}
\end{figure}

\vspace{3pt}  \noindent {\bf Scene dependency analysis of the multi-scale voxels.}
We analyze the scene dependency of the multi-scale voxels in~\cref{fig:scene_dependency_analysis}. The results indicate that scenes with foliage exhibit higher activations in high- and mid-frequency voxels, while textureless scenes show significant activations in low-frequency voxels. This confirms our approach’s adaptability to different scene configurations.



\vspace{3pt}  \noindent {\bf Novel view regularizations.}
We assessed novel view regularizations by omitting novel view rays $\bf{r}_\text{novel}$. As shown in \cref{tab:abalation_with_losses}, including these rays improves rendering quality. \cref{fig:ablation_vis} (\emph{Right}) shows omitting them risks local minima, causing incorrect geometry. Novel view regularizations add guidance, reducing overfitting and enhancing geometry accuracy.

\section{Conclusion}
\label{sec:conclusion}
We propose FrugalNeRF, a framework synthesizing novel views from extremely few inputs. To accelerate and regularize training, we use weight-sharing voxels across scales capturing varying scene frequencies and geometric adaptation via reprojection errors guiding multi-scale geometry. FrugalNeRF achieves state-of-the-art results on multiple datasets with shorter training and no external priors.

\vspace{3pt}  \noindent {\bf Limitations.}
Few-shot NeRF relies on accurate camera poses for training. In scenarios with significant changes in viewpoint or sparse training views, such as the 360° dataset, the model may face challenges in generalization. Although our method introduces novel-view losses to deal with those unseen regions in training views, it is still an issue for few-shot NeRF.

\paragraph{Acknowledgements.}
This research was funded by the National Science and Technology Council, Taiwan, under Grants NSTC 112-2222-E-A49-004-MY2 and 113-2628-E-A49-023-. The authors are grateful to Google, NVIDIA, and MediaTek Inc. for their generous donations. Yu-Lun Liu acknowledges the Yushan Young Fellow Program by the MOE in Taiwan.

{\small
\bibliographystyle{ieeenat_fullname}
\bibliography{11_references}

\begin{thebibliography}{104}
\providecommand{\natexlab}[1]{#1}
\providecommand{\url}[1]{\texttt{#1}}
\expandafter\ifx\csname urlstyle\endcsname\relax
  \providecommand{\doi}[1]{doi: #1}\else
  \providecommand{\doi}{doi: \begingroup \urlstyle{rm}\Url}\fi

\bibitem[Ahn et~al.(2022)Ahn, Jang, Park, Kim, and Kang]{ahn2022panerf}
Young~Chun Ahn, Seokhwan Jang, Sungheon Park, Ji-Yeon Kim, and Nahyup Kang.
\newblock Panerf: Pseudo-view augmentation for improved neural radiance fields based on few-shot inputs.
\newblock \emph{arXiv preprint arXiv:2211.12758}, 2022.

\bibitem[Bai et~al.(2023)Bai, Huang, Gong, Guo, and Guo]{bai2023self}
Jiayang Bai, Letian Huang, Wen Gong, Jie Guo, and Yanwen Guo.
\newblock Self-nerf: A self-training pipeline for few-shot neural radiance fields.
\newblock \emph{arXiv preprint arXiv:2303.05775}, 2023.

\bibitem[Barron et~al.(2022)Barron, Mildenhall, Verbin, Srinivasan, and Hedman]{barron2022mip}
Jonathan~T Barron, Ben Mildenhall, Dor Verbin, Pratul~P Srinivasan, and Peter Hedman.
\newblock Mip-nerf 360: Unbounded anti-aliased neural radiance fields.
\newblock In \emph{CVPR}, 2022.

\bibitem[Bian et~al.(2023)Bian, Wang, Li, Bian, and Prisacariu]{bian2023nope}
Wenjing Bian, Zirui Wang, Kejie Li, Jia-Wang Bian, and Victor~Adrian Prisacariu.
\newblock Nope-nerf: Optimising neural radiance field with no pose prior.
\newblock In \emph{CVPR}, 2023.

\bibitem[Bortolon et~al.(2022)Bortolon, Del~Bue, and Poiesi]{bortolon2022vm}
Matteo Bortolon, Alessio Del~Bue, and Fabio Poiesi.
\newblock Vm-nerf: Tackling sparsity in nerf with view morphing.
\newblock \emph{arXiv preprint arXiv:2210.04214}, 2022.

\bibitem[Caron et~al.(2021)Caron, Touvron, Misra, J{\'e}gou, Mairal, Bojanowski, and Joulin]{caron2021emerging}
Mathilde Caron, Hugo Touvron, Ishan Misra, Herv{\'e} J{\'e}gou, Julien Mairal, Piotr Bojanowski, and Armand Joulin.
\newblock Emerging properties in self-supervised vision transformers.
\newblock In \emph{ICCV}, 2021.

\bibitem[Chan et~al.(2021)Chan, Monteiro, Kellnhofer, Wu, and Wetzstein]{chan2021pi}
Eric~R Chan, Marco Monteiro, Petr Kellnhofer, Jiajun Wu, and Gordon Wetzstein.
\newblock pi-gan: Periodic implicit generative adversarial networks for 3d-aware image synthesis.
\newblock In \emph{CVPR}, 2021.

\bibitem[Chan et~al.(2022)Chan, Lin, Chan, Nagano, Pan, De~Mello, Gallo, Guibas, Tremblay, Khamis, et~al.]{chan2022efficient}
Eric~R Chan, Connor~Z Lin, Matthew~A Chan, Koki Nagano, Boxiao Pan, Shalini De~Mello, Orazio Gallo, Leonidas~J Guibas, Jonathan Tremblay, Sameh Khamis, et~al.
\newblock Efficient geometry-aware 3d generative adversarial networks.
\newblock In \emph{CVPR}, 2022.

\bibitem[Chen et~al.(2021)Chen, Xu, Zhao, Zhang, Xiang, Yu, and Su]{chen2021mvsnerf}
Anpei Chen, Zexiang Xu, Fuqiang Zhao, Xiaoshuai Zhang, Fanbo Xiang, Jingyi Yu, and Hao Su.
\newblock Mvsnerf: Fast generalizable radiance field reconstruction from multi-view stereo.
\newblock In \emph{ICCV}, 2021.

\bibitem[Chen et~al.(2022{\natexlab{a}})Chen, Xu, Geiger, Yu, and Su]{chen2022tensorf}
Anpei Chen, Zexiang Xu, Andreas Geiger, Jingyi Yu, and Hao Su.
\newblock Tensorf: Tensorial radiance fields.
\newblock In \emph{ECCV}, 2022{\natexlab{a}}.

\bibitem[Chen et~al.(2024)Chen, Chiu, and Liu]{chen2024improving}
Bo-Yu Chen, Wei-Chen Chiu, and Yu-Lun Liu.
\newblock Improving robustness for joint optimization of camera pose and decomposed low-rank tensorial radiance fields.
\newblock In \emph{AAAI}, 2024.

\bibitem[Chen et~al.(2022{\natexlab{b}})Chen, Liu, Huang, Wang, and Pan]{chen2022geoaug}
Di Chen, Yu Liu, Lianghua Huang, Bin Wang, and Pan Pan.
\newblock Geoaug: Data augmentation for few-shot nerf with geometry constraints.
\newblock In \emph{ECCV}, 2022{\natexlab{b}}.

\bibitem[Chen et~al.(2016)Chen, Fu, Yang, and Deng]{chen2016single}
Weifeng Chen, Zhao Fu, Dawei Yang, and Jia Deng.
\newblock Single-image depth perception in the wild.
\newblock In \emph{NeurIPS}, 2016.

\bibitem[Chen et~al.(2022{\natexlab{c}})Chen, Zhang, Li, Chen, Feng, Wang, and Wang]{chen2022hallucinated}
Xingyu Chen, Qi Zhang, Xiaoyu Li, Yue Chen, Ying Feng, Xuan Wang, and Jue Wang.
\newblock Hallucinated neural radiance fields in the wild.
\newblock In \emph{CVPR}, 2022{\natexlab{c}}.

\bibitem[Chen et~al.(2023{\natexlab{a}})Chen, Wang, Guo, and Zhang]{chen2023structnerf}
Zheng Chen, Chen Wang, Yuan-Chen Guo, and Song-Hai Zhang.
\newblock Structnerf: Neural radiance fields for indoor scenes with structural hints.
\newblock \emph{IEEE TPAMI}, 2023{\natexlab{a}}.

\bibitem[Chen et~al.(2023{\natexlab{b}})Chen, Yang, Lai, and Xie]{chen2023cunerf}
Zixuan Chen, Lingxiao Yang, Jian-Huang Lai, and Xiaohua Xie.
\newblock Cunerf: Cube-based neural radiance field for zero-shot medical image arbitrary-scale super resolution.
\newblock In \emph{ICCV}, 2023{\natexlab{b}}.

\bibitem[Chibane et~al.(2021)Chibane, Bansal, Lazova, and Pons-Moll]{chibane2021stereo}
Julian Chibane, Aayush Bansal, Verica Lazova, and Gerard Pons-Moll.
\newblock Stereo radiance fields (srf): Learning view synthesis for sparse views of novel scenes.
\newblock In \emph{CVPR}, 2021.

\bibitem[Dai et~al.(2017)Dai, Chang, Savva, Halber, Funkhouser, and Nie{\ss}ner]{dai2017scannet}
Angela Dai, Angel~X Chang, Manolis Savva, Maciej Halber, Thomas Funkhouser, and Matthias Nie{\ss}ner.
\newblock Scannet: Richly-annotated 3d reconstructions of indoor scenes.
\newblock In \emph{CVPR}, 2017.

\bibitem[Darmon et~al.(2022)Darmon, Bascle, Devaux, Monasse, and Aubry]{darmon2022improving}
Fran{\c{c}}ois Darmon, B{\'e}n{\'e}dicte Bascle, Jean-Cl{\'e}ment Devaux, Pascal Monasse, and Mathieu Aubry.
\newblock Improving neural implicit surfaces geometry with patch warping.
\newblock In \emph{CVPR}, 2022.

\bibitem[Deng et~al.(2023)Deng, Jiang, Qi, Yan, Zhou, Guibas, Anguelov, et~al.]{deng2023nerdi}
Congyue Deng, Chiyu Jiang, Charles~R Qi, Xinchen Yan, Yin Zhou, Leonidas Guibas, Dragomir Anguelov, et~al.
\newblock Nerdi: Single-view nerf synthesis with language-guided diffusion as general image priors.
\newblock In \emph{CVPR}, 2023.

\bibitem[Deng et~al.(2022)Deng, Liu, Zhu, and Ramanan]{deng2022depth}
Kangle Deng, Andrew Liu, Jun-Yan Zhu, and Deva Ramanan.
\newblock Depth-supervised nerf: Fewer views and faster training for free.
\newblock In \emph{CVPR}, 2022.

\bibitem[Fridovich-Keil et~al.(2022)Fridovich-Keil, Yu, Tancik, Chen, Recht, and Kanazawa]{fridovich2022plenoxels}
Sara Fridovich-Keil, Alex Yu, Matthew Tancik, Qinhong Chen, Benjamin Recht, and Angjoo Kanazawa.
\newblock Plenoxels: Radiance fields without neural networks.
\newblock In \emph{CVPR}, 2022.

\bibitem[Fu et~al.(2022)Fu, Xu, Ong, and Tao]{fu2022geo}
Qiancheng Fu, Qingshan Xu, Yew~Soon Ong, and Wenbing Tao.
\newblock Geo-neus: Geometry-consistent neural implicit surfaces learning for multi-view reconstruction.
\newblock In \emph{NeurIPS}, 2022.

\bibitem[Gao et~al.(2020)Gao, Shih, Lai, Liang, and Huang]{gao2020portrait}
Chen Gao, Yichang Shih, Wei-Sheng Lai, Chia-Kai Liang, and Jia-Bin Huang.
\newblock Portrait neural radiance fields from a single image.
\newblock \emph{arXiv preprint arXiv:2012.05903}, 2020.

\bibitem[Han et~al.(2022)Han, Wang, and Yang]{han2022single}
Yuxuan Han, Ruicheng Wang, and Jiaolong Yang.
\newblock Single-view view synthesis in the wild with learned adaptive multiplane images.
\newblock In \emph{ACM SIGGRAPH 2022 Conference Proceedings}, 2022.

\bibitem[Hong et~al.(2023)Hong, Chen, Lan, Pan, and Liu]{hong2022eva3d}
Fangzhou Hong, Zhaoxi Chen, Yushi Lan, Liang Pan, and Ziwei Liu.
\newblock Eva3d: Compositional 3d human generation from 2d image collections.
\newblock In \emph{ICLR}, 2023.

\bibitem[Hu et~al.(2021)Hu, Ravi, Berg, and Pathak]{hu2021worldsheet}
Ronghang Hu, Nikhila Ravi, Alexander~C Berg, and Deepak Pathak.
\newblock Worldsheet: Wrapping the world in a 3d sheet for view synthesis from a single image.
\newblock In \emph{ICCV}, 2021.

\bibitem[Hu et~al.(2023{\natexlab{a}})Hu, Hong, Pan, Mei, Yang, and Liu]{hu2023sherf}
Shoukang Hu, Fangzhou Hong, Liang Pan, Haiyi Mei, Lei Yang, and Ziwei Liu.
\newblock Sherf: Generalizable human nerf from a single image.
\newblock In \emph{ICCV}, 2023{\natexlab{a}}.

\bibitem[Hu et~al.(2023{\natexlab{b}})Hu, Zhou, Li, Yu, Hong, Hu, Li, Lee, and Liu]{hu2023consistentnerf}
Shoukang Hu, Kaichen Zhou, Kaiyu Li, Longhui Yu, Lanqing Hong, Tianyang Hu, Zhenguo Li, Gim~Hee Lee, and Ziwei Liu.
\newblock Consistentnerf: Enhancing neural radiance fields with 3d consistency for sparse view synthesis.
\newblock \emph{arXiv preprint arXiv:2305.11031}, 2023{\natexlab{b}}.

\bibitem[Jain et~al.(2021)Jain, Tancik, and Abbeel]{jain2021putting}
Ajay Jain, Matthew Tancik, and Pieter Abbeel.
\newblock Putting nerf on a diet: Semantically consistent few-shot view synthesis.
\newblock In \emph{ICCV}, 2021.

\bibitem[Jain et~al.(2022)Jain, Mildenhall, Barron, Abbeel, and Poole]{jain2022zero}
Ajay Jain, Ben Mildenhall, Jonathan~T Barron, Pieter Abbeel, and Ben Poole.
\newblock Zero-shot text-guided object generation with dream fields.
\newblock In \emph{CVPR}, 2022.

\bibitem[Jensen et~al.(2014)Jensen, Dahl, Vogiatzis, Tola, and Aan{\ae}s]{jensen2014large}
Rasmus Jensen, Anders Dahl, George Vogiatzis, Engin Tola, and Henrik Aan{\ae}s.
\newblock Large scale multi-view stereopsis evaluation.
\newblock In \emph{CVPR}, 2014.

\bibitem[Johari et~al.(2022)Johari, Lepoittevin, and Fleuret]{johari2022geonerf}
Mohammad~Mahdi Johari, Yann Lepoittevin, and Fran{\c{c}}ois Fleuret.
\newblock Geonerf: Generalizing nerf with geometry priors.
\newblock In \emph{CVPR}, 2022.

\bibitem[Jung et~al.(2023)Jung, Han, Kang, Kim, Kwak, and Kim]{jung2023self}
Jaewoo Jung, Jisang Han, Jiwon Kang, Seongchan Kim, Min-Seop Kwak, and Seungryong Kim.
\newblock Self-evolving neural radiance fields.
\newblock \emph{arXiv preprint arXiv:2312.01003}, 2023.

\bibitem[Kim et~al.(2022)Kim, Seo, and Han]{kim2022infonerf}
Mijeong Kim, Seonguk Seo, and Bohyung Han.
\newblock Infonerf: Ray entropy minimization for few-shot neural volume rendering.
\newblock In \emph{CVPR}, 2022.

\bibitem[Kingma and Ba(2014)]{kingma2014adam}
Diederik~P Kingma and Jimmy Ba.
\newblock Adam: A method for stochastic optimization.
\newblock \emph{arXiv preprint arXiv:1412.6980}, 2014.

\bibitem[Kopf et~al.(2021)Kopf, Rong, and Huang]{kopf2021robust}
Johannes Kopf, Xuejian Rong, and Jia-Bin Huang.
\newblock Robust consistent video depth estimation.
\newblock In \emph{CVPR}, 2021.

\bibitem[Krizhevsky et~al.(2012)Krizhevsky, Sutskever, and Hinton]{NIPS2012_c399862d}
Alex Krizhevsky, Ilya Sutskever, and Geoffrey~E Hinton.
\newblock Imagenet classification with deep convolutional neural networks.
\newblock In \emph{NeurIPS}, 2012.

\bibitem[Kwak et~al.(2023)Kwak, Song, and Kim]{kwak2023geconerf}
Minseop Kwak, Jiuhn Song, and Seungryong Kim.
\newblock Geconerf: Few-shot neural radiance fields via geometric consistency.
\newblock In \emph{ICML}, 2023.

\bibitem[Lee et~al.(2023)Lee, Choi, Kim, Kim, and Cho]{lee2023extremenerf}
SeokYeong Lee, JunYong Choi, Seungryong Kim, Ig-Jae Kim, and Junghyun Cho.
\newblock Extremenerf: Few-shot neural radiance fields under unconstrained illumination.
\newblock \emph{arXiv preprint arXiv:2303.11728}, 2023.

\bibitem[Li et~al.(2022)Li, Zhuang, Wang, Liang, Chang, and Yang]{li2022automated}
Changlin Li, Bohan Zhuang, Guangrun Wang, Xiaodan Liang, Xiaojun Chang, and Yi Yang.
\newblock Automated progressive learning for efficient training of vision transformers.
\newblock In \emph{CVPR}, 2022.

\bibitem[Li et~al.(2024)Li, Zhang, Bai, Zheng, Ning, Zhou, and Gu]{li2024dngaussian}
Jiahe Li, Jiawei Zhang, Xiao Bai, Jin Zheng, Xin Ning, Jun Zhou, and Lin Gu.
\newblock Dngaussian: Optimizing sparse-view 3d gaussian radiance fields with global-local depth normalization.
\newblock \emph{arXiv preprint arXiv:2403.06912}, 2024.

\bibitem[Li et~al.(2021)Li, Luo, Zhu, Zhao, Li, and Shan]{li2021enforcing}
Siyuan Li, Yue Luo, Ye Zhu, Xun Zhao, Yu Li, and Ying Shan.
\newblock Enforcing temporal consistency in video depth estimation.
\newblock In \emph{ICCV}, 2021.

\bibitem[Lin et~al.(2023)Lin, Lin, Lai, Lin, Shih, and Ramamoorthi]{lin2023vision}
Kai-En Lin, Yen-Chen Lin, Wei-Sheng Lai, Tsung-Yi Lin, Yi-Chang Shih, and Ravi Ramamoorthi.
\newblock Vision transformer for nerf-based view synthesis from a single input image.
\newblock In \emph{WACV}, 2023.

\bibitem[Liu et~al.(2023)Liu, Gao, Meuleman, Tseng, Saraf, Kim, Chuang, Kopf, and Huang]{liu2023robust}
Yu-Lun Liu, Chen Gao, Andreas Meuleman, Hung-Yu Tseng, Ayush Saraf, Changil Kim, Yung-Yu Chuang, Johannes Kopf, and Jia-Bin Huang.
\newblock Robust dynamic radiance fields.
\newblock In \emph{CVPR}, 2023.

\bibitem[Luo et~al.(2020)Luo, Huang, Szeliski, Matzen, and Kopf]{luo2020consistent}
Xuan Luo, Jia-Bin Huang, Richard Szeliski, Kevin Matzen, and Johannes Kopf.
\newblock Consistent video depth estimation.
\newblock \emph{ACM Transactions on Graphics (ToG)}, 2020.

\bibitem[Martin-Brualla et~al.(2021)Martin-Brualla, Radwan, Sajjadi, Barron, Dosovitskiy, and Duckworth]{martin2021nerf}
Ricardo Martin-Brualla, Noha Radwan, Mehdi~SM Sajjadi, Jonathan~T Barron, Alexey Dosovitskiy, and Daniel Duckworth.
\newblock Nerf in the wild: Neural radiance fields for unconstrained photo collections.
\newblock In \emph{CVPR}, 2021.

\bibitem[Meuleman et~al.(2023)Meuleman, Liu, Gao, Huang, Kim, Kim, and Kopf]{meuleman2023progressively}
Andreas Meuleman, Yu-Lun Liu, Chen Gao, Jia-Bin Huang, Changil Kim, Min~H Kim, and Johannes Kopf.
\newblock Progressively optimized local radiance fields for robust view synthesis.
\newblock In \emph{CVPR}, 2023.

\bibitem[Mildenhall et~al.(2019{\natexlab{a}})Mildenhall, Srinivasan, Ortiz-Cayon, Kalantari, Ramamoorthi, Ng, and Kar]{mildenhall2019llff}
Ben Mildenhall, Pratul~P. Srinivasan, Rodrigo Ortiz-Cayon, Nima~Khademi Kalantari, Ravi Ramamoorthi, Ren Ng, and Abhishek Kar.
\newblock Local light field fusion: Practical view synthesis with prescriptive sampling guidelines.
\newblock \emph{ACM Transactions on Graphics (TOG)}, 2019{\natexlab{a}}.

\bibitem[Mildenhall et~al.(2019{\natexlab{b}})Mildenhall, Srinivasan, Ortiz-Cayon, Kalantari, Ramamoorthi, Ng, and Kar]{mildenhall2019local}
Ben Mildenhall, Pratul~P Srinivasan, Rodrigo Ortiz-Cayon, Nima~Khademi Kalantari, Ravi Ramamoorthi, Ren Ng, and Abhishek Kar.
\newblock Local light field fusion: Practical view synthesis with prescriptive sampling guidelines.
\newblock \emph{ACM Transactions on Graphics (TOG)}, 2019{\natexlab{b}}.

\bibitem[Mildenhall et~al.(2020)Mildenhall, Srinivasan, Tancik, Barron, Ramamoorthi, and Ng]{mildenhall2020nerf}
Ben Mildenhall, Pratul~P. Srinivasan, Matthew Tancik, Jonathan~T. Barron, Ravi Ramamoorthi, and Ren Ng.
\newblock Nerf: Representing scenes as neural radiance fields for view synthesis.
\newblock In \emph{ECCV}, 2020.

\bibitem[Mildenhall et~al.(2022)Mildenhall, Hedman, Martin-Brualla, Srinivasan, and Barron]{mildenhall2022nerf}
Ben Mildenhall, Peter Hedman, Ricardo Martin-Brualla, Pratul~P Srinivasan, and Jonathan~T Barron.
\newblock Nerf in the dark: High dynamic range view synthesis from noisy raw images.
\newblock In \emph{CVPR}, 2022.

\bibitem[M{\"u}ller et~al.(2022)M{\"u}ller, Evans, Schied, and Keller]{muller2022instant}
Thomas M{\"u}ller, Alex Evans, Christoph Schied, and Alexander Keller.
\newblock Instant neural graphics primitives with a multiresolution hash encoding.
\newblock \emph{ACM Transactions on Graphics (ToG)}, 2022.

\bibitem[Niemeyer et~al.(2022)Niemeyer, Barron, Mildenhall, Sajjadi, Geiger, and Radwan]{niemeyer2022regnerf}
Michael Niemeyer, Jonathan~T Barron, Ben Mildenhall, Mehdi~SM Sajjadi, Andreas Geiger, and Noha Radwan.
\newblock Regnerf: Regularizing neural radiance fields for view synthesis from sparse inputs.
\newblock In \emph{CVPR}, 2022.

\bibitem[Oechsle et~al.(2021)Oechsle, Peng, and Geiger]{oechsle2021unisurf}
Michael Oechsle, Songyou Peng, and Andreas Geiger.
\newblock Unisurf: Unifying neural implicit surfaces and radiance fields for multi-view reconstruction.
\newblock In \emph{ICCV}, 2021.

\bibitem[Peng et~al.(2021)Peng, Zhang, Li, Wang, Liang, and Lin]{peng2021pi}
Jiefeng Peng, Jiqi Zhang, Changlin Li, Guangrun Wang, Xiaodan Liang, and Liang Lin.
\newblock Pi-nas: Improving neural architecture search by reducing supernet training consistency shift.
\newblock In \emph{ICCV}, 2021.

\bibitem[Pumarola et~al.(2021)Pumarola, Corona, Pons-Moll, and Moreno-Noguer]{pumarola2021d}
Albert Pumarola, Enric Corona, Gerard Pons-Moll, and Francesc Moreno-Noguer.
\newblock D-nerf: Neural radiance fields for dynamic scenes.
\newblock In \emph{CVPR}, 2021.

\bibitem[Radford et~al.(2021)Radford, Kim, Hallacy, Ramesh, Goh, Agarwal, Sastry, Askell, Mishkin, Clark, et~al.]{radford2021learning}
Alec Radford, Jong~Wook Kim, Chris Hallacy, Aditya Ramesh, Gabriel Goh, Sandhini Agarwal, Girish Sastry, Amanda Askell, Pamela Mishkin, Jack Clark, et~al.
\newblock Learning transferable visual models from natural language supervision.
\newblock In \emph{ICML}, 2021.

\bibitem[Ranftl et~al.(2020)Ranftl, Lasinger, Hafner, Schindler, and Koltun]{Ranftl2020}
Ren\'{e} Ranftl, Katrin Lasinger, David Hafner, Konrad Schindler, and Vladlen Koltun.
\newblock Towards robust monocular depth estimation: Mixing datasets for zero-shot cross-dataset transfer.
\newblock \emph{IEEE TPAMI}, 2020.

\bibitem[Ranftl et~al.(2021{\natexlab{a}})Ranftl, Bochkovskiy, and Koltun]{Ranftl2021}
Ren\'{e} Ranftl, Alexey Bochkovskiy, and Vladlen Koltun.
\newblock Vision transformers for dense prediction.
\newblock \emph{arXiv preprint arXiv:2103.13413}, 2021{\natexlab{a}}.

\bibitem[Ranftl et~al.(2021{\natexlab{b}})Ranftl, Bochkovskiy, and Koltun]{ranftl2021vision}
Ren{\'e} Ranftl, Alexey Bochkovskiy, and Vladlen Koltun.
\newblock Vision transformers for dense prediction.
\newblock In \emph{ICCV}, 2021{\natexlab{b}}.

\bibitem[Roessle et~al.(2022)Roessle, Barron, Mildenhall, Srinivasan, and Nie{\ss}ner]{roessle2022dense}
Barbara Roessle, Jonathan~T Barron, Ben Mildenhall, Pratul~P Srinivasan, and Matthias Nie{\ss}ner.
\newblock Dense depth priors for neural radiance fields from sparse input views.
\newblock In \emph{CVPR}, 2022.

\bibitem[Sch\"{o}nberger and Frahm(2016)]{schoenberger2016sfm}
Johannes~Lutz Sch\"{o}nberger and Jan-Michael Frahm.
\newblock Structure-from-motion revisited.
\newblock In \emph{CVPR}, 2016.

\bibitem[Sch\"{o}nberger et~al.(2016)Sch\"{o}nberger, Zheng, Pollefeys, and Frahm]{schoenberger2016mvs}
Johannes~Lutz Sch\"{o}nberger, Enliang Zheng, Marc Pollefeys, and Jan-Michael Frahm.
\newblock Pixelwise view selection for unstructured multi-view stereo.
\newblock In \emph{ECCV}, 2016.

\bibitem[Seo et~al.(2023{\natexlab{a}})Seo, Chang, and Kwak]{seo2023flipnerf}
Seunghyeon Seo, Yeonjin Chang, and Nojun Kwak.
\newblock Flipnerf: Flipped reflection rays for few-shot novel view synthesis.
\newblock In \emph{ICCV}, 2023{\natexlab{a}}.

\bibitem[Seo et~al.(2023{\natexlab{b}})Seo, Han, Chang, and Kwak]{seo2023mixnerf}
Seunghyeon Seo, Donghoon Han, Yeonjin Chang, and Nojun Kwak.
\newblock Mixnerf: Modeling a ray with mixture density for novel view synthesis from sparse inputs.
\newblock In \emph{CVPR}, 2023{\natexlab{b}}.

\bibitem[Shi et~al.(2024)Shi, Wei, Wang, and Su]{shi2023zerorf}
Ruoxi Shi, Xinyue Wei, Cheng Wang, and Hao Su.
\newblock Zerorf: Fast sparse view 360 $\{$$\backslash$deg$\}$ reconstruction with zero pretraining.
\newblock In \emph{CVPR}, 2024.

\bibitem[Sitzmann et~al.(2019{\natexlab{a}})Sitzmann, Thies, Heide, Nie{\ss}ner, Wetzstein, and Zollhofer]{sitzmann2019deepvoxels}
Vincent Sitzmann, Justus Thies, Felix Heide, Matthias Nie{\ss}ner, Gordon Wetzstein, and Michael Zollhofer.
\newblock Deepvoxels: Learning persistent 3d feature embeddings.
\newblock In \emph{CVPR}, 2019{\natexlab{a}}.

\bibitem[Sitzmann et~al.(2019{\natexlab{b}})Sitzmann, Zollh{\"o}fer, and Wetzstein]{sitzmann2019scene}
Vincent Sitzmann, Michael Zollh{\"o}fer, and Gordon Wetzstein.
\newblock Scene representation networks: Continuous 3d-structure-aware neural scene representations.
\newblock In \emph{NeurIPS}, 2019{\natexlab{b}}.

\bibitem[Sitzmann et~al.(2020)Sitzmann, Martel, Bergman, Lindell, and Wetzstein]{sitzmann2020implicit}
Vincent Sitzmann, Julien Martel, Alexander Bergman, David Lindell, and Gordon Wetzstein.
\newblock Implicit neural representations with periodic activation functions.
\newblock In \emph{NeurIPS}, 2020.

\bibitem[Somraj and Soundararajan(2023)]{somraj2023vip}
Nagabhushan Somraj and Rajiv Soundararajan.
\newblock Vip-nerf: Visibility prior for sparse input neural radiance fields.
\newblock In \emph{ACM SIGGRAPH}, 2023.

\bibitem[Somraj et~al.(2023)Somraj, Karanayil, and Soundararajan]{somraj2023simplenerf}
Nagabhushan Somraj, Adithyan Karanayil, and Rajiv Soundararajan.
\newblock Simplenerf: Regularizing sparse input neural radiance fields with simpler solutions.
\newblock In \emph{ACM SIGGRAPH Asia}, 2023.

\bibitem[Song et~al.(2023)Song, Park, An, Cho, Kwak, Cho, and Kim]{song2023d}
Jiuhn Song, Seonghoon Park, Honggyu An, Seokju Cho, Min-Seop Kwak, Sungjin Cho, and Seungryong Kim.
\newblock D$\backslash$" arf: Boosting radiance fields from sparse inputs with monocular depth adaptation.
\newblock In \emph{NeurIPS}, 2023.

\bibitem[Su et~al.(2024)Su, Hu, Tsai, Lee, Lin, and Liu]{su2024boostmvsnerfs}
Chih-Hai Su, Chih-Yao Hu, Shr-Ruei Tsai, Jie-Ying Lee, Chin-Yang Lin, and Yu-Lun Liu.
\newblock Boostmvsnerfs: Boosting mvs-based nerfs to generalizable view synthesis in large-scale scenes.
\newblock In \emph{ACM SIGGRAPH 2024 Conference Papers}, 2024.

\bibitem[Sun et~al.(2022)Sun, Sun, and Chen]{sun2022direct}
Cheng Sun, Min Sun, and Hwann-Tzong Chen.
\newblock Direct voxel grid optimization: Super-fast convergence for radiance fields reconstruction.
\newblock In \emph{CVPR}, 2022.

\bibitem[Sun et~al.(2023)Sun, Zhang, Chen, Li, Ji, Zhao, and Xing]{sun2023vgos}
Jiakai Sun, Zhanjie Zhang, Jiafu Chen, Guangyuan Li, Boyan Ji, Lei Zhao, and Wei Xing.
\newblock Vgos: Voxel grid optimization for view synthesis from sparse inputs.
\newblock In \emph{IJCAI}, 2023.

\bibitem[Tancik et~al.(2020)Tancik, Srinivasan, Mildenhall, Fridovich-Keil, Raghavan, Singhal, Ramamoorthi, Barron, and Ng]{tancik2020fourier}
Matthew Tancik, Pratul Srinivasan, Ben Mildenhall, Sara Fridovich-Keil, Nithin Raghavan, Utkarsh Singhal, Ravi Ramamoorthi, Jonathan Barron, and Ren Ng.
\newblock Fourier features let networks learn high frequency functions in low dimensional domains.
\newblock In \emph{NeurIPS}, 2020.

\bibitem[Tao et~al.(2023)Tao, Gao, Wang, Chen, Hao, Liang, Salzmann, and Yu]{tao2023lidar}
Tang Tao, Longfei Gao, Guangrun Wang, Peng Chen, Dayang Hao, Xiaodan Liang, Mathieu Salzmann, and Kaicheng Yu.
\newblock Lidar-nerf: Novel lidar view synthesis via neural radiance fields.
\newblock \emph{arXiv preprint arXiv:2304.10406}, 2023.

\bibitem[Tucker and Snavely(2020)]{tucker2020single}
Richard Tucker and Noah Snavely.
\newblock Single-view view synthesis with multiplane images.
\newblock In \emph{CVPR}, 2020.

\bibitem[Uy et~al.(2023)Uy, Martin-Brualla, Guibas, and Li]{uy2023scade}
Mikaela~Angelina Uy, Ricardo Martin-Brualla, Leonidas Guibas, and Ke Li.
\newblock Scade: Nerfs from space carving with ambiguity-aware depth estimates.
\newblock In \emph{CVPR}, 2023.

\bibitem[Wang and Torr(2022)]{wang2022traditional}
Guangrun Wang and Philip~HS Torr.
\newblock Traditional classification neural networks are good generators: They are competitive with ddpms and gans.
\newblock \emph{arXiv preprint arXiv:2211.14794}, 2022.

\bibitem[Wang et~al.(2023)Wang, Chen, Loy, and Liu]{wang2023sparsenerf}
Guangcong Wang, Zhaoxi Chen, Chen~Change Loy, and Ziwei Liu.
\newblock Sparsenerf: Distilling depth ranking for few-shot novel view synthesis.
\newblock In \emph{ICCV}, 2023.

\bibitem[Wang et~al.(2021{\natexlab{a}})Wang, Wang, Genova, Srinivasan, Zhou, Barron, Martin-Brualla, Snavely, and Funkhouser]{wang2021ibrnet}
Qianqian Wang, Zhicheng Wang, Kyle Genova, Pratul~P Srinivasan, Howard Zhou, Jonathan~T Barron, Ricardo Martin-Brualla, Noah Snavely, and Thomas Funkhouser.
\newblock Ibrnet: Learning multi-view image-based rendering.
\newblock In \emph{CVPR}, 2021{\natexlab{a}}.

\bibitem[Wang et~al.(2022)Wang, Skorokhodov, and Wonka]{wang2022hf}
Yiqun Wang, Ivan Skorokhodov, and Peter Wonka.
\newblock Hf-neus: Improved surface reconstruction using high-frequency details.
\newblock In \emph{NeurIPS}, 2022.

\bibitem[Wang et~al.(2004)Wang, Bovik, Sheikh, and Simoncelli]{wang2004image}
Zhou Wang, Alan~C Bovik, Hamid~R Sheikh, and Eero~P Simoncelli.
\newblock Image quality assessment: from error visibility to structural similarity.
\newblock \emph{IEEE TIP}, 2004.

\bibitem[Wang et~al.(2021{\natexlab{b}})Wang, Wu, Xie, Chen, and Prisacariu]{wang2021nerf}
Zirui Wang, Shangzhe Wu, Weidi Xie, Min Chen, and Victor~Adrian Prisacariu.
\newblock Nerf--: Neural radiance fields without known camera parameters.
\newblock \emph{arXiv preprint arXiv:2102.07064}, 2021{\natexlab{b}}.

\bibitem[Wiles et~al.(2020)Wiles, Gkioxari, Szeliski, and Johnson]{wiles2020synsin}
Olivia Wiles, Georgia Gkioxari, Richard Szeliski, and Justin Johnson.
\newblock Synsin: End-to-end view synthesis from a single image.
\newblock In \emph{CVPR}, 2020.

\bibitem[Wimbauer et~al.(2023)Wimbauer, Yang, Rupprecht, and Cremers]{wimbauer2023behind}
Felix Wimbauer, Nan Yang, Christian Rupprecht, and Daniel Cremers.
\newblock Behind the scenes: Density fields for single view reconstruction.
\newblock In \emph{CVPR}, 2023.

\bibitem[Xu et~al.(2022)Xu, Jiang, Wang, Fan, Shi, and Wang]{xu2022sinnerf}
Dejia Xu, Yifan Jiang, Peihao Wang, Zhiwen Fan, Humphrey Shi, and Zhangyang Wang.
\newblock Sinnerf: Training neural radiance fields on complex scenes from a single image.
\newblock In \emph{ECCV}, 2022.

\bibitem[Xu et~al.(2023)Xu, Jiang, Wang, Fan, Wang, and Wang]{xu2023neurallift}
Dejia Xu, Yifan Jiang, Peihao Wang, Zhiwen Fan, Yi Wang, and Zhangyang Wang.
\newblock Neurallift-360: Lifting an in-the-wild 2d photo to a 3d object with 360deg views.
\newblock In \emph{CVPR}, 2023.

\bibitem[Xu et~al.(2024)Xu, Liu, Tang, Deng, and He]{xu2024learning}
Yingjie Xu, Bangzhen Liu, Hao Tang, Bailin Deng, and Shengfeng He.
\newblock Learning with unreliability: Fast few-shot voxel radiance fields with relative geometric consistency.
\newblock In \emph{CVPR}, 2024.

\bibitem[Yang et~al.(2023)Yang, Pavone, and Wang]{yang2023freenerf}
Jiawei Yang, Marco Pavone, and Yue Wang.
\newblock Freenerf: Improving few-shot neural rendering with free frequency regularization.
\newblock In \emph{CVPR}, 2023.

\bibitem[Yariv et~al.(2020)Yariv, Kasten, Moran, Galun, Atzmon, Ronen, and Lipman]{yariv2020multiview}
Lior Yariv, Yoni Kasten, Dror Moran, Meirav Galun, Matan Atzmon, Basri Ronen, and Yaron Lipman.
\newblock Multiview neural surface reconstruction by disentangling geometry and appearance.
\newblock In \emph{NeurIPS}, 2020.

\bibitem[Ye et~al.(2022)Ye, Li, Tucker, Kanazawa, and Snavely]{ye2022deformable}
Vickie Ye, Zhengqi Li, Richard Tucker, Angjoo Kanazawa, and Noah Snavely.
\newblock Deformable sprites for unsupervised video decomposition.
\newblock In \emph{CVPR}, 2022.

\bibitem[Younes et~al.(2024)Younes, Ouasfi, and Boukhayma]{younes2024sparsecraft}
Mae Younes, Amine Ouasfi, and Adnane Boukhayma.
\newblock Sparsecraft: Few-shot neural reconstruction through stereopsis guided geometric linearization.
\newblock 2024.

\bibitem[Yu et~al.(2021)Yu, Ye, Tancik, and Kanazawa]{yu2021pixelnerf}
Alex Yu, Vickie Ye, Matthew Tancik, and Angjoo Kanazawa.
\newblock pixelnerf: Neural radiance fields from one or few images.
\newblock In \emph{CVPR}, 2021.

\bibitem[Yuan et~al.(2022)Yuan, Sun, Lai, Ma, Jia, and Gao]{yuan2022nerf}
Yu-Jie Yuan, Yang-Tian Sun, Yu-Kun Lai, Yuewen Ma, Rongfei Jia, and Lin Gao.
\newblock Nerf-editing: geometry editing of neural radiance fields.
\newblock In \emph{CVPR}, 2022.

\bibitem[Zhang et~al.(2021)Zhang, Yang, Tulsiani, and Ramanan]{zhang2021ners}
Jason Zhang, Gengshan Yang, Shubham Tulsiani, and Deva Ramanan.
\newblock Ners: Neural reflectance surfaces for sparse-view 3d reconstruction in the wild.
\newblock In \emph{NeurIPS}, 2021.

\bibitem[Zhang et~al.(2020)Zhang, Riegler, Snavely, and Koltun]{zhang2020nerf++}
Kai Zhang, Gernot Riegler, Noah Snavely, and Vladlen Koltun.
\newblock Nerf++: Analyzing and improving neural radiance fields.
\newblock \emph{arXiv preprint arXiv:2010.07492}, 2020.

\bibitem[Zhang et~al.(2018)Zhang, Isola, Efros, Shechtman, and Wang]{zhang2018perceptual}
Richard Zhang, Phillip Isola, Alexei~A Efros, Eli Shechtman, and Oliver Wang.
\newblock The unreasonable effectiveness of deep features as a perceptual metric.
\newblock In \emph{CVPR}, 2018.

\bibitem[Zheng et~al.(2023)Zheng, Lin, and Xu]{zheng2023editablenerf}
Chengwei Zheng, Wenbin Lin, and Feng Xu.
\newblock Editablenerf: Editing topologically varying neural radiance fields by key points.
\newblock In \emph{CVPR}, 2023.

\bibitem[Zhou et~al.(2018)Zhou, Tucker, Flynn, Fyffe, and Snavely]{zhou2018stereo}
Tinghui Zhou, Richard Tucker, John Flynn, Graham Fyffe, and Noah Snavely.
\newblock Stereo magnification: Learning view synthesis using multiplane images.
\newblock In \emph{ACM TOG}, 2018.

\bibitem[Zhou and Tulsiani(2023)]{zhou2023sparsefusion}
Zhizhuo Zhou and Shubham Tulsiani.
\newblock Sparsefusion: Distilling view-conditioned diffusion for 3d reconstruction.
\newblock In \emph{CVPR}, 2023.

\bibitem[Zhu et~al.(2024)Zhu, Fan, Jiang, and Wang]{zhu2023FSGS}
Zehao Zhu, Zhiwen Fan, Yifan Jiang, and Zhangyang Wang.
\newblock Fsgs: Real-time few-shot view synthesis using gaussian splatting.
\newblock In \emph{ECCV}, 2024.

\end{thebibliography}
}

\ifarxiv \clearpage \appendix 


This supplementary material presents additional results to complement the main manuscript.
First, we discuss the difference between competing methods in ~\cref{sec:discussion}
Second, we explain the implementation details in calculating reprojection errors in~\cref{sec:reproj}.
%
%
Then, we provide all the training losses in our training process in~\cref{sec:implement_details}.
Next, we provide additional experiments in ~\cref{sec:additional_exp}.
Moreover, we describe the experimental setup, including the dataset and training time measurement of compared methods in our evaluations in~\cref{sec:exp_setup}.
%
%
In addition to this document, we provide an interactive HTML interface to compare our video results with state-of-the-art methods and show ablation videos and failure cases.
We also attach the source code of our implementation for reference and will make it publicly available for reproducibility.









\section{Discussions on Competing Models}
\label{sec:discussion}
\paragraph{GeCoNeRF.} GeCoNeRF~\citep{kwak2023geconerf} is a few-shot NeRF that uses warped features as pseudo labels, which is sufficiently different from our method. Our method primarily focuses on cross-scale geometric adaptation, selecting render depths with minimal reprojection error across different scales as pseudo labels to adaptively learn the most suitable geometry for each scale. In contrast, GeCoNeRF, besides requiring a pre-trained feature extractor, directly optimizes warped features, making it highly sensitive to geometric noise and resulting in many floaters in its rendering result as shown in our supplementary videos. Our approach, on the other hand, is more robust due to our proposed multi-scale voxels. Low-resolution voxels represent coarse geometry, which is less likely to produce floaters. Using this as supervision effectively suppresses the generation of floaters. 

\paragraph{ZeroRF.} 
ZeroRF~\citep{shi2023zerorf} is a concurrent work to ours, also aimed at training NeRF with sparse input views and achieving fast training times. Unlike TensoRF~\citep{chen2022tensorf}, which directly optimizes the decomposed feature grid, ZeroRF parameterizes the feature grids with a randomly initialized deep neural network (generator). This decision is based on the belief in the higher resilience to noise and artifacts ability of deep neural networks. Although ZeroRF claims to achieve fast convergence stemming from its voxel representation, the need to train the generator results in slower training speeds compared to ours (refer to the main paper Table 2). Our method directly optimizes the feature grid and utilizes cross-scale geometry adaptation to avoid overfitting under sparse views, without requiring a generator that slows down convergence to form decomposed tensorial feature volumes. Additionally, we found that ZeroRF is not suitable for scenes with a background (\emph{e.g.}, LLFF~\citep{mildenhall2019llff}) or datasets like the DTU~\citep{jensen2014large} Dataset, where ZeroRF must extensively use object masks for training. These object masks are not provided directly in these two datasets. Otherwise, ZeroRF may produce many artifacts and floaters, or the feature volume may be filled up to fit the background, leading to severe memory consumption issues causing training failures due to out-of-memory errors.

\paragraph{SparseNeRF.} 
SparseNeRF~\citep{wang2023sparsenerf} proposes a spatial continuity regularization that distills depth continuity priors, but it requires a pre-trained depth prior and is extremely slow by using MLP representation. Additionally, because monocular depth prediction results lack detail, SparseNeRF's rendered results tend to be blurry and lack detail. In contrast, our proposed cross-scale geometric adaptation does not rely on pre-trained priors and ensures the generation of overall geometry while paying attention to details.


\paragraph{SimpleNeRF.} SimpleNeRF~\citep{somraj2023simplenerf} introduces a data augmentation method for few-shot NeRF, employing an MLP with fewer positional encoding frequencies for augmentation, but this simultaneously increases the training time. In contrast, we propose an efficient cross-scale geometric adaptation that achieves multi-scale representation through shared-weight voxels, eliminating the need for an additional model to reconstruct the same scene. This approach yields better results with lower costs. 

\paragraph{FreeNeRF.} 
FreeNeRF~\citep{yang2023freenerf} is an MLP-based few-shot NeRF model. FreeNeRF proposes using a scheduling mechanism to gradually increase input frequency, allowing the model to learn low-frequency geometry during the early stages of training and then ramp up positional encoding to enable the model to learn more detailed geometry later on. However, our approach takes advantage of the explicit voxel representation, which converges faster and allows for direct cross-scaled geometry operations. Additionally, because we employ cross-scale geometry adaptation, our model dynamically determines which frequency of geometry to learn at different training stages. We do not require the complex frequency scheduling of FreeNeRF, nor are we limited to learning only high-frequency components in the later stages of training like FreeNeRF. This makes our method simpler, more general, and more robust.

\paragraph{VGOS.} 
VGOS~\citep{sun2023vgos} introduces an incremental voxel training strategy and a voxel smoothing method for Few-shot NeRF, aimed at reducing training time. It employs a complex scheduling strategy to freeze the outer part of the voxel, leading to a leaky reconstruction of the background scene. Additionally, VGOS requires ground truth poses for novel pose sampling, which results in a quality drop when using random sampling. However, while VGOS's training time is shorter than ours, its performance significantly lags behind. Our cross-scale geometric adaptation strategy eliminates the need for complex scheduling and ground truth pose sampling.

\paragraph{FSGS.} 
FSGS~\citep{zhu2023FSGS} addresses the challenge of limited 3D Gaussian splatting (3DGS) by introducing Proximity-guided Gaussian Unpooling, which adaptively densifies the Gaussians between existing points. Although this method mitigates the issue of insufficient GS, it still relies on a sufficient initial set of Gaussians to perform effectively. In few-shot scenarios, the initial number of GS can be extremely sparse, leading to suboptimal results. Furthermore, FSGS frequently requires novel view inference using monocular depth models during training, which significantly increases the training time. In contrast, our cross-scale geometric adaptation approach ensures rapid convergence without relying on novel view inference or monocular depth models, providing efficient and robust performance even with minimal initial data.

\paragraph{ReVoRF.}
ReVoRF~\citep{xu2024learning} introduces a voxel-based framework that strategically utilizes unreliable regions in pseudo-novel view synthesis for few-shot NeRF. By leveraging a bilateral geometric consistency loss and reliability-aware voxel smoothing, ReVoRF achieves significant improvements in reconstruction quality and training efficiency. However, its reliability mask requires rendering the entire frame to infer the monocular depth model, which limits the frequency of updates. Additionally, its smoothing process does not account for the balance between high- and low-frequency details, resulting in reconstructions that lack fine details. In contrast, FrugalNeRF employs cross-scale geometric adaptation to preserve high-frequency details while preventing floaters. Its high efficiency allows geometric adaptation to be computed at every iteration, ensuring robust and detailed scene reconstruction.

\section{Details of Calculating Reprojection Errors}
\label{sec:reproj}
Mathematically, let ${\bf{p}}_{i}$ be a 2D pixel coordinate in frame $i$, and $\bf{\widetilde{p}}_{i}$ be its homogeneous augmentation. The depth $D_{i}^{l}({\bf{p}}_{i})$ at scale $l$ obtained from volume rendering, and camera intrinsics $K_{i}$ are used to reproject ${\bf{p}}_{i}$ onto the 3D point ${\bf{x}}_{i}^{l}$ in camera coordinate system of frame $i$. Subsequently, utilizing the rotation matrix $R_{i}$ and translation matrix $t_{i}$ of frame $i$, ${\bf{x}}_{i}^{l}$ are transformed into world coordinates system ${\bf{x}}^{l}$:
\begin{equation}
    {\bf{x}}_{i}^{l} = D_{i}^{l}({\bf{p}}_{i}) K_{i}^{-1} {\bf{\widetilde{p}}}_{i}
\end{equation}
\begin{equation}
    {\bf{x}}^{l} =  R_{i} {\bf{x}}_{i}^{l} + t_{i}
\end{equation}
We simplify the previous two equations because the position of the 3D point ${\bf{x}}^{l}$ in world coordinates can also be determined directly from the ray defined by the starting point $o_{i}({\bf{p}}_{i})$ and the direction $v_{i}({\bf{p}}_{i})$:
\begin{equation}
    {\bf{x}}^{l} = o_{i}({\bf{p}}_{i}) + D_{i}^{l}({\bf{p}}_{i}) v_{i}({\bf{p}}_{i})
\end{equation}
Following this, the 3D point ${\bf{x}}^{l}$ in the world coordinate system is transformed to the camera coordinate system of frame $j$ using its rotation matrices $R_{j}$, and translation matrices $T_{j}$:
\begin{equation}
    {\bf{x}}_{i\rightarrow j}^{l} = R_{j}^{T}\left ( {\bf{x}}^{l} - t_{j} \right )
\end{equation}
Finally, project it back to the 2D pixel coordinate system of frame $j$,
\begin{equation}
    \widetilde{{\bf{p}}}_{i\rightarrow j}^{l} = \pi (K_{j} {\bf{x}}_{i\rightarrow j}^{l})
\end{equation}
where $\pi([x, y, z]^T) = \begin{bmatrix} \frac{x}{z}, \frac{y}{z} \end{bmatrix}$. Using coordinates ${\bf{p}}_{i}$ and ${{\bf{p}}}_{i\rightarrow j}^{l}$ to index the RGB maps of frames $i$ (denoted as $C_{i}$) and $j$ (denoted as $C_{j}$), facilitating the computation of the reprojection error:
\begin{equation}
    e^l({\bf{p}}_{i}) = \left \| C_{i}({\bf{p}}_{i}) - C_{j}({{\bf{p}}}_{i\rightarrow j}^{l})\right \|^2 
\end{equation}
Therefore, for each ray sampled from the training view, the pseudo-GT depth of the scale with the minimum reprojection error is obtained,
\begin{equation}
    D'(\mathbf{r}_\text{train}) = \arg \min _{l}(e^{l}(\mathbf{r}_\text{train})).
\end{equation}
where the pseudo-GT depth is utilized to compute the geometric adaptation loss (MSE) $\mathcal{L}_\text{geo}$.
\begin{equation}
    \mathcal{L}_\text{geo}(\mathbf{r}_\text{train}) = \sum_{l=0}^{L} \sum_{r_\text{train} \in \mathcal{R}_\text{train}} {\left \| \hat{D}^l(\mathbf{r}_\text{train}) - D'(\mathbf{r}_\text{train}) \right \|^2}.
\end{equation}
This mechanism provides a supervisory signal for geometry, ensuring that the model can effectively maintain the geometric integrity of the scene across different scales, even in the absence of explicit depth ground truth. It is a pivotal part of the training process, allowing the model to adapt and refine its understanding of the scene's geometric structure in a self-adaptive manner. In our implementation, instead of using a single pixel to calculate reprojection error, we use a patch with $5 \times 5$ pixels to calculate reprojection error. This avoids warping noise caused by similar patterns in scenes, for example, in the case of the LLFF fortress and room. Furthermore, we set a threshold for reprojection error that allows us to ignore cases of image warping with occlusions and prevents crashes during initial training processes, which typically have high reprojection errors.

\section{Losses}
\label{sec:implement_details}
\paragraph{Voxel TV loss ($\mathcal{L}_\text{tv}$).} We use the TV loss on voxel to smooth the result in voxel space.

\paragraph{Patch-wise depth smoothness loss ($\mathcal{L}_\text{ds}$).} We sample patches of rays and calculate the total variance of depth to smooth the geometry in the depth space.

\paragraph{L1 sparsity loss ($\mathcal{L}_\text{l1}$).} We suppress the voxel density in air space by introducing a density L1 regularization loss.

\paragraph{Distortion loss ($\mathcal{L}_\text{dist}$).} We adopt the approach from Mip-NeRF 360~\cite{barron2022mip}, integrating distortion loss to remove floaters from the novel views.

\paragraph{Occlusion loss ($\mathcal{L}_\text{occ}$).} In the DTU dataset, we follow FreeNeRF~\cite {yang2023freenerf} by incorporating an occlusion loss that utilizes black and white background priors to push floaters into the background.

\paragraph{Novel pose sampling form spiraling trajectory.}
We follow the implementation of a spiraling trajectory from TensoRF~\cite{chen2022tensorf}. For the LLFF dataset, we sample 60 novel poses from the spiraling trajectory sampled from training views with 1 rotations, radius scale 1.0, and $z_\text{rate}$ 0.5. 
For the DTU dataset, we sample 60 novel poses from the spiraling trajectory sampled from training views with 4 rotations, radius scale 0.5, and $z_\text{rate}$ 0.5.
For the RealEstate-10K dataset, we sample 60 novel poses from the spiraling trajectory sampled from training views with 2 rotations, radius scale 2.0, and $z_\text{rate}$ 0.5.

\subsection{Details of adding Pretrained Monocular Depth Prior}
\label{sec:depth_prior}
We utilize the pre-trained Dense Prediction Transformer (DPT)~\citep{ranftl2021vision} to generate monocular depth maps from training views. DPT is trained on 1.4 million image-depth pairs, making it a convenient and effective choice for our setup. To address the scale ambiguity between the true scene scale and the estimated depth, we introduce a relaxed relative loss based on Pearson correlation between the estimated and rendered depth maps. This loss is applied at multiple scales, enhancing the monocular depth prior's constraint across different scales and improving the overall geometric consistency. ~\cref{fig:mono_vis} show the render depth on adding Pretrained Monocular Depth Prior.


\begin{figure}[t]
\centering
\resizebox{1.0\columnwidth}{!} 
{
\includegraphics[width=\textwidth]{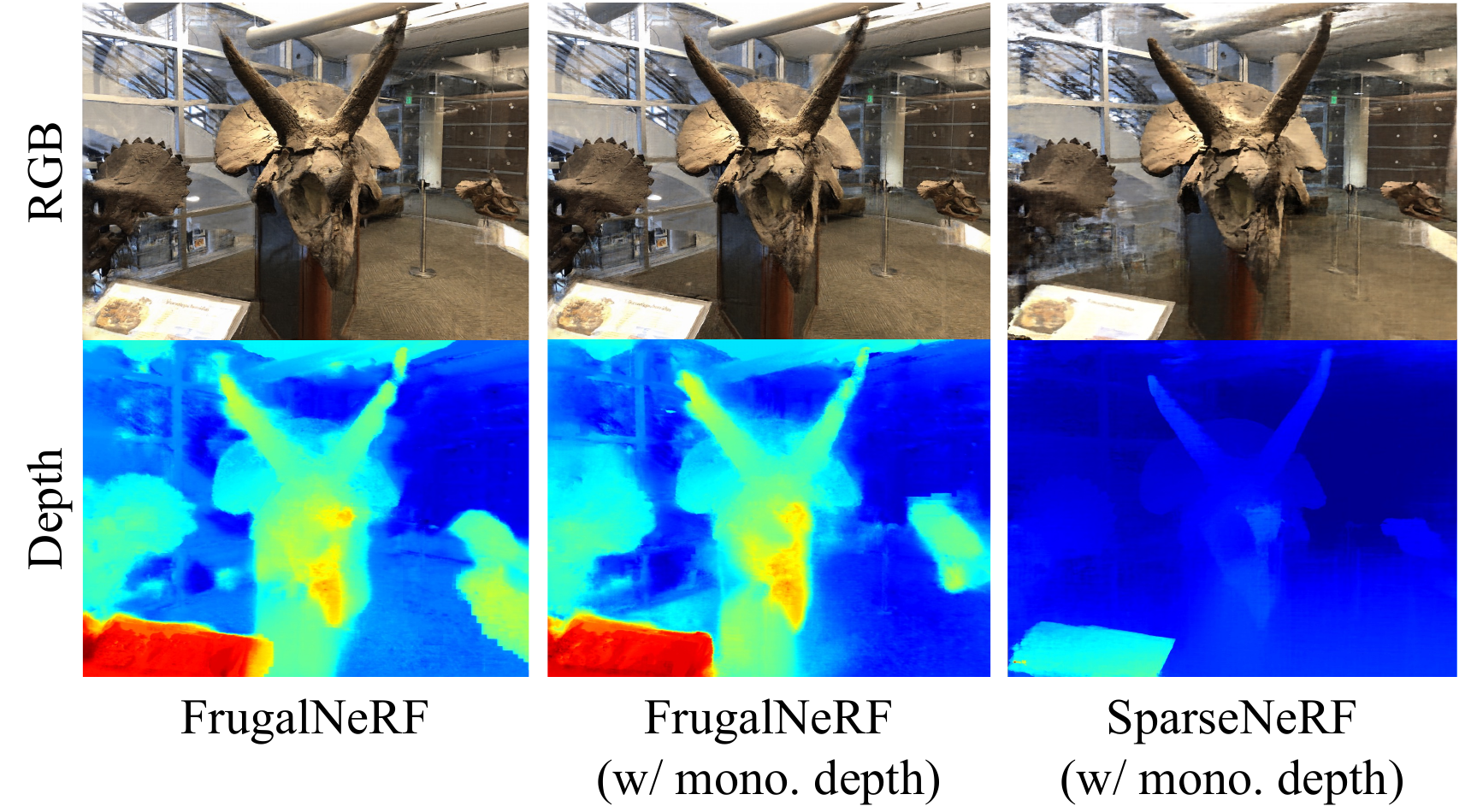}
}
\caption{\textbf{Visual comparisons on adding a pre-trained monocular depth prior.} 
}
\label{fig:mono_vis}
\end{figure}

\section{Additional experiments.}
\label{sec:additional_exp}

\subsection{Number of training views analysis.}
We plot the number of training views experiment in~\cref{fig:number_of_training_views_analysis} and \cref{tab:number_of_training_views_analysis}, demonstrating that FrugalNeRF outperforms TensoRF on sparse views (2 to 8 views) and continues to lead as the number of views increases.

\begin{figure}[t]
\includegraphics[width=\columnwidth]{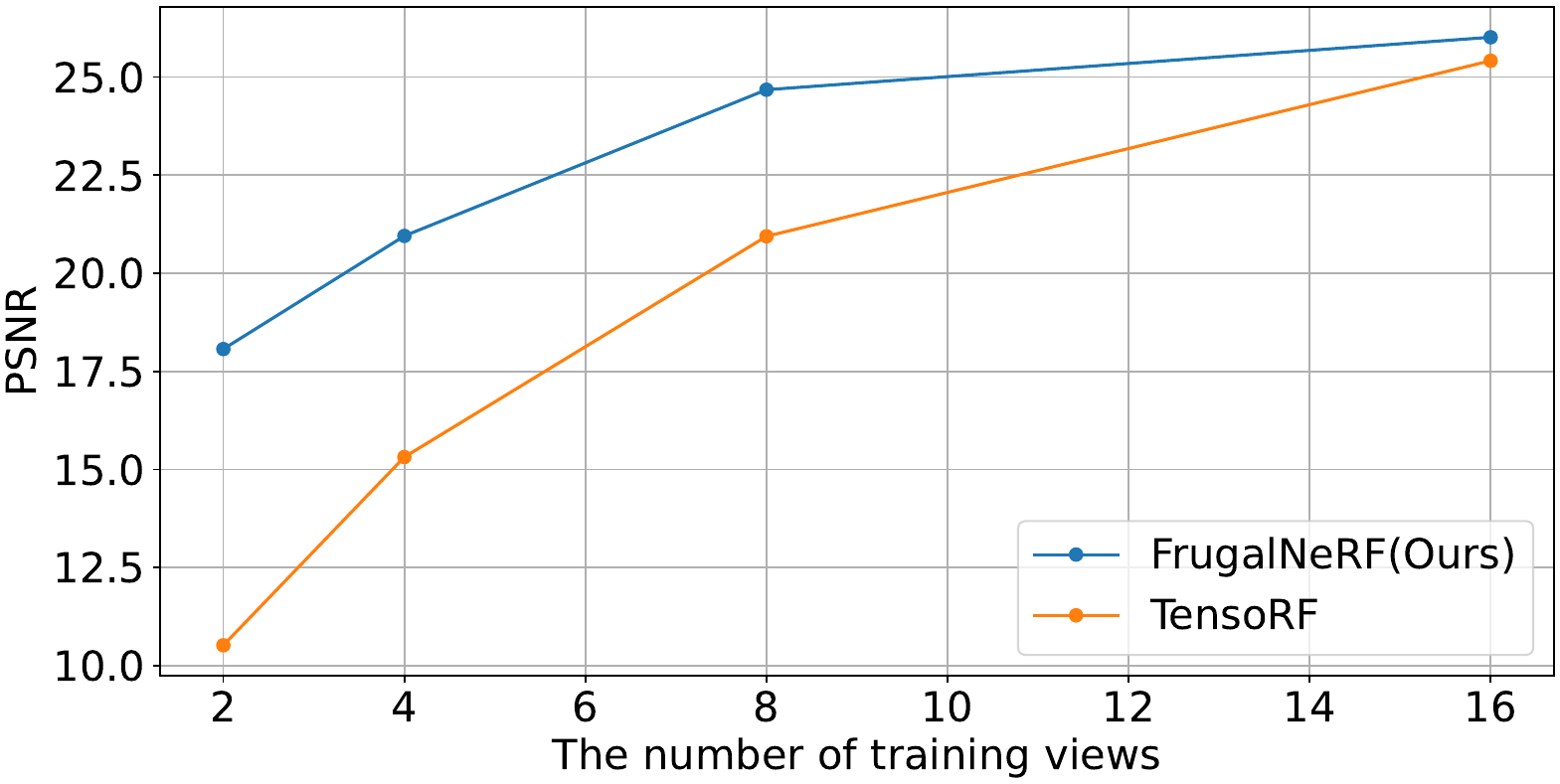}
\caption{\textbf{Number of training views analysis.} FrugalNeRF significantly outperforms the base TensoRF on sparse views.
}
\label{fig:number_of_training_views_analysis}
\end{figure}

\begin{table}[h]
\caption{\textbf{Number of training views analysis.}}
\vspace{-0.9pc}
\centering
\scriptsize
\resizebox{1.0\columnwidth}{!} {
\label{tab:number_of_training_views_analysis}
\begin{tabular}{l|ccc|ccc|ccc}
\toprule
 & \multicolumn{3}{c|}{2-view} & 
 \multicolumn{3}{c|}{16-view} &
 \multicolumn{3}{c}{Dense view (17- to 54-view)} \\
Method & PSNR $\uparrow$ & SSIM $\uparrow$ & LPIPS $\downarrow$ & PSNR $\uparrow$ & SSIM $\uparrow$ & LPIPS $\downarrow$ & PSNR $\uparrow$ & SSIM $\uparrow$ & LPIPS $\downarrow$ \\
\midrule
TensoRF & 11.97 & 0.32 & 0.64 & 25.42 & 0.813 & 0.151 & \textbf{26.73} & \textbf{0.839} & 0.124 \\
Ours & \textbf{18.07} & \textbf{0.54} & \textbf{0.35} & \textbf{25.82} & \textbf{0.822} & \textbf{0.128} & 26.69 & 0.835 & \textbf{0.114} \\
\bottomrule
\end{tabular}
}
\end{table}

\subsection{Downsampling Strategy.}
We use nearest-neighbor downsampling for stability and to prevent floaters. In ~\cref{tab:ablation_downsample} and ~\cref{fig:ablation_downsample} we compare it to bilinear interpolation.

\begin{table}[h]
\caption{\textbf{Ablation on downsampling strategy on LLFF dataset.}}
\centering
\scriptsize
\label{tab:ablation_downsample}
\begin{tabular}{lccc}
\toprule
Method & PSNR $\uparrow$ & SSIM $\uparrow$ & LPIPS $\downarrow$ \\
\midrule
Bilinear & 17.08 & 0.49 & 0.36 \\
Nearest-neighbor (ours) & \textbf{18.07} & \textbf{0.54} & \textbf{0.35} \\
\bottomrule
\end{tabular}
\vspace{-1pc}
\end{table}

\begin{table}[h]
\vspace{-0.7pc}
\centering
\setlength{\tabcolsep}{1pt}
\renewcommand{\arraystretch}{0.4}
\resizebox{1.0\columnwidth}{!} {
\begin{tabular}{cc}
\includegraphics[width=0.67\columnwidth]{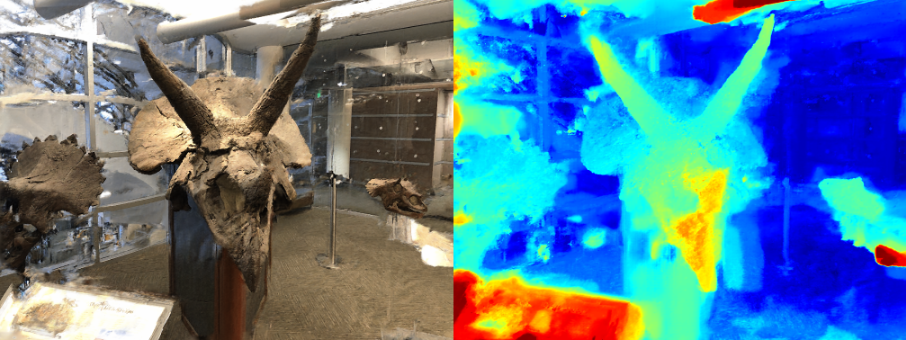} &
\includegraphics[width=0.67\columnwidth]{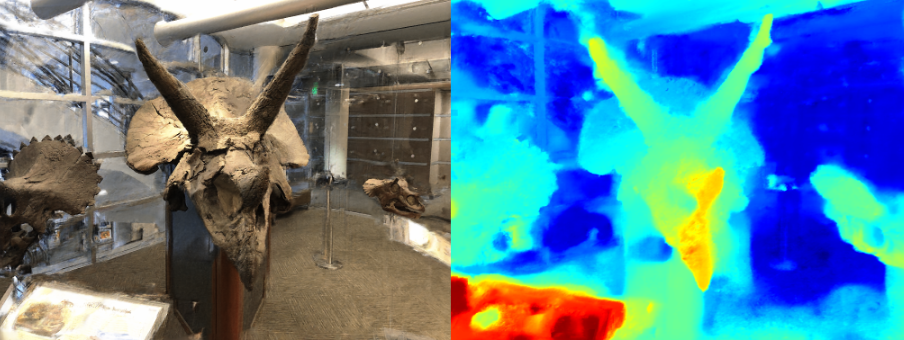} \\
Bilinear & Nearest-neighbor (ours) \\
\end{tabular}
}
\caption{\textbf{Visual comparison on different downsampling strategies .} 
}
\label{fig:ablation_downsample}
\end{table}

\section{Experimental Setup}
\label{sec:exp_setup}
We compare the result of few-shot NeRF on LLFF and DTU with $n=2, 3, 4$ input views.
\paragraph{LLFF dataset.} The LLFF dataset comprises 8 forward-facing unbounded scenes with variable frame counts at a resolution of 1008 × 756. In line with prior work~\citep{somraj2023vip}, we use every 8th frame for testing in each scene. For training, we uniformly sample 
$n$ views from the remaining frames.
\paragraph{DTU dataset.} The DTU dataset is a large-scale multi-view collection that includes 124 different scenes. Follow the Pixel-NeRF~\citep{yu2021pixelnerf} and ViP-NeRF~\citep{somraj2023vip} approach, we use the same test sets. However, because COLMAP will fail to generate sparse depth at scans 8, 30, and 110, we can only test on 12 scenes. Test scan IDs are 21, 31, 34, 38, 40, 41, 45, 55, 63, 82, 103, and 114. We use specific image IDs as input views and downsample images to 300 × 400 pixels for consistency with prior studies~\citep{yu2021pixelnerf, somraj2023vip}.
\paragraph{RealEstate-10K dataset.} RealEstate-10K is a comprehensive database of approximately 80,000 video segments, each with over 30 frames, widely utilized for novel view synthesis. For our study, we select five scenes from its extensive test set, following the approach outlined in ViP-NeRF~\citep{somraj2023vip}. We selected frames 0, 10, 20, and 30 for the training set with a resolution of 1024 × 576, in accordance with the SimpleNeRF~\citep{somraj2023simplenerf} methodology, while testing on the same test set as SimpleNeRF~\citep{somraj2023simplenerf} due to the unobserved region problem, which NeRF cannot handle, in some testing view. 


\subsection{Training Time Measurement and Time Complexity}
\paragraph{RegNeRF.} 
We use the official implementation of RegNeRF~\citep{niemeyer2022regnerf} and follow most of the default configuration, while the batch size or other hyperparameters might be adjusted due to the GPU memory issue. For the LLFF dataset, the training requires roughly 2.35 hours per scene with 69769 iterations and a batch size of 2,048. Note that RegNeRF samples 10000 random poses by its default configuration on the DTU dataset, leading to out-of-memory on a single NVIDIA RTX 4090 GPU. While reducing the number of random poses to about 1/8 could potentially resolve this issue, such a reduction is likely to adversely affect the performance, so we simply exclude this method from our experiments.

\paragraph{FreeNeRF.} 
We use the official implementation of FreeNeRF~\citep{yang2023freenerf} and follow most of the default configuration, while the batch size or other hyperparameters might be adjusted due to the GPU memory issue. For the LLFF dataset, the training requires roughly 1.5 hours per scene with 69,769 iterations and a batch size of 2,048. For the DTU dataset, the training requires about 1 hour per scene with 43,945 iterations and a batch size of 2,048.

\paragraph{SparseNeRF.} 
We use the official implementation of SparseNeRF.~\citep{wang2023sparsenerf} and follow most of the default configuration, while the batch size or other hyperparameters might be adjusted due to the GPU memory issue. For the LLFF dataset, the training requires roughly 1 hour per scene with 70,000 iterations and a batch size of 512. For the DTU dataset, the training requires about 30 minutes per scene with 70,000 iterations and a batch size of 256.
\paragraph{SimpleNeRF.} 
We use the official implementation of SimpleNeRF~\citep{somraj2023simplenerf} and follow most of the default configuration, while the batch size or other hyperparameters might be adjusted due to the GPU memory issue. For the LLFF dataset, we use the model weights released by the author directly. Since there's no official implemented dataloader for the DTU dataset, we use the dataloader and configuration from ViP-NeRF~\citep{somraj2023vip}, which requires about 1.38 hours per scene with 25,000 iterations and batch size of 2,048.

\paragraph{VGOS.} 
We furter provide VGOS result. We use the official implementation of VGOS~\citep{sun2023vgos} and follow most of the default configuration, while the batch size or other hyperparameters might be adjusted due to the GPU memory issue. Note that VGOS samples random poses directly from the entire dataset, which is unreasonable under the few-shot setting, so we replace the sampling with the interpolation from training poses implemented in the official repo.  For the LLFF dataset, the training requires roughly 5 minutes per scene with 9,000 iterations and a batch size of 16,384. For the DTU dataset, the training requires about 3 minutes per scene with 9,000 iterations and a batch size of 16,384. Note that VGOS seems invalid on the DTU dataset (\cref{fig:supp_dtu_more_visual}) and they does not evaluate the DTU dataset in their paper.

\paragraph{GeCoNeRF.}
As mentioned in GeCoNeRF~\citep{kwak2023geconerf}'s 
official github repo, their current code is unexecutable. To complete our experiment, we still try our best to implement their method based on the code provided. For the LLFF dataset, the training requires roughly 4 hours per scene with 85,000 iterations and a batch size of 1024. It is important to note that we utilized 2 GPUs for training this method, so the training time reported in our paper might be shorter than what is actually required. 
\paragraph{ZeroRF.}
We use the official implementation of ZeroRF~\citep{shi2023zerorf} and follow most of the default configurations. For the LLFF dataset, ZeroRF does not provide the dataloader for the LLFF, and their paper mentions its inability to be used for unbounded scenes. Therefore, our primary testing was conducted on the DTU dataset. In the DTU dataset, the original implementation of ZeroRF necessitates masking out the background area of the input frame before training, which is incompatible with our evaluation benchmark. Consequently, we trained it without object masks. Training requires approximately 25 minutes per scene with 10,000 iterations and a batch size of $2^{14}$.

\paragraph{FSGS.} We use the official implementation of FSGS~\citep{zhu2023FSGS} and follow most of the default configurations. For the LLFF dataset, we adjust the input views to match the settings used in ViP-NeRF, which differs from the original FSGS paper. Training takes approximately 25 minutes per scene with 10,000 iterations. Since there is no official dataloader for the DTU dataset, we convert the DTU camera poses to the LLFF format and use the default LLFF configuration. Training on the DTU dataset requires around 20 minutes per scene with 10,000 iterations.

\paragraph{Time complexity.}
To verify the efficiency of our method, besides comparing the training time of various methods, we also calculated the MFLOPs per pixel in ~\cref{tab:ablation_timecomplexity}.
\begin{table}[t]
\centering
\small
\caption{\textbf{Comparison of the time complexity.}}
\label{tab:ablation_timecomplexity}
\begin{tabular}{l|r}
\toprule
Method & MFLOPs / pixel $\downarrow$ \\
\midrule
FreeNeRF~\citep{yang2023freenerf} & 288.57 \\
ViP-NeRF~\citep{somraj2023vip} & 149.26 \\
SimpleNeRF~\citep{somraj2023simplenerf} & 303.82 \\
SparseNeRF~\citep{wang2023sparsenerf} & 287.92 \\
Ours & 13.77 \\
\bottomrule
\end{tabular}%
\end{table}


\section{Complete Quantitative Evaluations}
\label{sec:full_quantitative}
\paragraph{LLFF dataset.}
We show all 8 scenes of the quantitative comparisons with two, three, and four input views on the LLFF dataset in~\cref{tab:suppl_LLFF},~\cref{tab:suppl_LLFF_3v}, and~\cref{tab:suppl_LLFF_4v}, respectively.

\newcolumntype{C}[1]{>{\centering\arraybackslash}p{#1}}

\begin{table*}[t]

\centering
\caption{\textbf{Quantitative results on the LLFF~\citep{mildenhall2019llff} dataset with two input views. The three rows show LPIPS, SSIM, and PSNR scores, respectively.}}
\label{tab:suppl_LLFF}
\resizebox{\textwidth}{!}{%
\begin{tabular}{l|*{9}{C{1.2cm}}}
\toprule
\multicolumn{1}{r|}{Scene} & Fern & Flower & Fortress & Horns & Leaves & Orchids & Room & Trex & Average\\
\multicolumn{1}{l|}{Method} & & & & & & & & & \\
\midrule
& 0.51 & 0.43 & 0.37 & 0.51 & 0.35 & 0.45 & 0.38 & 0.42 & 0.43 \\
RegNeRF~\citep{niemeyer2022regnerf}& 0.45 & 0.51 & 0.46 & 0.42 & 0.37 & 0.30 & 0.74 & 0.54 & 0.49 \\
& 15.8 & 17.0 & 20.6 & 15.9 & 14.5 & 13.9 & 18.7 & 16.7 & 16.9 \\
\midrule
& 0.50 & 0.43 & 0.30 & 0.49 & 0.47 & 0.43 & 0.35 & 0.41 & 0.42 \\
DS-NeRF~\citep{deng2022depth}& 0.46 & 0.44 & 0.65 & 0.49 & 0.24 & 0.32 & 0.76 & 0.53 & 0.51 \\
& 16.4 & 16.1 & 23.0 & 16.6 & 12.4 & 13.7 & 18.9 & 15.7 & 16.9 \\
\midrule
& 0.44 & 0.46 & 0.17 & 0.46 & 0.52 & 0.41 & 0.30 & 0.43 & 0.39 \\
DDP-NeRF~\citep{roessle2022dense}& 0.49 & 0.45 & 0.77 & 0.52 & 0.23 & 0.38 & 0.76 & 0.54 & 0.54 \\
& 17.2 & 16.2 & 22.7 & 17.1 & 12.6 & 15.1 & 18.7 & 15.7 & 17.2 \\
\midrule
& 0.46 & 0.38 & 0.33 & 0.43 & 0.36 & 0.42 & 0.34 & 0.33 & 0.38 \\
FreeNeRF~\citep{yang2023freenerf}& 0.49 & 0.55 & 0.53 & 0.53 & 0.38 & 0.35 & 0.76 & 0.60 & 0.54 \\
& 17.1 & 17.6 & 21.3 & 17.1 & 14.4 & 14.1 & 18.3 & 18.1 & 17.6 \\
\midrule
& 0.45 & 0.42 & 0.21 & 0.39 & 0.46 & 0.40 & 0.36 & 0.38 & 0.37 \\
ViP-NeRF~\citep{somraj2023vip}& 0.45 & 0.43 & 0.71 & 0.54 & 0.21 & 0.36 & 0.72 & 0.54 & 0.52 \\
& 16.2 & 14.9 & 22.6 & 17.1 & 11.7 & 14.2 & 17.7 & 15.9 & 16.7\\
\midrule
& 0.51 & 0.43 & 0.25 & 0.42 & 0.44 & 0.41 & 0.35 & 0.39 & 0.39 \\
SimpleNeRF~\citep{somraj2023simplenerf}& 0.50 & 0.53 & 0.67 & 0.54 & 0.30 & 0.37 & 0.77 & 0.58 & 0.55 \\
& 17.0 & 16.9 & 22.5 & 17.1 & 13.5 & 14.7 & 19.5 & 16.8 & 17.6 \\
\midrule
& 0.48&	0.44&	0.37&	0.47&	0.36&	0.42&	0.38&	0.40&	0.42 \\ 
VGOS~\citep{sun2023vgos}& 0.51&	0.55&	0.53&	0.55&	0.38&	0.40&	0.77&	0.59&	0.55 \\
& 
16.5&	17.5&	19.4&	15.7&	14.7&	14.4&	18.8&	16.0&	16.7 \\
\midrule
& 0.56 & 0.49 & 0.50 & 0.61 & 0.49 & 0.51 & 0.54 & 0.49 & 0.52 \\
GeCoNeRF~\citep{kwak2023geconerf} & 0.47 & 0.49 & 0.43 & 0.41 & 0.28 & 0.29 & 0.68 & 0.52 & 0.45 \\ 
& 16.4 & 16.9 & 17.9 & 15.4 & 13.3 & 13.4 & 17.3 & 16.1 & 15.8 \\ 
\midrule
& 0.48 & 0.55 & 0.40 & 0.52 & 0.52 & 0.55 & 0.29 & 0.37 & 0.45 \\
SparseNeRF~\citep{wang2023sparsenerf}& 0.52 & 0.41 & 0.61 & 0.51 & 0.244 & 0.24 & 0.82 & 0.62 & 0.52\\
& 18.2 & 15.4 & 21.7 & 17.4 & 13.4 & 13.3 & 22.8& 18.6 & 18.0\\
\midrule
& 0.46 & 0.45 & 0.35 & 0.42 & 0.33 & 0.41 & 0.38 & 0.45 & 0.41 \\
FSGS~\citep{zhu2023FSGS} & 0.40 & 0.38 & 0.47 & 0.42 & 0.34 & 0.24 & 0.72 & 0.46 & 0.45 \\
& 15.0 & 14.8 & 16.9 & 16.2 & 14.2 & 12.6 & 17.6 & 13.8 & 15.3 \\
\midrule
& 0.41 & 0.41 & 0.27 & 0.36 & 0.32 & 0.42 & 0.34 & 0.32 & 0.35 \\
FrugalNeRF (Ours) & 0.47&	0.50&	0.54&	0.55&	0.41&	0.33&	0.75&	0.61&	0.54 \\
& 17.4&	17.5&	20.3&	18.5&	15.5&	15.0&	19.2&	18.6&	18.1 \\
\midrule
& 0.40 & 0.40 & 0.27 & 0.37 & 0.33 & 0.39 & 0.32 & 0.35 & 0.35 \\
FrugalNeRF w/ mono. depth (Ours) & 0.46&	0.53&	0.54&	0.54&	0.41&	0.37&	0.76&	0.59&	0.54 \\
& 17.7&	17.9&	20.9&	18.5&	15.4&	15.6&	19.6&	18.2&	18.3 \\
\bottomrule
\end{tabular}

}
\end{table*}

\newcolumntype{C}[1]{>{\centering\arraybackslash}p{#1}}

\begin{table*}[t]
\centering
\caption{\textbf{Quantitative results on the LLFF~\citep{mildenhall2019llff} dataset with three input views. The three rows show LPIPS, SSIM, and PSNR scores, respectively.}}
\label{tab:suppl_LLFF_3v}
\resizebox{\textwidth}{!}{%
\begin{tabular}{l|*{9}{C{1.2cm}}}
\toprule
\multicolumn{1}{r|}{Scene} & Fern & Flower & Fortress & Horns & Leaves & Orchids & Room & Trex & Average\\
\multicolumn{1}{l|}{Method} & & & & & & & & & \\
\midrule
& 0.47 & 0.27 & 0.31 & 0.44 & 0.39 & 0.44 & 0.25 & 0.36 & 0.36 \\
RegNeRF~\citep{niemeyer2022regnerf}& 0.48 & 0.58 & 0.64 & 0.53 & 0.37 & 0.31 & 0.81 & 0.63 & 0.57 \\
& 17.9 & 19.6 & 22.7 & 18.2 & 14.6 & 14.2 & 21.0 & 18.4 & 18.7 \\
\midrule
& 0.47 & 0.25 & 0.25 & 0.47 & 0.50 & 0.45 & 0.22 & 0.37 & 0.36 \\
DS-NeRF~\citep{deng2022depth}& 0.52 & 0.66 & 0.72 & 0.52 & 0.25 & 0.33 & 0.84 & 0.59 & 0.58 \\
& 18.5 & 21.3 & 24.8 & 17.5 & 12.6 & 14.1 & 23.0 & 17.1 & 19.0 \\
\midrule
& 0.47 & 0.29 & 0.20 & 0.48 & 0.52 & 0.45 & 0.32 & 0.42 & 0.39 \\
DDP-NeRF~\citep{roessle2022dense}& 0.53 & 0.63 & 0.75 & 0.53 & 0.24 & 0.35 & 0.76 & 0.54 & 0.56 \\
& 18.5 & 20.2 & 22.1 & 17.4 & 12.8 & 15.1 & 18.3 & 16.0 & 17.7 \\
\midrule
& 0.40 & 0.28 & 0.32 & 0.41 & 0.40 & 0.41 & 0.22 & 0.33 & 0.34\\
FreeNeRF~\citep{yang2023freenerf}& 0.54 & 0.61 & 0.60 & 0.58 & 0.40 & 0.37 & 0.85 & 0.64 & 0.60 \\
& 18.9 & 20.7 & 22.0 & 18.7 & 15.0 & 14.7 & 22.6 & 19.0 & 19.3 \\
\midrule
& 0.51 & 0.24 & 0.19 & 0.42 & 0.44 & 0.41 & 0.27 & 0.32 & 0.34 \\
ViP-NeRF~\citep{somraj2023vip}& 0.49 & 0.65 & 0.76 & 0.57 & 0.25 & 0.34 & 0.81 & 0.62 & 0.59 \\
& 17.3 & 20.8 & 24.5 & 18.2 & 12.4 & 14.2 & 21.7 & 18.1 & 18.9\\
\midrule
& 0.43 & 0.24 & 0.17 & 0.42 & 0.42 & 0.39 & 0.26 & 0.34 & 0.33 \\
SimpleNeRF~\citep{somraj2023simplenerf}& 0.52 & 0.66 & 0.78 & 0.57 & 0.38 & 0.38 & 0.83 & 0.66 & 0.62 \\
& 18.2 & 20.7 & 24.7 & 18.4 & 14.8 & 15.0 & 22.0 & 18.9 & 19.5 \\
\midrule
& 0.40&	0.31&	0.33&	0.46&	0.40&	0.41&	0.31&	0.35&	0.37 \\ 
VGOS~\citep{sun2023vgos}& 0.58&	0.61&	0.69&	0.58&	0.40&	0.40&	0.83&	0.66&	0.61 \\
&
19.0&	20.0&	23.0&	17.0&	15.0&	15.2&	21.8&	18.0&	18.8 \\
\midrule
& 0.57 & 0.36 & 0.45 & 0.60 & 0.50 & 0.51 & 0.34 & 0.43 & 0.47 \\
GeCoNeRF~\citep{kwak2023geconerf} & 0.46 & 0.57 & 0.53 & 0.44 & 0.32 & 0.30 & 0.80 & 0.59 & 0.50 \\ 
& 17.0 & 19.5 & 20.6 & 15.8 & 13.8 & 13.6 & 21.1 & 18.1 & 17.4 \\ 
\midrule
& 0.43&	0.33&	0.37&	0.50&	0.35&	0.41&	0.28&	0.31&	0.37 \\
SparseNeRF~\citep{wang2023sparsenerf} & 0.57&	0.60&	0.59&	0.53&	0.45&	0.37&	0.81&	0.67&	0.59 \\
& 19.6&	19.8&	23.0&	18.4&	16.5&	15.2&	21.5&	20.1&	19.5 \\
\midrule
& 0.48 & 0.30 & 0.15 & 0.36 & 0.26 & 0.35 & 0.28 & 0.28 & 0.30 \\
FSGS~\citep{zhu2023FSGS}& 0.55 & 0.68 & 0.72 & 0.65 & 0.28 & 0.37 & 0.84 & 0.62 & 0.61 \\
& 17.9 & 21.5 & 23.9 & 19.4 & 13.3 & 14.1 & 22.6 & 17.4 & 19.2 \\
\midrule
& 0.39&	0.32&	0.24&	0.34&	0.37&	0.42&	0.27&	0.29 &	0.32 \\
FrugalNeRF (Ours) & 0.50&	0.55&	0.63&	0.59&	0.39&	0.35&	0.81&	0.66&	0.59 \\
& 18.2&	18.8&	23.4&	19.3&	15.5&	15.3&	22.2&	19.3&19.4 \\
\midrule
& 0.40 & 0.23 & 0.22 & 0.33 & 0.37 & 0.40 & 0.25 & 0.29 & 0.30 \\
FrugalNeRF w/ mono. depth (Ours) & 0.49&	0.63&	0.69&	0.60&	0.39&	0.36&	0.83&	0.67&	0.61 \\
& 18.6&	21.4&	23.5&	19.0&	15.4&	15.7&	22.3&	20.0&	19.9 \\
 \bottomrule
\end{tabular}

}
\end{table*}

\newcolumntype{C}[1]{>{\centering\arraybackslash}p{#1}}

\begin{table*}[t]
\centering
\caption{\textbf{Quantitative results on the LLFF~\citep{mildenhall2019llff} dataset with four input views. The three rows show LPIPS, SSIM, and PSNR scores, respectively.}}
\label{tab:suppl_LLFF_4v}
\resizebox{\textwidth}{!}{%
\begin{tabular}{l|*{9}{C{1.2cm}}}
\toprule
\multicolumn{1}{r|}{Scene} & Fern & Flower & Fortress & Horns & Leaves & Orchids & Room & Trex & Average\\
\multicolumn{1}{l|}{Method} & & & & & & & & & \\
\midrule
& 0.35 & 0.29 & 0.37 & 0.34 & 0.32 & 0.43 & 0.19 & 0.32 & 0.32 \\
RegNeRF~\citep{niemeyer2022regnerf} & 0.63 & 0.64 & 0.55 & 0.64 & 0.44 & 0.34 & 0.87 & 0.66 & 0.62 \\
& 20.8 & 19.8 & 22.4 & 20.1 & 15.9 & 14.8 & 23.9 & 18.9 & 19.9 \\
\midrule
& 0.35 & 0.28 & 0.31 & 0.41 & 0.41 & 0.41 & 0.16 & 0.39 & 0.34 \\
DS-NeRF~\citep{deng2022depth} & 0.63 & 0.64 & 0.66 & 0.59 & 0.39 & 0.38 & 0.89 & 0.59 & 0.61 \\
& 20.9 & 20.6 & 24.1 & 19.5 & 15.8 & 15.2 & 25.6 & 17.1 & 20.1 \\
\midrule
& 0.40 & 0.30 & 0.18 & 0.42 & 0.45 & 0.42 & 0.26 & 0.39 & 0.35 \\
DDP-NeRF~\citep{roessle2022dense} & 0.60 & 0.63 & 0.73 & 0.59 & 0.37 & 0.41 & 0.82 & 0.60 & 0.61 \\
& 20.1 & 20.0 & 23.4 & 19.3 & 15.1 & 15.8 & 20.8 & 17.3 & 19.2 \\
\midrule
& 0.37 & 0.30 & 0.35 & 0.37 & 0.35 & 0.42 & 0.19 & 0.31 & 0.33 \\
FreeNeRF~\citep{yang2023freenerf} & 0.64 & 0.64 & 0.60 & 0.63 & 0.47 & 0.37 & 0.88 & 0.68 & 0.63 \\
& 21.1 & 20.5 & 23.2 & 20.4 & 16.6 & 14.9 & 24.8 & 19.6 & 20.5 \\
\midrule
& 0.39 & 0.27 & 0.25 & 0.38 & 0.36 & 0.40 & 0.23 & 0.32 & 0.32 \\
ViP-NeRF~\citep{somraj2023vip} & 0.58 & 0.63 & 0.70 & 0.60 & 0.40 & 0.39 & 0.85 & 0.64 & 0.62 \\
& 18.2 & 19.5 & 23.3 & 19.0 & 14.8 & 14.8 & 23.2 & 18.6 & 19.3 \\
\midrule
& 0.33 & 0.27 & 0.28 & 0.38 & 0.35 & 0.36 & 0.19 & 0.32 & 0.31 \\
SimpleNeRF~\citep{somraj2023simplenerf} & 0.65 & 0.67 & 0.69 & 0.63 & 0.46 & 0.42 & 0.88 & 0.68 & 0.65 \\
& 21.1 & 20.8 & 24.3 & 19.7 & 16.3 & 15.7 & 24.3 & 19.3 & 20.4 \\
\midrule
& 0.40&	0.35&	0.40&	0.43&	0.34&	0.41&	0.28&	0.35&	0.37 \\
VGOS~\citep{sun2023vgos} & 0.64&	0.63&	0.64&	0.62&	0.49&	0.43&	0.86&	0.68&	0.64 \\
&
19.6&	20.3&	22.7&	18.6&	16.6&	15.8&	23.6&	18.7&	19.7 \\
\midrule
& 0.45 & 0.36 & 0.44 & 0.47 & 0.44 & 0.51 & 0.27 & 0.40 & 0.42 \\ 
GeCoNeRF~\citep{kwak2023geconerf} & 0.61 & 0.61 & 0.51 & 0.59 & 0.40 & 0.30 & 0.85 & 0.63 & 0.56 \\ 
& 20.5 & 19.9 & 21.2 & 19.6 & 15.5 & 13.9 & 23.5 & 19.0 & 19.1 \\ 
\midrule
& 0.42&	0.32&	0.31&	0.39&	0.36&	0.42&	0.25&	0.29&	0.34 \\
SparseNeRF~\citep{wang2023sparsenerf} & 0.62&	0.64&	0.70&	0.63&	0.49&	0.39&	0.85&	0.70&	0.65 \\
& 21.4&	20.7&	24.6&	20.4&	17.5&	15.7&	23.5&	20.9&	20.9 \\
\midrule
& 0.26 & 0.22 & 0.17 & 0.24 & 0.22 & 0.28 & 0.17 & 0.23 & 0.22 \\
FSGS~\citep{zhu2023FSGS}& 0.67 & 0.65 & 0.65 & 0.70 & 0.46 & 0.45 & 0.88 & 0.71 & 0.66 \\
& 20.5 & 20.2 & 22.6 & 20.9 & 15.6 & 15.4 & 23.7 & 19.2 & 20.1 \\
\midrule
& 0.30&	0.28&	0.24&	0.30&	0.26&	0.38&	0.19&	0.27&	0.27 \\
FrugalNeRF (Ours) & 0.63&	0.64&	0.60&	0.66&	0.52&	0.41&	0.87&	0.72&	0.65 \\
& 21.1&	20.8&	23.6&	21.6&	16.9&	16.3&	24.2&	19.7&	20.9 \\
\midrule
& 0.30 & 0.27 & 0.25 & 0.28 & 0.24 & 0.37 & 0.18 & 0.27 & 0.26 \\
FrugalNeRF w/ mono. depth (Ours) & 0.64 & 0.65 & 0.64 & 0.68 & 0.53 & 0.41 & 0.88 & 0.71 & 0.66 \\
& 21.5 & 20.9 & 23.9 & 21.1 & 17.2 & 16.3 & 24.1 & 19.6 & 20.9 \\
 \bottomrule
\end{tabular}
}
\end{table*}

\paragraph{DTU dataset.}
We show all 12 scenes of the quantitative comparisons with two, three, and four input views on the DTU dataset in~\cref{tab:quantitative_dtu},~\cref{tab:suppl_DTU},~\cref{tab:suppl_DTU_3v}, and~\cref{tab:suppl_DTU_4v}, respectively.

\begin{table*}[t]
\centering
\caption{\textbf{Quantitative results on the DTU ~\cite{jensen2014large} dataset.} 
FurgalNeRF synthesizes better images than most of the other baselines under extreme few-shot settings but with shorter training time and does not rely on any externally learned priors. Additionally, integrating monocular depth model regularization further improves quality while maintaining fast convergence. We follow SparseNeRF~\cite{wang2023sparsenerf} to remove the background when computing metrics.
}
\label{tab:quantitative_dtu}
\resizebox{\textwidth}{!}{%
\begin{tabular}{l|c|c|ccc|ccc|ccc|c}
\toprule
 & & Learned & \multicolumn{3}{c|}{2-view} & \multicolumn{3}{c|}{3-view} & \multicolumn{3}{c|}{4-view} & Training \\
Method & Venue & priors & PSNR $\uparrow$ & SSIM $\uparrow$ & LPIPS $\downarrow$ & PSNR $\uparrow$ & SSIM $\uparrow$ & LPIPS $\downarrow$ & PSNR $\uparrow$ & SSIM $\uparrow$ & LPIPS $\downarrow$ & time $\downarrow$\\
\midrule
TensoRF~\cite{chen2022tensorf} & ECCV22 & - & 8.81 & 0.34 & 0.71 & 9.13 & 0.34 & 0.9.11 & 9.15 & 0.33 & 0.71  & 5 mins \\
\midrule

FreeNeRF~\cite{yang2023freenerf} & CVPR23 & - & \cellcolor{orange!25}18.05 & \cellcolor{orange!25}0.73 & \cellcolor{orange!25}0.22 & \cellcolor{orange!25}22.40 & \cellcolor{orange!25}0.82 & \cellcolor{red!25}0.14 & \cellcolor{red!25}24.98 & \cellcolor{red!25}0.86 & \cellcolor{red!25}0.12  & 1 hrs \\
ViP-NeRF~\cite{somraj2023vip} & SIGGRAPH23 & - & 14.91 & 0.49 & 0.24 & 16.62 & 0.55 & \cellcolor{orange!25}0.22 & 17.64 & 0.57 & \cellcolor{orange!25}0.21 & 2.2 hrs \\
SimpleNeRF~\cite{somraj2023simplenerf} & SIGGRAPH Asia23 & - & 14.41 & 0.79 & 0.25 & 14.01 & 0.77 & 0.25 & 13.90 & \cellcolor{orange!25}0.78 & 0.26 & 1.38 hrs \\
ZeroRF~\cite{shi2023zerorf} & CVPR24 & - & 
14.84 & 0.60 & 0.30 & 
14.47 & 0.61 & 0.31 &
15.73 & 0.67 & 0.28 & 25 mins
\\ 
FrugalNeRF (Ours) & - & - & \cellcolor{red!25}19.72 & \cellcolor{red!25}0.78 & \cellcolor{red!25}0.16 & \cellcolor{red!25}22.43 & \cellcolor{red!25}0.83 & \cellcolor{red!25}0.14 & \cellcolor{orange!25}24.51 & \cellcolor{red!25}0.86 & \cellcolor{red!25}0.12 & \cellcolor{red!25}6 mins \\
\midrule

SparseNeRF~\cite{wang2023sparsenerf} & ICCV23 & monocular depth & \cellcolor{orange!25}19.83 & \cellcolor{orange!25}0.75 & \cellcolor{orange!25}0.20 & \cellcolor{orange!25}22.47 & \cellcolor{red!25}0.83 & \cellcolor{orange!25}0.14 & \cellcolor{orange!25}24.03 & \cellcolor{red!25}0.86 & \cellcolor{red!25}0.12 & 30 mins \\
FSGS~\cite{zhu2023FSGS} & ECCV24 & monocular depth & 16.82 & 0.64 & 0.27 & 18.29 & \cellcolor{orange!25}0.69 & 0.21 & 20.08 & \cellcolor{orange!25}0.75 & \cellcolor{orange!25}0.16 & 20 mins \\
FrugalNeRF (Ours) & - & monocular depth & \cellcolor{red!25}20.77 & \cellcolor{red!25}0.79 & \cellcolor{red!25}0.15  & \cellcolor{red!25}22.84 & \cellcolor{red!25}0.83 & \cellcolor{red!25}0.13 & \cellcolor{red!25}24.81 & \cellcolor{red!25}0.86 & \cellcolor{red!25}0.12 & \cellcolor{red!25}7 mins\\
\bottomrule
\end{tabular}%
}
\end{table*}


\begin{table*}[t]
\centering
\caption{\textbf{Quantitative results on the DTU~\cite{jensen2014large} dataset with two input views. The three rows show LPIPS, SSIM and PSNR scores, respectively.}}
\label{tab:suppl_DTU}
\resizebox{\textwidth}{!}{%
\begin{tabular}{l|ccccccccccccc}
\toprule
\multicolumn{1}{r|}{Scene} & Scan21 & Scan31 & Scan34 & Scan38 & Scan40 & Scan41 & Scan45 & Scan55 & Scan63 & Scan82 & Scan103 & Scan114 & Average\\
\multicolumn{1}{l|}{Method} & & & & & & & & & & & & & \\
\midrule
& 0.33 & 0.18 & 0.31 & 0.34 & 0.41 & 0.35 & 0.19 & 0.11 & 0.07 & 0.08 & 0.17 & 0.12 & 0.22 \\
FreeNeRF~\cite{yang2023freenerf}& 0.51 & 0.75 & 0.63 & 0.61 & 0.58 & 0.63 & 0.76 & 0.80 & 0.93 & 0.90 & 0.82 & 0.85 & 0.73 \\
& 13.21 & 19.33 & 14.66 & 16.76 & 11.42 & 14.50 & 18.66 & 21.62 & 23.19 & 21.56 & 17.55 & 24.19 & 18.05 \\
\midrule
& 0.37 & 0.24	& 0.27 & 0.38 & 0.31 & 0.23	& 0.31 & 0.21 & 0.09 & 0.12 & 0.18 & 0.17 & 0.24 \\
ViP-NeRF~\cite{somraj2023vip}& 0.26	& 0.49 & 0.52 & 0.43 & 0.47 & 0.58 & 0.37 & 0.39 & 0.63	& 0.57 & 0.65 & 0.49 & 0.49 \\
& 11.31 & 13.57 & 17.13 & 13.25 & 15.08 & 17.81 & 11.35 & 16.92 & 16.71 & 13.37 & 16.15 & 16.24 & 14.91\\
\midrule
& 0.23 & 0.32 & 0.23 & 0.21 & 0.24 & 0.19 & 0.28 & 0.22 & 0.30 & 0.27 & 0.19 & 0.27 & 0.25 \\
SimpleNeRF~\cite{somraj2023simplenerf} & 0.73 & 0.71 & 0.76 & 0.77 & 0.77 & 0.84 & 0.70 & 0.88 & 0.75 & 0.79 & 0.81 & 0.82 & 0.79 \\
& 12.71 & 11.91 & 14.39 & 14.50 & 13.76 & 15.57 & 11.88 & 19.58 & 12.73 & 14.37 & 16.64 & 14.86 & 14.41 \\
\midrule
& 0.28 & 0.36 & 0.33 & 0.31 & 0.30 & 0.27 & 0.37 & 0.15 & 0.49 & 0.45 & 0.34 & 0.18 & 0.32 \\ 
VGOS~\cite{sun2023vgos}& 0.69 & 0.67 & 0.69 & 0.71 & 0.73 & 0.78 & 0.64 & 0.90 & 0.56 & 0.57 & 0.73 & 0.85 & 0.71 \\
& 9.69 & 8.97 & 9.75 & 10.27 & 8.79 & 9.75 & 7.54 & 19.24 & 5.17 & 5.63 & 11.29 & 15.81 & 10.16 \\
\midrule
& 0.39 & 0.22 & 0.26 & 0.33 & 0.24 & 0.21 & 0.20 & 0.14 & 0.08 & 0.08 & 0.15 & 0.13 & 0.20 \\
SparseNeRF~\cite{wang2023sparsenerf}& 0.45 & 0.69 & 0.70 & 0.60 & 0.72 & 0.76 & 0.75 & 0.78 & 0.92 & 0.91 & 0.84 & 0.85 & 0.75 \\
& 14.25 & 17.95 & 20.65 & 17.93 & 16.33 & 20.13 & 18.22 & 22.29 & 20.70 & 23.46 & 21.70 & 24.40 & 19.83 \\
\midrule
& 0.45 & 0.27 & 0.35 & 0.44 & 0.29 & 0.28 & 0.39 & 0.25 & 0.13 & 0.18 & 0.25 & 0.29 & 0.30 \\
ZeroRF~\cite{shi2023zerorf} & 0.30 & 0.61 & 0.50 & 0.39 & 0.59 & 0.63 & 0.49 & 0.68 & 0.88 & 0.82 & 0.73 & 0.63 & 0.60 \\
& 10.99 & 14.40 & 13.93 & 12.16 & 15.41 & 16.73 & 11.24 & 17.08 & 20.39 & 15.36 & 16.23 & 14.12 & 14.84 \\
\midrule
& 0.25 & 0.16 & 0.20 & 0.24 & 0.24 & 0.17 & 0.16 & 0.13 & 0.09 & 0.07 & 0.13 & 0.11 & 0.16 \\
FrugalNeRF (Ours) & 0.57 & 0.73 & 0.73 & 0.64 & 0.73 & 0.78 & 0.77 & 0.86 & 0.92 & 0.92 & 0.85 & 0.89 & 0.78 \\
& 14.67 & 17.86 & 19.47 & 17.66 & 14.51 & 19.74 & 16.94 & 24.87 & 21.21 & 22.67 & 21.45 & 25.60 & 19.72 \\
\midrule
& 0.25 & 0.15 & 0.19 & 0.21 & 0.23 & 0.16 & 0.15 & 0.12 & 0.08 & 0.07 & 0.10 & 0.10 & 0.15 \\
FrugalNeRF w/ mono. depth (Ours) & 0.56 & 0.73 & 0.75 & 0.68 & 0.74 & 0.79 & 0.78 & 0.86 & 0.93 & 0.91 & 0.88 & 0.90 & 0.79 \\
& 14.14 & 18.46 & 21.27 & 19.40 & 15.56 & 20.53 & 18.05 & 25.65 & 23.46 & 22.72 & 23.76 & 26.25 & 20.77 \\
 \bottomrule
\end{tabular}
}
\end{table*}
\begin{table*}[t]
\centering
\caption{\textbf{Quantitative results on the DTU~\citep{jensen2014large} dataset with three input views. The three rows show LPIPS, SSIM and PSNR scores, respectively.}}
\label{tab:suppl_DTU_3v}
\resizebox{\textwidth}{!}{%
\begin{tabular}{l|ccccccccccccc}
\toprule
\multicolumn{1}{r|}{Scene} & Scan21 & Scan31 & Scan34 & Scan38 & Scan40 & Scan41 & Scan45 & Scan55 & Scan63 & Scan82 & Scan103 & Scan114 & Average\\
\multicolumn{1}{l|}{Method} & & & & & & & & & & & & & \\
\midrule
& 15.93 & 19.53 & 23.23 & 19.88 & 18.38 & 22.83 & 21.07 & 22.88 & 25.28 & 26.39 & 26.68 & 26.68 & 22.40 \\
FreeNeRF~\citep{yang2023freenerf}& 0.58 & 0.76 & 0.80 & 0.70 & 0.80 & 0.84 & 0.84 & 0.80 & 0.94 & 0.94 & 0.92 & 0.90 & 0.82 \\
& 15.93 & 19.53 & 23.23 & 19.88 & 18.38 & 22.83 & 21.07 & 22.88 & 25.28 & 26.39 & 26.68 & 26.68 & 22.40 \\
\midrule
& 0.34 & 0.18 & 0.26 & 0.32 & 0.32 & 0.28 & 0.22 & 0.22 & 0.09 & 0.11 & 0.12 & 0.12 & 0.22\\
ViP-NeRF~\citep{somraj2023vip}& 0.33 & 0.58 & 0.58 & 0.53 & 0.47 & 0.55 & 0.50 & 0.43 & 0.66 & 0.65 & 0.77 & 0.60 & 0.55 \\
& 12.97 & 16.58 & 18.63 & 16.12 & 14.82 & 16.25 & 14.14 & 18.04 & 17.67 & 14.75 & 20.85 & 18.65 & 16.62 \\
\midrule
& 0.22 & 0.32 & 0.24 & 0.24 & 0.28 & 0.27 & 0.23 & 0.15 & 0.31 & 0.36 & 0.17 & 0.25 & 0.25 \\
SimpleNeRF~\citep{somraj2023simplenerf}& 0.74 & 0.68 & 0.74 & 0.75 & 0.75 & 0.77 & 0.79 & 0.90 & 0.77 & 0.67 & 0.84 & 0.81 & 0.77 \\
& 12.90 & 11.29 & 14.17 & 13.42 & 11.44 & 12.23 & 15.31 & 20.41 & 13.97 & 10.93 & 17.41 & 14.66 & 14.01 \\
\midrule
& 0.28 & 0.38 & 0.29 & 0.26 & 0.28 & 0.27 & 0.38 & 0.16 & 0.51 & 0.47 & 0.29 & 0.15 & 0.31 \\ 
VGOS~\citep{sun2023vgos}& 0.69 & 0.65 & 0.71 & 0.76 & 0.74 & 0.76 & 0.62 & 0.90 & 0.58 & 0.58 & 0.75 & 0.87 & 0.72 \\
& 9.84 & 8.34 & 10.50 & 11.91 & 8.51 & 9.14 & 7.27 & 18.86 & 5.38 & 5.80 & 11.81 & 16.74 & 10.34 \\
\midrule
& 0.23 & 0.12 & 0.15 & 0.37 & 0.14 & 0.14 & 0.12 & 0.14 & 0.04 & 0.04 & 0.11 & 0.08 & 0.14 \\
SparseNeRF~\citep{wang2023sparsenerf}& 0.63 & 0.81 & 0.79 & 0.59 & 0.84 & 0.84 & 0.84 & 0.84 & 0.96 & 0.95 & 0.90 & 0.92 & 0.83 \\
& 17.14 & 21.11 & 24.88 & 12.36 & 22.25 & 23.05 & 20.85 & 19.75 & 27.52 & 28.98 & 23.74 & 28.00 & 22.47 \\
\midrule
& 0.45 & 0.36 & 0.41 & 0.45 & 0.29 & 0.30 & 0.33 & 0.27 & 0.19 & 0.19 & 0.24 & 0.30 & 0.31 \\
ZeroRF~\citep{shi2023zerorf} & 0.33 & 0.55 & 0.47 & 0.41 & 0.65 & 0.68 & 0.57 & 0.68 & 0.84 & 0.83 & 0.74 & 0.63 & 0.61 \\
& 11.55 & 12.43 & 11.81 & 12.84 & 15.66 & 16.01 & 12.77 & 16.50 & 17.81 & 15.34 & 16.64 & 14.25 & 14.47 \\ 

\midrule
& 0.19 & 0.14 & 0.18 & 0.22 & 0.21 & 0.13 & 0.13 & 0.12 & 0.06 & 0.05 & 0.10 & 0.11 & 0.14 \\
FrugalNeRF (Ours) & 0.69 & 0.76 & 0.77 & 0.69 & 0.79 & 0.84 & 0.82 & 0.89 & 0.94 & 0.94 & 0.89 & 0.90 & 0.83 \\
& 17.38 & 19.06 & 22.38 & 18.96 & 17.77 & 24.01 & 20.35 & 26.11 & 24.57 & 25.85 & 25.43 & 27.28 & 22.43 \\
\midrule
& 0.19 & 0.13 & 0.17 & 0.21 & 0.20 & 0.13 & 0.13 & 0.12 & 0.06 & 0.05 & 0.08 & 0.10 & 0.13 \\
FrugalNeRF w/ mono. depth (Ours) & 0.68 & 0.78 & 0.78 & 0.73 & 0.79 & 0.84 & 0.82 & 0.88 & 0.95 & 0.93 & 0.91 & 0.91 & 0.83 \\
& 17.14 & 19.89 & 23.17 & 20.33 & 17.18 & 23.71 & 20.59 & 26.60 & 25.52 & 25.04 & 27.84 & 27.10 & 22.84 \\
 \bottomrule
\end{tabular}
}
\end{table*}
\begin{table*}[t]
\centering
\caption{\textbf{Quantitative results on the DTU~\citep{jensen2014large} dataset with four input views. The three rows show LPIPS, SSIM and PSNR scores, respectively.}}
\label{tab:suppl_DTU_4v}
\resizebox{\textwidth}{!}{%
\begin{tabular}{l|ccccccccccccc}
\toprule
\multicolumn{1}{r|}{Scene} & Scan21 & Scan31 & Scan34 & Scan38 & Scan40 & Scan41 & Scan45 & Scan55 & Scan63 & Scan82 & Scan103 & Scan114 & Average\\
\multicolumn{1}{l|}{Method} & & & & & & & & & & & & & \\
\midrule
& 0.18 & 0.14 & 0.13 & 0.24 & 0.14 & 0.12 & 0.09 & 0.06 & 0.04 & 0.03 & 0.08 & 0.07 & 0.11 \\
FreeNeRF~\citep{yang2023freenerf} & 0.72 & 0.81 & 0.83 & 0.72 & 0.85 & 0.86 & 0.86 & 0.92 & 0.96 & 0.96 & 0.93 & 0.93 & 0.86 \\
& 18.72 & 21.29 & 25.97 & 19.43 & 22.88 & 25.59 & 22.39 & 28.63 & 27.35 & 31.51 & 27.30 & 28.65 & 24.98 \\
\midrule
& 0.33 & 0.19 & 0.21 & 0.31 & 0.35 & 0.24 & 0.23 & 0.24 & 0.08 & 0.08 & 0.10 & 0.12 & 0.21 \\
ViP-NeRF~\citep{somraj2023vip}& 0.39 & 0.61 & 0.59 & 0.59 & 0.45 & 0.61 & 0.52 & 0.38 & 0.67 & 0.67 & 0.76 & 0.64 & 0.57 \\ & 14.24 & 17.22 & 19.44 & 18.19 & 15.76 & 18.84 & 15.57 & 16.62 & 17.19 & 16.45 & 22.67 & 19.50 & 17.64 \\
\midrule
& 0.27 & 0.28 & 0.23 & 0.25 & 0.32 & 0.27 & 0.25 & 0.21 & 0.27 & 0.27 & 0.18 & 0.29 & 0.26 \\
SimpleNeRF~\citep{somraj2023simplenerf} & 0.71 & 0.73 & 0.78 & 0.75 & 0.72 & 0.76 & 0.78 & 0.88 & 0.82 & 0.80 & 0.84 & 0.81 & 0.78 \\
& 11.81 & 12.95 & 14.72 & 12.71 & 10.42 & 11.67 & 14.12 & 18.84 & 14.05 & 14.43 & 16.87 & 14.23 & 13.90 \\
\midrule
& 0.27 & 0.35 & 0.31 & 0.28 & 0.27 & 0.27 & 0.37 & 0.16 & 0.43 & 0.42 & 0.28 & 0.18 & 0.30 \\ 
VGOS~\citep{sun2023vgos} & 0.73 & 0.69 & 0.71 & 0.74 & 0.76 & 0.78 & 0.64 & 0.90 & 0.66 & 0.66 & 0.75 & 0.85 & 0.74 \\
& 11.09 & 9.53 & 10.57 & 11.15 & 9.12 & 10.00 & 8.10 & 19.53 & 6.55 & 7.14 & 12.69 & 15.65 & 10.93 \\
\midrule
& 0.16 & 0.14 & 0.15 & 0.21 & 0.21 & 0.14 & 0.10 & 0.09 & 0.04 & 0.05 & 0.09 & 0.06 & 0.12 \\
SparseNeRF~\citep{wang2023sparsenerf}& 0.72 & 0.80 & 0.85 & 0.74 & 0.80 & 0.86 & 0.86 & 0.88 & 0.95 & 0.95 & 0.93 & 0.93 & 0.86 \\
& 18.60 & 20.99 & 25.87 & 20.92 & 19.45 & 24.81 & 22.15 & 26.37 & 26.20 & 26.72 & 28.10 & 28.19 & 24.03 \\
\midrule
& 0.43 & 0.32 & 0.28 & 0.44 & 0.28 & 0.25 & 0.20 & 0.29 & 0.17 & 0.14 & 0.26 & 0.32 & 0.28 \\
ZeroRF~\citep{shi2023zerorf} & 0.36 & 0.62 & 0.66 & 0.47 & 0.68 & 0.73 & 0.73 & 0.67 & 0.87 & 0.87 & 0.72 & 0.62 & 0.67 \\
& 11.75 & 13.48 & 16.47 & 13.53 & 16.87 & 17.26 & 16.48 & 15.92 & 19.33 & 19.12 & 15.18 & 13.36 & 15.73 \\ 
\midrule
& 0.17 & 0.12 & 0.16 & 0.17 & 0.19 & 0.12 & 0.12 & 0.12 & 0.05 & 0.04 & 0.07 & 0.10 & 0.12 \\
FrugalNeRF (Ours)  & 0.73 & 0.81 & 0.81 & 0.79 & 0.81 & 0.85 & 0.85 & 0.89 & 0.95 & 0.95 & 0.93 & 0.92 & 0.86 \\
& 19.21 & 21.84 & 24.99 & 23.08 & 19.47 & 25.64 & 21.59 & 27.31 & 26.27 & 27.26 & 29.27 & 28.21 & 24.51 \\
\midrule
& 0.17 & 0.12 & 0.15 & 0.17 & 0.19 & 0.12 & 0.11 & 0.12 & 0.05 & 0.03 & 0.07 & 0.09 & 0.12 \\
FrugalNeRF w/ mono. depth (Ours) & 0.73 & 0.81 & 0.82 & 0.80 & 0.82 & 0.86 & 0.86 & 0.90 & 0.96 & 0.95 & 0.93 & 0.92 & 0.86 \\
& 19.07 & 21.65 & 25.82 & 23.13 & 18.96 & 25.55 & 22.21 & 28.02 & 26.87 & 28.28 & 29.27 & 28.92 & 24.81 \\
 \bottomrule
\end{tabular}
}
\end{table*}

\paragraph{RealEstate-10K dataset.}
We show all 12 scenes of the quantitative comparisons with two, three, and four input views on the RealEstate-10K dataset in~\cref{tab:quantitative_real},~\cref{tab:suppl_Real_2v},~\cref{tab:suppl_Real_3v}, and~\cref{tab:suppl_Real_4v}.

\begin{table*}[t]
\centering
\caption{\textbf{Quantitative results on the RealEstate-10K~\cite{zhou2018stereo} dataset.} For SimpleNeRF~\cite{somraj2023simplenerf} and ViP-NeRF~\cite{somraj2023vip}, we calculate metrics using testing data provided in their respective clouds. As for other models, we rely on the scores provided in the SimpleNeRF paper.}
\label{tab:quantitative_real}
\resizebox{\textwidth}{!}{%
\begin{tabular}{l|c|c|ccc|ccc|ccc|c}

\toprule
 & & Learned & \multicolumn{3}{c|}{2-view} & \multicolumn{3}{c|}{3-view} & \multicolumn{3}{c|}{4-view} & Training \\
Method & Venue & priors & PSNR $\uparrow$ & SSIM $\uparrow$ & LPIPS $\downarrow$ & PSNR $\uparrow$ & SSIM $\uparrow$ & LPIPS $\downarrow$ & PSNR $\uparrow$ & SSIM $\uparrow$ & LPIPS $\downarrow$ & time $\downarrow$\\
\midrule
TensoRF~\cite{chen2022tensorf} & ECCV 2022 & - & 24.81 & 0.78 & 0.17 & 24.86 & 0.78 & 0.17 & 24.84 & 0.78 & 0.17 & 11 mins\\
\midrule
RegNeRF~\cite{niemeyer2022regnerf} & CVPR 2022 & normalizing flow & 16.87 & 0.59 & 0.45 & 17.73 & 0.61 & 0.44 & 18.25 & 0.62 & 0.44 & 2.35 hrs\\
DS-NeRF~\cite{deng2022depth} & CVPR 2022 & - & 25.44 & 0.79 & 0.32 & 25.94 & 0.79 & 0.32 & 26.28 & 0.79 & 0.33 & 3.5 hrs\\
DDP-NeRF~\cite{roessle2022dense} & CVPR 2022 & depth completion & 26.15 & 0.85 & 0.15 & 25.92 & 0.85 & 0.16 & 26.48 & 0.86 & 0.16 & 3.5 hrs\\
FreeNeRF~\cite{yang2023freenerf} & CVPR 2023 & - & 14.50 & 0.54 & 0.55 & 15.12 & 0.57 & 0.54 & 16.25 & 0.60 & 0.54 & \cellcolor{orange!25}1.5 hrs\\
ViP-NeRF~\cite{somraj2023vip} & SIGGRAPH 2023 & - & 29.55 & 0.87 & \cellcolor{orange!25}0.09 & 29.75 & \cellcolor{orange!25}0.88 & 0.11 & 30.47 & 0.88 & 0.11 & 13.5 hrs\\
SimpleNeRF~\cite{somraj2023simplenerf} & SIGGRAPH Asia 2023 & - & \cellcolor{red!25}30.30 & \cellcolor{red!25}0.88 & \cellcolor{red!25}0.07 & \cellcolor{red!25}31.40 & \cellcolor{red!25}0.89 & \cellcolor{orange!25}0.08 & \cellcolor{orange!25}31.73 & \cellcolor{orange!25}0.89 & \cellcolor{orange!25}0.09 & 9.5 hrs\\
FrugalNeRF (Ours) & - & - & \cellcolor{orange!25}30.12 & \cellcolor{orange!25}0.87 & \cellcolor{red!25}0.07 & 
\cellcolor{orange!25}31.04 & \cellcolor{red!25}0.89 & \cellcolor{red!25}0.06 & 
\cellcolor{red!25}31.78 & \cellcolor{red!25}0.90 & \cellcolor{red!25}0.06 & \cellcolor{red!25}20 mins\\
 \bottomrule
\end{tabular}%
}
\end{table*}
\newcolumntype{C}[1]{>{\centering\arraybackslash}p{#1}}

\begin{table*}[t]

\centering
\caption{\textbf{Quantitative results on the RealEstate-10K~\cite{zhou2018stereo} dataset with two input views. The three rows show LPIPS, SSIM, and PSNR scores, respectively.}}
\label{tab:suppl_Real_2v}
\resizebox{\textwidth}{!}{%
\begin{tabular}{l|*{6}{C{2.5cm}}}
\toprule
\multicolumn{1}{r|}{Scene} & 0 & 1 & 3 & 4 & 6 & Average\\
\multicolumn{1}{l|}{Method} & & & & & & \\
\midrule
& 0.35 & 0.32 & 0.49 & 0.54 & 0.54 & 0.45\\
RegNeRF~\cite{niemeyer2022regnerf}& 0.60 & 0.83 & 0.30 & 0.61 & 0.59 & 0.59 \\
& 16.51 & 21.04 & 13.88 & 17.13 & 15.79 & 16.87\\
\midrule 
& 0.26 & 0.27 & 0.51 & 0.24 & 0.31 & 0.32\\
DS-NeRF~\cite{deng2022depth}& 0.81 & 0.91 & 0.50 & 0.88 & 0.83 & 0.79\\
& 24.68 & 27.93 & 19.24 & 29.18	& 26.18 & 25.44\\
\midrule
& 0.11 & 0.12 & 0.34 & 0.06 & 0.11 & 0.15 \\
DDP-NeRF~\cite{roessle2022dense}& 0.89 & 0.95 & 0.56 & 0.94 & 0.92 & 0.85\\
& 25.90 & 25.87 & 18.97 & 32.01 & 28.00 & 26.15 \\
\midrule
& 0.45 & 0.50 & 0.64 & 0.67 & 0.48 & 0.55\\
FreeNeRF~\cite{yang2023freenerf}& 0.54 & 0.77 & 0.28 & 0.49	& 0.58 & 0.53\\
& 15.00 & 17.00 & 12.15 & 12.84	& 15.50 & 14.50\\
\midrule
& 0.05 & 0.05 & 0.22 & 0.04 & 0.08 & 0.09 \\
ViP-NeRF~\cite{somraj2023vip}& 0.94 & 0.97 & 0.56 & 0.95 & 0.93 & 0.87  \\
& 30.41 & 32.03 & 18.96 & 34.74 & 31.61 & 29.55\\
\midrule
& 0.04 & 0.04 & 0.21 & 0.03 & 0.05 & 0.07\\
SimpleNeRF~\cite{somraj2023simplenerf}& 0.95 & 0.97 & 0.56 & 0.95 & 0.96 & 0.88 \\
& 31.89 & 33.8 & 18.65 & 34.93 & 32.24 & 30.30 \\
\midrule
& 0.04 & 0.04 & 0.20 & 0.04 & 0.05 & 0.07 \\
FrugalNeRF (Ours) & 0.94 & 0.97 & 0.56 & 0.95 & 0.95 & 0.87 \\
& 30.13 & 34.69 & 18.35 & 35.00 & 32.45 & 30.12 \\
 \bottomrule
\end{tabular}

}
\end{table*}

\newcolumntype{C}[1]{>{\centering\arraybackslash}p{#1}}

\begin{table*}[t]

\centering
\caption{\textbf{Quantitative results on the RealEstate-10K~\cite{zhou2018stereo} dataset with three input views. The three rows show LPIPS, SSIM, and PSNR scores, respectively.}}
\label{tab:suppl_Real_3v}
\resizebox{\textwidth}{!}{%
\begin{tabular}{l|*{6}{C{2.5cm}}}
\toprule
\multicolumn{1}{r|}{Scene} & 0 & 1 & 3 & 4 & 6 & Average\\
\multicolumn{1}{l|}{Method} & & & & & & \\
\midrule
& 0.40 & 0.32 & 0.53 & 0.56	& 0.37 & 0.44 \\
RegNeRF~\cite{niemeyer2022regnerf}& 0.60 & 0.82	& 0.29 & 0.62	& 0.71 & 0.61 \\
& 15.99 & 20.89 & 13.87 & 17.60	& 20.28 & 17.73\\
\midrule 
& 0.24 & 0.26 & 0.53 & 0.26	& 0.31 & 0.32 \\
DS-NeRF~\cite{deng2022depth}& 0.83 & 0.91 & 0.49 & 0.87	& 0.85 & 0.79 \\
& 25.24 & 28.68 & 19.14 & 29.08	& 27.58 & 25.94\\
\midrule
& 0.11 & 0.11 & 0.38 & 0.06	& 0.13 & 0.16\\
DDP-NeRF~\cite{roessle2022dense}& 0.89 & 0.96 & 0.55 & 0.94 & 0.92 & 0.85\\
& 25.27 & 26.67 & 18.81 & 31.84	& 26.99 & 25.92 \\
\midrule
& 0.54 & 0.51 & 0.64 & 0.59	& 0.42 & 0.54\\
FreeNeRF~\cite{yang2023freenerf}& 0.53 & 0.75 & 0.29 & 0.61 & 0.66 & 0.57\\
& 13.79 & 15.59 & 12.45 & 15.72 & 18.05 & 15.12 \\
\midrule
& 0.06 & 0.10 & 0.26 & 0.04 & 0.08 & 0.11\\
ViP-NeRF~\cite{somraj2023vip} & 0.94 & 0.95 & 0.60 & 0.95 & 0.95 & 0.88 \\
& 30.66 & 29.89 & 19.59 & 35.17 & 33.43 & 29.75\\
\midrule
& 0.04 & 0.04 & 0.23 & 0.03 & 0.08 & 0.08 \\
SimpleNeRF~\cite{somraj2023simplenerf}& 0.95 & 0.98 & 0.61 & 0.95 & 0.95 & 0.89 \\
& 32.23 & 36.44 & 19.65 & 35.85 & 32.81 & 31.40 \\
\midrule
& 0.04 & 0.03 & 0.18 & 0.03 & 0.04 & 0.06 \\
FrugalNeRF (Ours) & 0.95 & 0.98 & 0.61 & 0.95 & 0.96 & 0.89 \\
& 31.11 & 35.39 & 18.85 & 35.78 & 34.07 & 31.04 \\
 \bottomrule
\end{tabular}

}
\end{table*}

\newcolumntype{C}[1]{>{\centering\arraybackslash}p{#1}}

\begin{table*}[t]

\centering
\caption{\textbf{Quantitative results on the RealEstate-10K~\cite{zhou2018stereo} dataset with four input views. The three rows show LPIPS, SSIM, and PSNR scores, respectively.}}
\label{tab:suppl_Real_4v}
\resizebox{\textwidth}{!}{%
\begin{tabular}{l|*{6}{C{2.7cm}}}
\toprule
\multicolumn{1}{r|}{Scene} & 0 & 1 & 3 & 4 & 6 & Average\\
\multicolumn{1}{l|}{Method} & & & & & & \\
\midrule
& 0.43 & 0.35 & 0.59 & 0.56 & 0.27 & 0.44 \\
RegNeRF~\cite{niemeyer2022regnerf}& 0.59 & 0.83	& 0.29 & 0.65	& 0.75 & 0.62\\
& 16.09 & 20.98 & 13.91 & 18.48 & 21.78 & 18.25\\
\midrule 
& 0.27 & 0.26 & 0.56 & 0.25	& 0.31 & 0.33 \\
DS-NeRF~\cite{deng2022depth}& 0.82 & 0.92 & 0.50 & 0.87	& 0.85 & 0.79\\
& 25.40 & 29.40 & 19.64 & 29.26	& 27.69 & 26.28\\
\midrule
& 0.12 & 0.08 & 0.39 & 0.06 & 0.13 & 0.16 \\
DDP-NeRF~\cite{roessle2022dense}& 0.89 & 0.96 & 0.58 & 0.93 & 0.91 & 0.86\\
& 25.14 & 28.57 & 19.57 & 31.73 & 27.36 & 26.48\\
\midrule
& 0.56 & 0.48 & 0.65 & 0.58 & 0.39 & 0.53 \\
FreeNeRF~\cite{yang2023freenerf}& 0.53 & 0.80 & 0.31 & 0.66	& 0.69 & 0.60 \\
& 13.84 & 17.93 & 12.69 & 17.29 & 19.48 & 16.25\\
\midrule
& 0.06 & 0.08 & 0.27 & 0.05 & 0.09 & 0.11 \\
ViP-NeRF~\cite{somraj2023vip}& 0.94 & 0.96 & 0.62 & 0.94 & 0.95 & 0.88 \\
& 31.64 & 32.24 & 20.35 & 34.84 & 33.28 & 30.47 \\
\midrule
& 0.04 & 0.05 & 0.24 & 0.03 & 0.09 & 0.09 \\
SimpleNeRF~\cite{somraj2023simplenerf}& 0.96 & 0.97 & 0.64 & 0.95 & 0.94 & 0.89\\
& 32.95 & 36.44 & 20.52 & 35.97 & 32.77 & 31.73 \\
\midrule
& 0.04 & 0.03 & 0.17 & 0.03 & 0.05 & 0.06 \\
FrugalNeRF (Ours) & 0.96 & 0.98 & 0.64 & 0.95 & 0.96 & 0.90 \\
& 32.29 & 36.06 & 19.81 & 36.54 & 34.22 & 31.78 \\
 \bottomrule
\end{tabular}

}
\end{table*}

\section{Additional Visual Comparisons}
\label{sec:supp_visual}
\paragraph{LLFF dataset.}
We show additional visual comparisons on the LLFF dataset with two input views in~\cref{fig:supp_llff_more_visual} and ~\cref{fig:supp_llff_more_depth}.

\begin{figure*}[t]
\centering
\resizebox{1.0\textwidth}{!} 
{
\includegraphics[width=\textwidth]{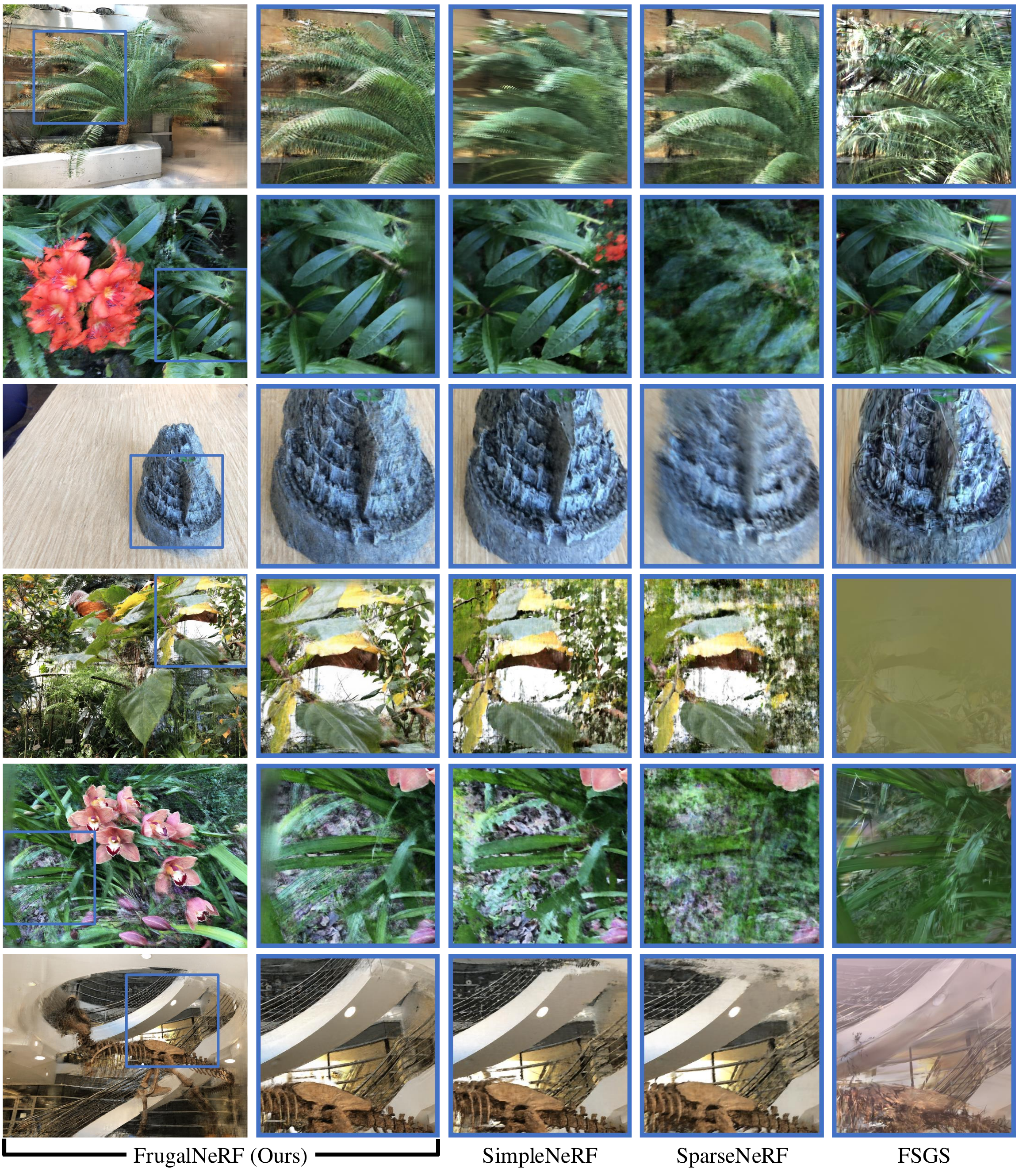}
}
\caption{\textbf{More qualitative comparisons on the LLFF~\cite{mildenhall2019llff} dataset with two input views.} FrugalNeRF achieves better synthesis quality in different scenes.}
\label{fig:supp_llff_more_visual}
\end{figure*}

\paragraph{DTU dataset.}
We show additional visual comparisons on the DTU dataset with two input views in~\cref{fig:supp_dtu_more_visual} and ~\cref{fig:supp_dtu_more_depth}.

\begin{figure*}[t]
\centering
\resizebox{1.0\textwidth}{!} 
{
\includegraphics[width=\textwidth]{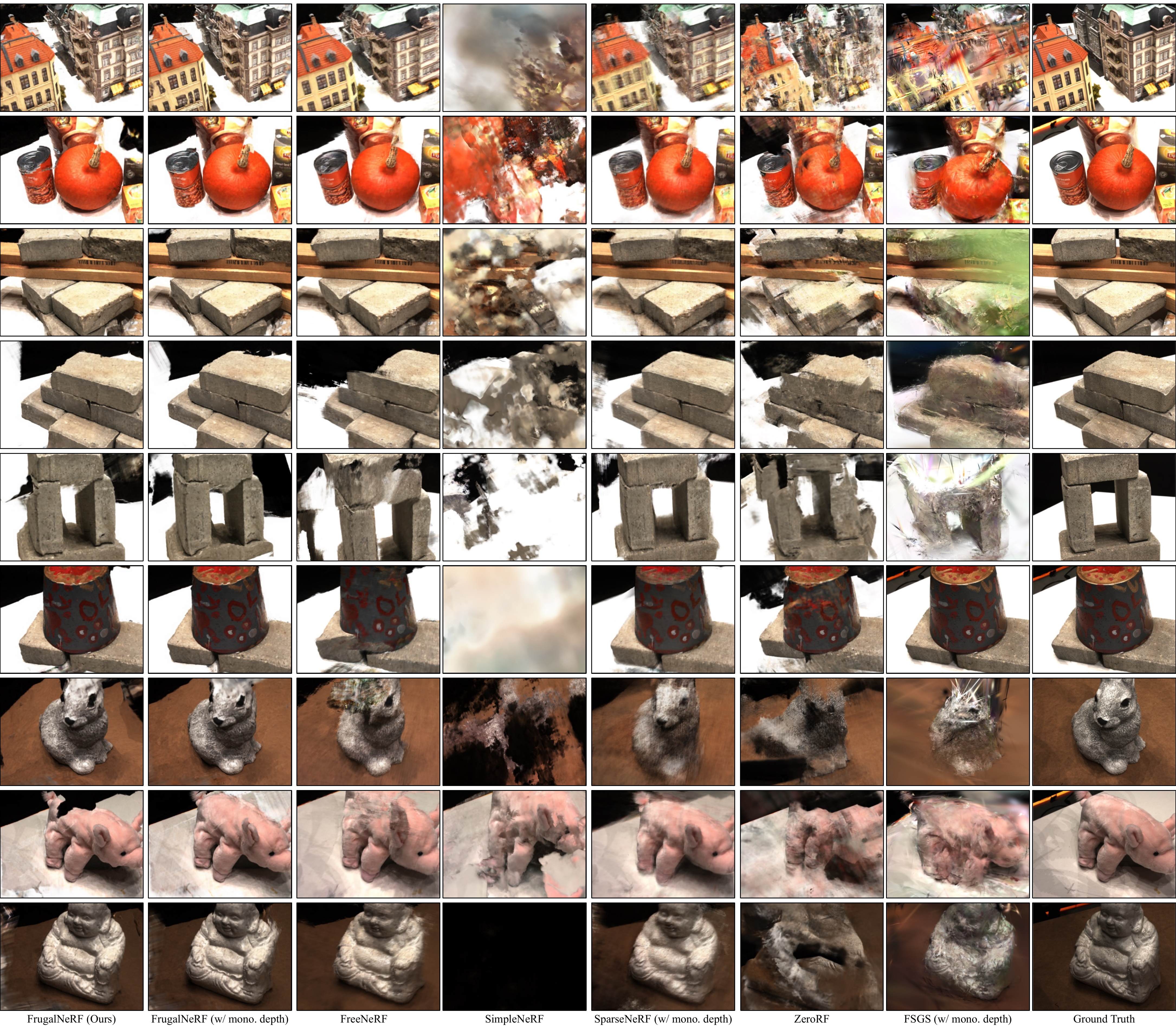}
}
\caption{\textbf{More qualitative comparisons on the DTU~\cite{jensen2014large} dataset with two input views.} FrugalNeRF achieves better synthesis quality in different scenes.}
\label{fig:supp_dtu_more_visual}
\end{figure*}

\paragraph{RealEstate-10K dataset.}
We further present the qualitative comparisons of novel view synthesis on the RealEstate-10K dataset with two input views in ~\cref{fig:real_visual} and ~\cref{fig:supp_real_more_visual}. Compared to SimpleNeRF~\citep{somraj2023simplenerf}, which requires hours of training, FrugalNeRF needs only less than 20 minutes and can render comparable results, demonstrating FrugalNeRF's effectiveness in more in-the-wild scenes.

\begin{figure*}[t]
\centering
\resizebox{1.0\textwidth}{!} 
{
\includegraphics[width=\textwidth]{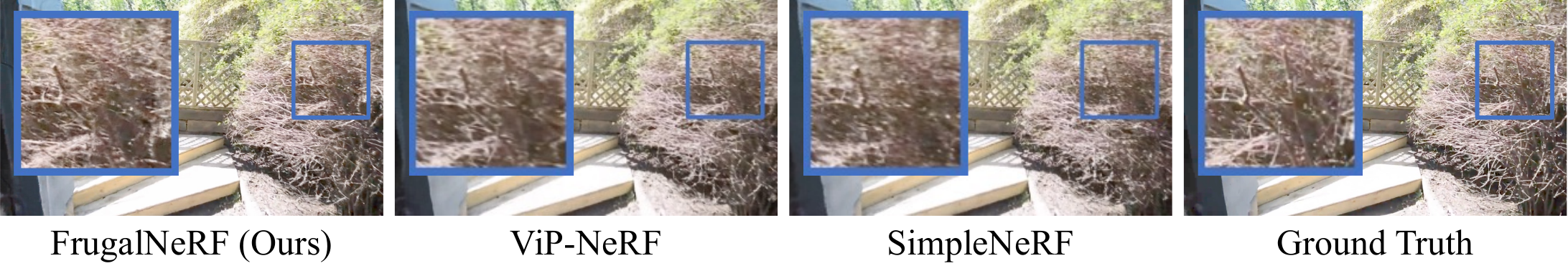}
}
\caption{\textbf{Qualitative comparisons on the RealEstate-10K~\cite{zhou2018stereo} dataset with two input views.} Compared to Vip-NeRF~\cite{somraj2023vip} and SimpleNeRF~\cite{somraj2023simplenerf}, our FrugalNeRF renders sharper details in the scene.}
\label{fig:real_visual}
\end{figure*}
\begin{figure*}[t]
\centering
\resizebox{1.0\textwidth}{!} 
{
\includegraphics[width=\textwidth]{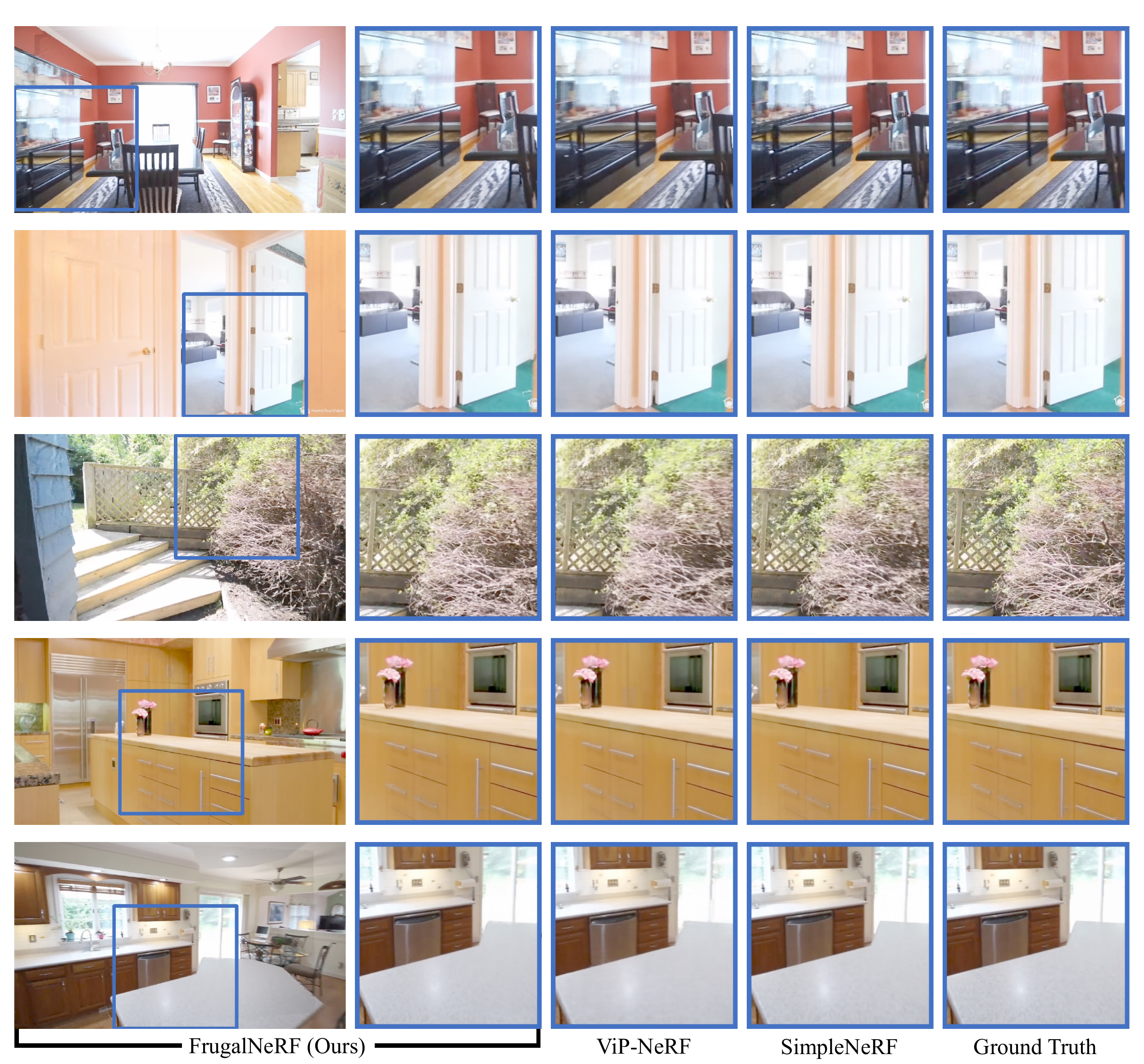}
}
\caption{\textbf{More qualitative comparisons on the RealEstate-10K~\cite{zhou2018stereo} dataset with two input views.} FrugalNeRF achieves synthesis quality comparable to the state-of-the-art methods.}
\label{fig:supp_real_more_visual}
\end{figure*}

\begin{figure*}[t]
\centering
\resizebox{1.0\textwidth}{!} 
{
\includegraphics[width=\textwidth]{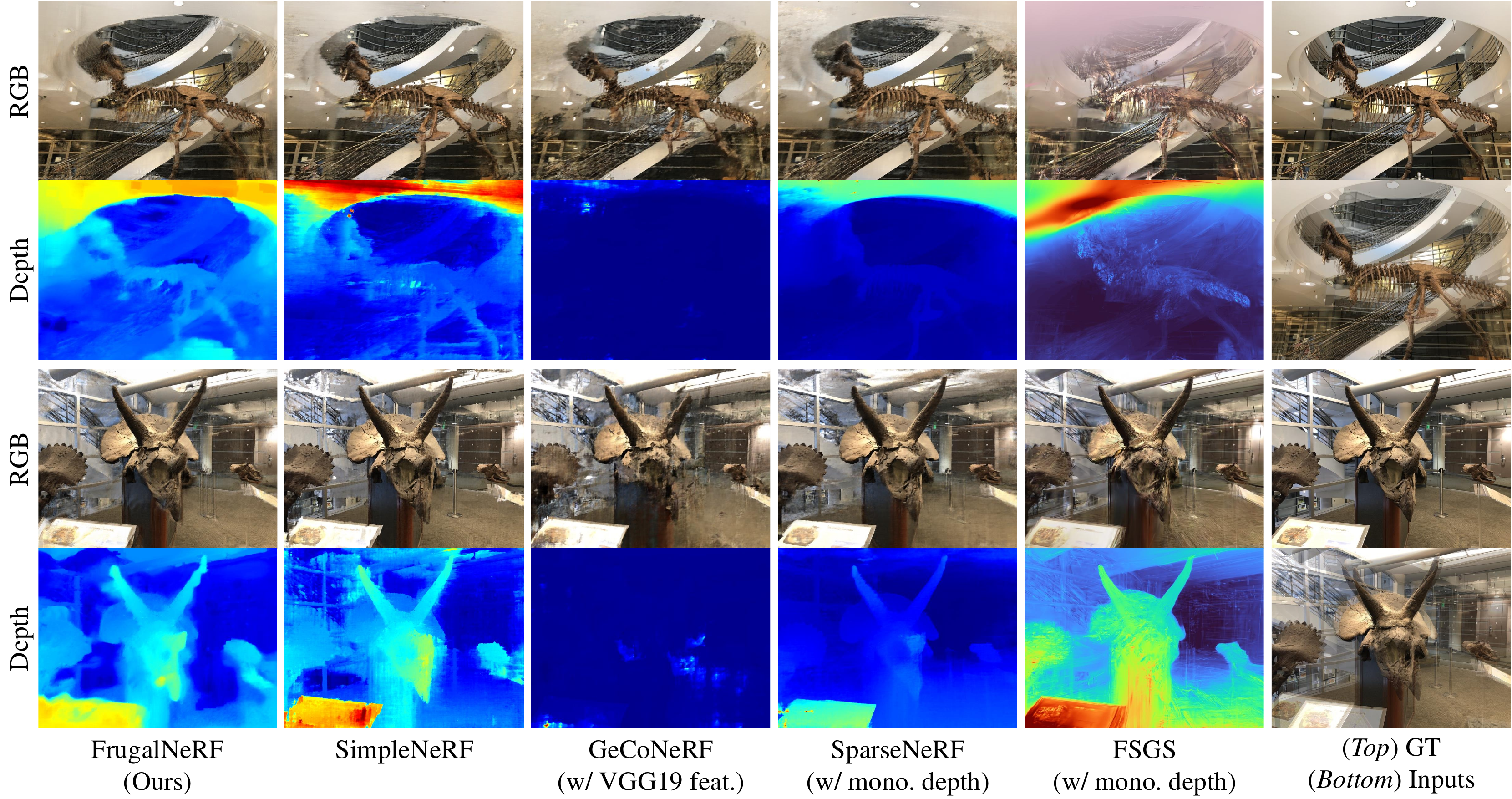}
}
\caption{\textbf{Render depth map comparisons on the LLFF~\cite{mildenhall2019llff} dataset with two input views.} FrugalNeRF achieves better synthesis quality in different scenes.}
\label{fig:supp_llff_more_depth}
\end{figure*}

\begin{figure*}[t]
\centering
\resizebox{1.0\textwidth}{!} 
{
\includegraphics[width=\textwidth]{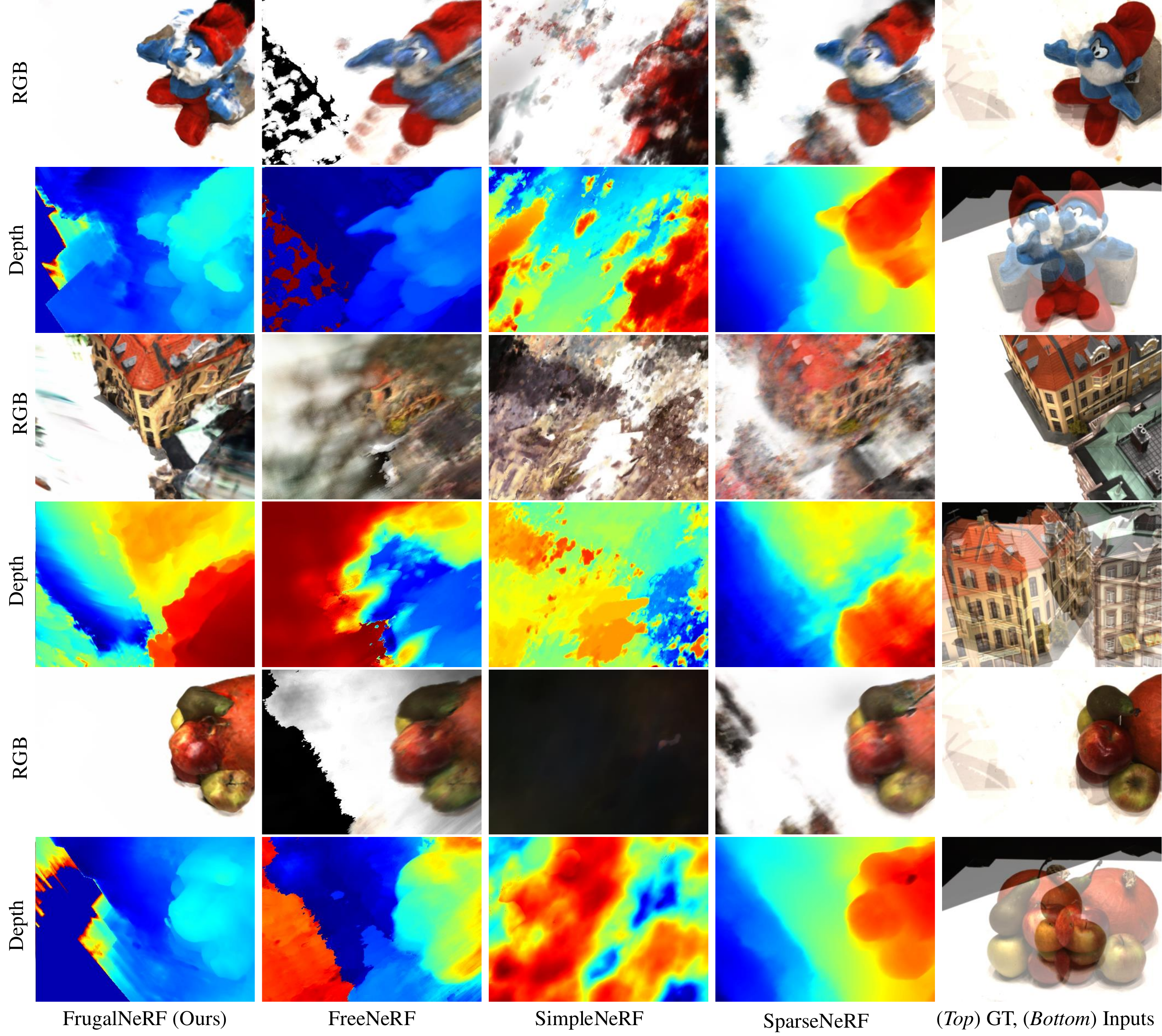}
}
\caption{\textbf{Render depth map comparisons on the DTU~\cite{jensen2014large} dataset with two input views.} FrugalNeRF achieves better synthesis quality in different scenes.}
\label{fig:supp_dtu_more_depth}
\end{figure*}

 \fi

\end{document}